\documentclass[journal]{IEEEtran}
\usepackage{amsmath,amsfonts}
\usepackage{algorithmic}
\usepackage{algorithm}
\usepackage{array}
\usepackage[caption=false,font=normalsize,labelfont=sf,textfont=sf]{subfig}
\usepackage{textcomp}
\usepackage{stfloats}
\usepackage{url}
\usepackage{verbatim}
\usepackage{graphicx}
\usepackage{cite}
\usepackage{hyperref}
\usepackage{color}
\usepackage{colortbl}

\newcommand{\nologo}{
  \raisebox{-0.25\height}{\includegraphics[width=0.4cm]{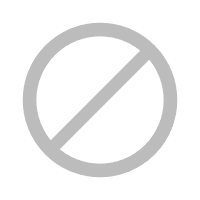}}
}
\newcommand{\rightlogo}{
  \raisebox{-0.25\height}{\includegraphics[width=0.4cm]{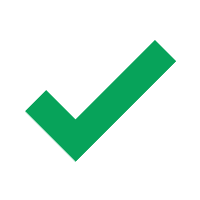}}
}

\newcommand{\visuallogo}{
  \raisebox{-0.3\height}{\includegraphics[width=0.4cm]{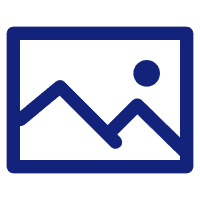}}
}
\newcommand{\textlogo}{
  \raisebox{-0.3\height}{\includegraphics[width=0.5cm]{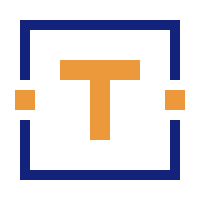}}
}
\newcommand{\audiologo}{
  \raisebox{-0.3\height}{\includegraphics[width=0.45cm]{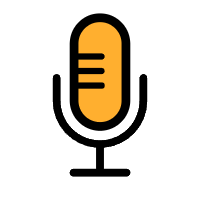}}
}
\newcommand{\partlogo}{
  \raisebox{-0.3\height}{\includegraphics[width=0.35cm]{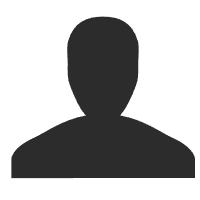}}
}
\newcommand{\holisticlogo}{
  \raisebox{-0.25\height}{\includegraphics[width=0.4cm]{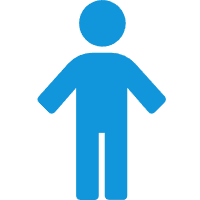}}
}
\newcommand{\mappinglogo}{
  \raisebox{-0.25\height}{\includegraphics[width=0.4cm]{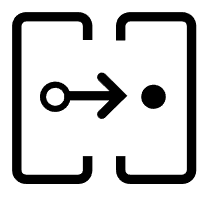}}
}
\newcommand{\llmlogo}{
  \raisebox{-0.25\height}{\includegraphics[width=0.4cm]{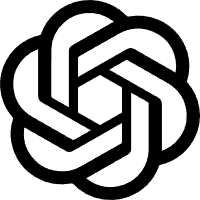}}
}
\newcommand{\handlogo}{
  \raisebox{-0.25\height}{\includegraphics[width=0.4cm]{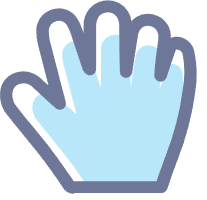}}
}
\newcommand{\facelogo}{
  \raisebox{-0.25\height}{\includegraphics[width=0.5cm]{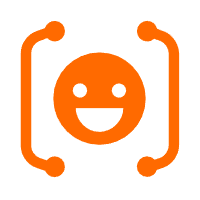}}
}

\newcommand{\acceleratelogo}{
  \raisebox{-0.4\height}{\includegraphics[width=0.5cm]{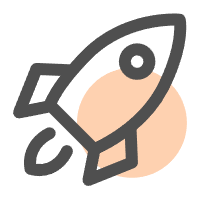}}
}

\newcommand{\realtimelogo}{
  \raisebox{-0.25\height}{\includegraphics[width=0.4cm]{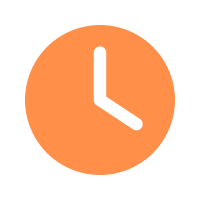}}
}

\newcommand{\reallogo}{
  \raisebox{-0.25\height}{\includegraphics[width=0.4cm]{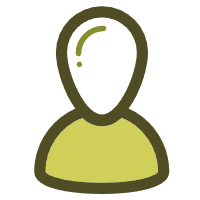}}
}
\newcommand{\poselogo}{
  \raisebox{-0.25\height}{\includegraphics[width=0.35cm]{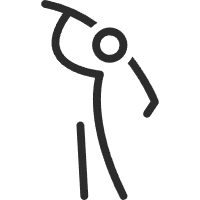}}
}
\newcommand{\drivingreallogo}{
  \raisebox{-0.25\height}{\includegraphics[width=0.35cm]{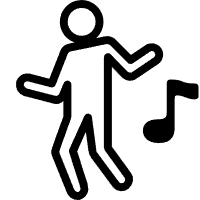}}
}

\newcommand{\palogo}{
  \raisebox{-0.25\height}{\includegraphics[width=0.4cm]{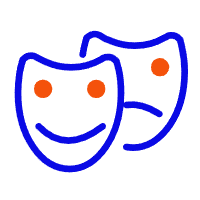}}
}
\newcommand{\dancelogo}{
  \raisebox{-0.25\height}{\includegraphics[width=0.35cm]{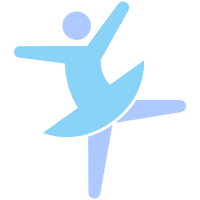}}
}
\newcommand{\posevideologo}{
  \raisebox{-0.25\height}{\includegraphics[width=0.35cm]{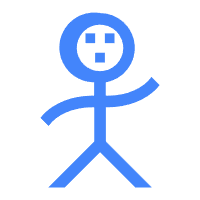}}
}
\newcommand{\textfacePartlogo}{
  \raisebox{-0.2\height}{\includegraphics[width=0.3cm]{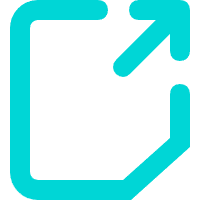}}
}
\newcommand{\textmotionlogo}{
  \raisebox{-0.25\height}{\includegraphics[width=0.4cm]{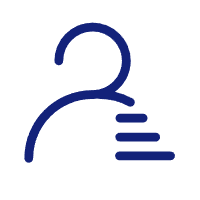}}
}
\newcommand{\liplogo}{
  \raisebox{-0.25\height}{\includegraphics[width=0.4cm]{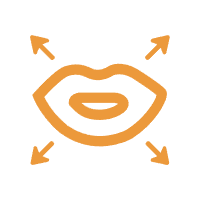}}
}
\newcommand{\talkingheadlogo}{
  \raisebox{-0.25\height}{\includegraphics[width=0.3cm]{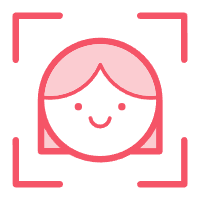}}
}
\newcommand{\holisticbodylogo}{
  \raisebox{-0.25\height}{\includegraphics[width=0.3cm]{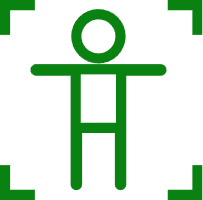}}
}
\usepackage{xcolor,colortbl}
\usepackage{multirow}
\usepackage{booktabs}
\usepackage[T1]{fontenc}
\usepackage{newtxtext}  
\newcommand{\new}[1]{#1}
\newcommand{\newR}[1]{\textcolor{black}{#1}}

\begin{document}

\title{Human Motion Video Generation: A Survey}

\author{
Haiwei Xue, 
Xiangyang Luo, 
Zhanghao Hu, 
Xin Zhang, 
Xunzhi Xiang, 
Yuqin Dai, \\
Jianzhuang Liu,~\IEEEmembership{Senior Member,~IEEE},
Zhensong Zhang, 
Minglei Li,  
Jian Yang, 
Fei Ma,\\
Zhiyong Wu,~\IEEEmembership{Member,~IEEE},
Changpeng Yang, 
Zonghong Dai,
and Fei Richard Yu,~\IEEEmembership{Fellow,~IEEE}
\IEEEcompsocitemizethanks{
\IEEEcompsocthanksitem 
H. Xue, X. Luo, and Z. Wu are with Tsinghua University. (E-mail: xhw22@mails.tsinghua.edu.cn, luo-xy24@mails.tsinghua.edu.cn, zywu@sz.
tsinghua.edu.cn)
\IEEEcompsocthanksitem 
J. Liu is with Shenzhen Institutes of Advanced Technology, Chinese Academy of Sciences. (E-mail: jz.liu@siat.ac.cn)
\IEEEcompsocthanksitem 
Z. Zhang is with Huawei Noah’s Ark Lab. (E-mail: zhangzhensong@huawei.com)
\IEEEcompsocthanksitem 
Z. Hu, M. Li, and C. Yang are with 01.AI. (E-mail: huzh666295@gmail.com, liminglei29@gmail.com, yangchangpeng@01.ai)
\IEEEcompsocthanksitem X. Zhang is with School of Mathematics and Statistics, Xi’an Jiaotong University. (E-mail: zhangxin0526@stu.xjtu.edu.cn)
\IEEEcompsocthanksitem Z. Dai is with Artificial Intelligence Innovation and Incubation (Al') Institute of Fudan University. (E-mail: zhdai23@m.fudan.edu.cn)
\IEEEcompsocthanksitem X. Xiang is with University of Chinese Academy of Sciences. (E-mail: xiangxunzhi21@mails.ucas.ac.cn)
\IEEEcompsocthanksitem Y. Dai and J. Yang are with PCA Lab, Nanjing University of Science and Technology. (E-mail: daiy@njust.edu.cn, csjyang@njust.edu.cn)
\IEEEcompsocthanksitem F. Ma and H. Xue are with Guangdong Laboratory of Artificial Intelligence and Digital Economy (SZ). (E-mail: mafei@gml.ac.cn)
\IEEEcompsocthanksitem Fei Richard Yu is with Shenzhen University and  Carleton University. (E-mail: richard.yu@ieee.org)
\IEEEcompsocthanksitem 
F. Ma and Z. Wu are the corresponding authors.}
}

\markboth{Xue \MakeLowercase{\textit{et al.}}: Human Motion Video Generation: A Survey}{}


\maketitle

\begin{abstract}
Human motion video generation has garnered significant research interest due to its broad applications, enabling innovations such as photorealistic singing heads or dynamic avatars that seamlessly dance to music. However, existing surveys in this field focus on individual methods, lacking a comprehensive overview of the entire generative process. This paper addresses this gap by providing an in-depth survey of human motion video generation, encompassing over ten sub-tasks, and detailing the five key phases of the generation process: input, motion planning, motion video generation, refinement, and output. Notably, this is the first survey that discusses the potential of large language models in enhancing human motion video generation. Our survey reviews the latest developments and technological trends in human motion video generation across three primary modalities: vision, text, and audio. By covering over two hundred papers, we offer a thorough overview of the field and highlight milestone works that have driven significant technological breakthroughs. Our goal for this survey is to unveil the prospects of human motion video generation and serve as a valuable resource for advancing the comprehensive applications of digital humans. 
A complete list of the models examined in this survey is available in \href{https://github.com/Winn1y/Awesome-Human-Motion-Video-Generation}{\textcolor{cyan}{Our Repository}}.
\end{abstract}


\section{Introduction}

\begin{figure}[t!]
    \centering
    \vspace{-10px}

    \subfloat[Quantity]{\includegraphics[width=0.56\linewidth]{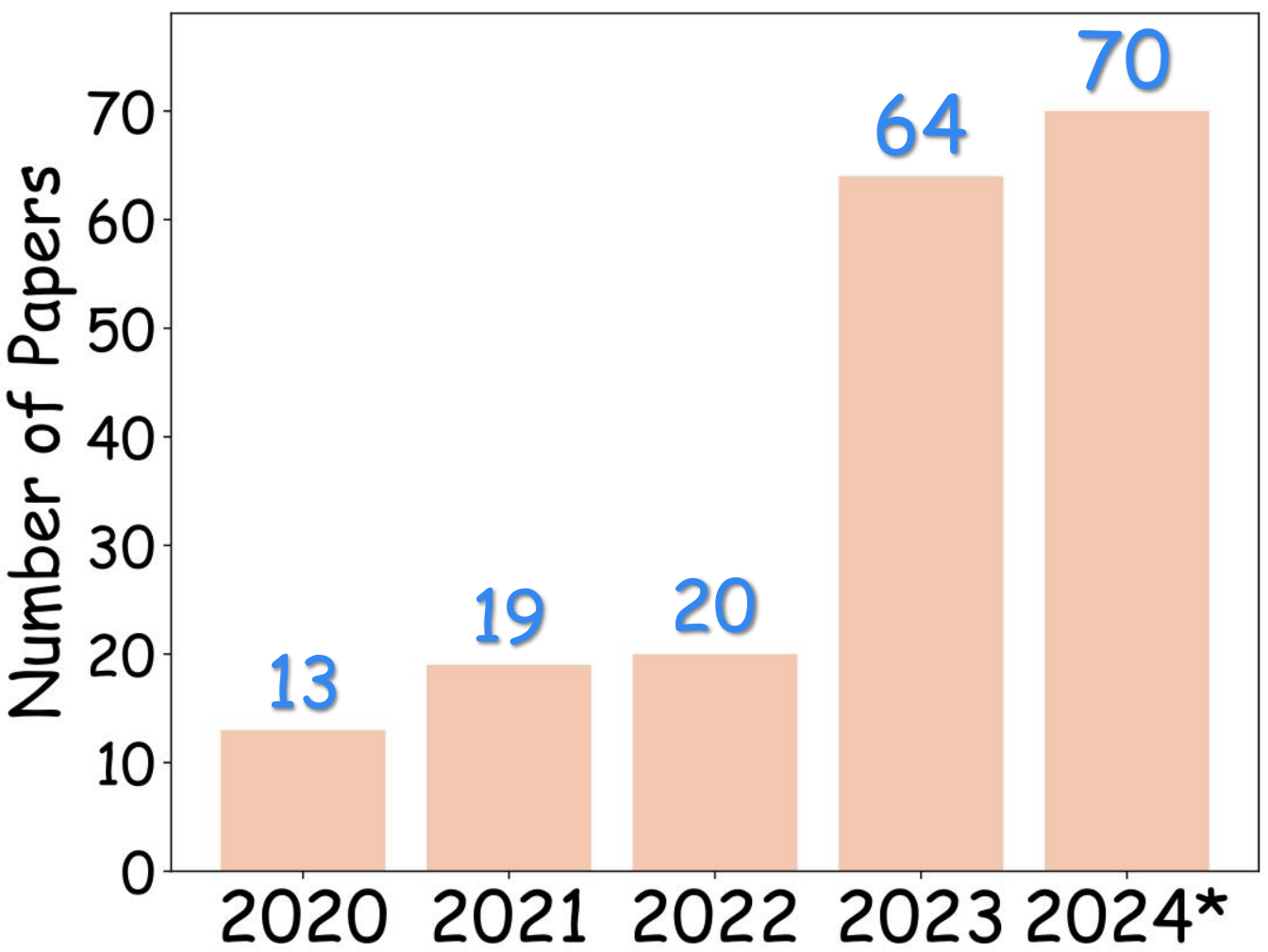}}
    \hfill
    \subfloat[Categories]{\includegraphics[width=0.44\linewidth]{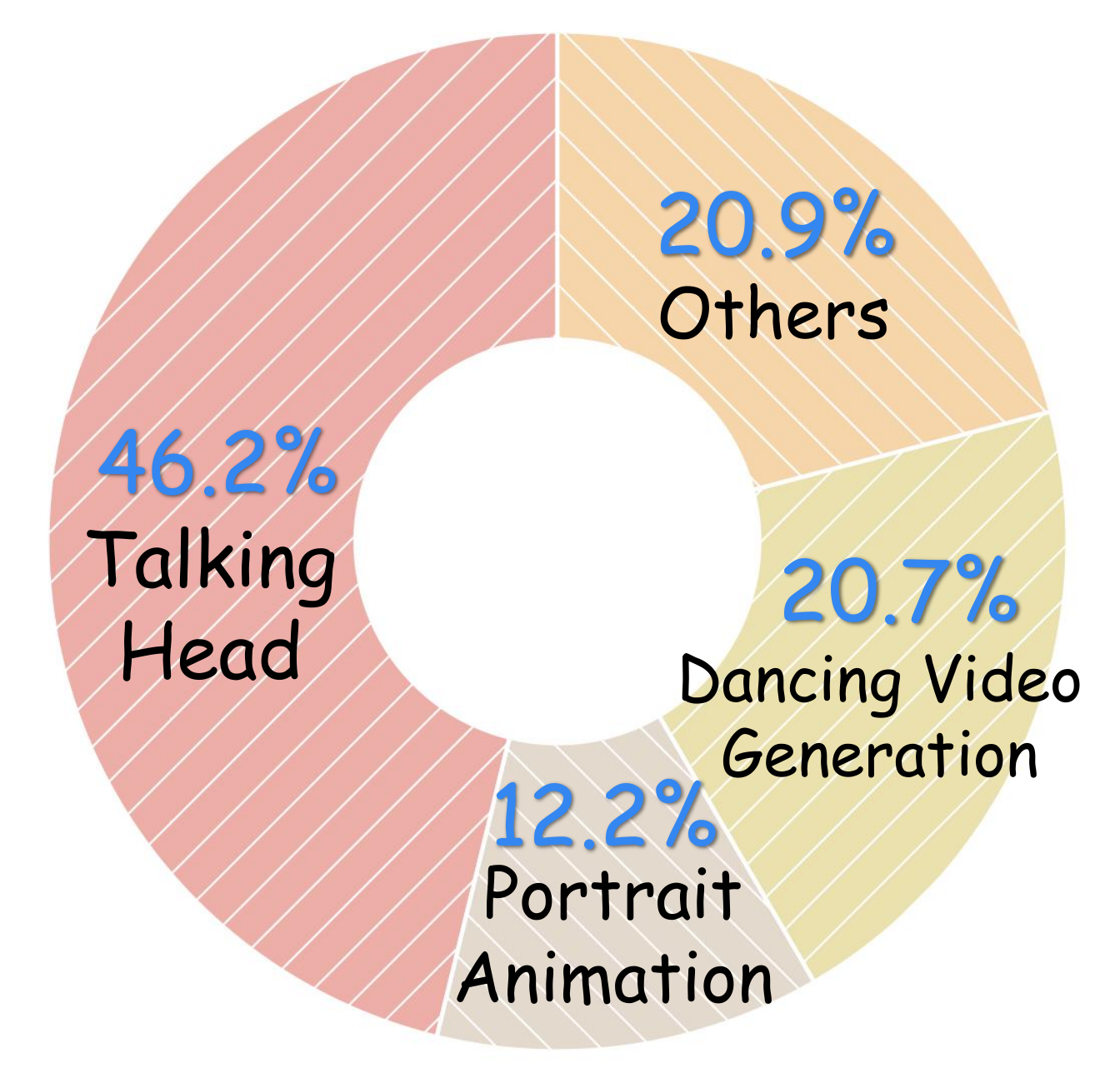}}
    
    \caption{Quantity of papers in the four categories reviewed in this survey, showing rapid growth in human motion video generation, emphasizing key areas such as talking head and dance videos. (2024* denotes the period from Jan. to Aug. in 2024.)
    }
    \label{fig:static}
    \vspace{-10px}
\end{figure}

\IEEEPARstart{H}{uman} motion video generation refers to the process of creating video sequences that depict realistic or stylized human movements based on various inputs such as vision cues, text prompts, and audio signals.
Early works in this field are limited to creating cartoon characters or human models lacking realistic textures~\cite{kim2018deep, fan2022faceformer}.
With the advent of general text-to-video generation models \cite{singer2022makeavideotexttovideogenerationtextvideo, guo2023animatediff,blattmann2023stable}, the field has expanded to produce videos with realistic textures and human-like quality. 

Recently, research on human-centered video synthesis \cite{jang2024faces, chang2023magicdance} has flourished, gaining significant interest in areas such as talking head, portrait animation, and dance video generation, as shown in Fig. \ref{fig:static}. 
To minimize the uncanny valley effect and enhance human-computer interactions, the generation of photorealistic human motion videos has emerged as a 
prominent topic, involving the creation of videos with human-like appearances, realistic motions, and natural expressions.

\begin{figure*}[t]
    \vspace{-15px}
    \centering
    \includegraphics[width=0.9\textwidth]{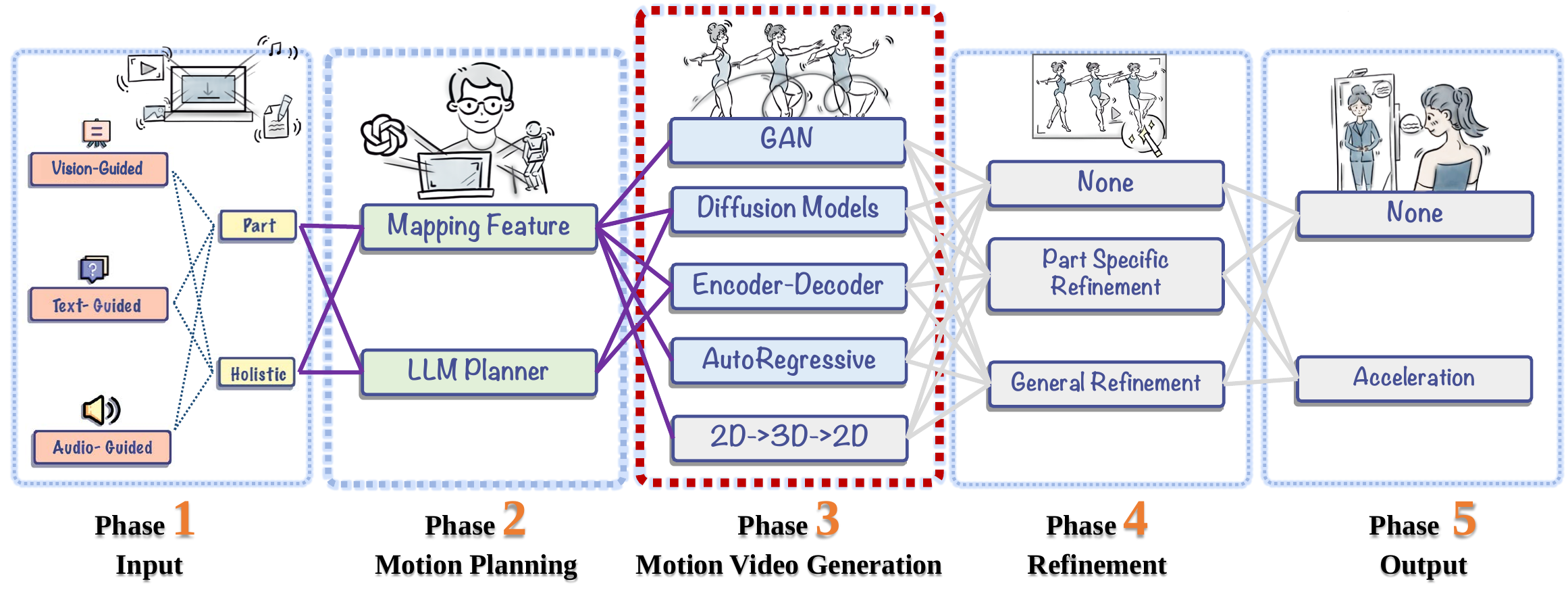}
    \vspace{-10px}
    \caption{
    Pipeline of generating human motion videos, which can be divided into five key phases.
    Initially, diverse input sources such as vision cues, text prompts, and audio signals are identified. 
    Next, these inputs guide the planning of human motions, either through feature mapping or utilizing LLMs. 
    The third phase focuses on modelling and translating these motion signals into video outputs according to the input conditions.
    Subsequently, the generated videos undergo refinement, with particular attention to optimizing details such as hand movements, synchronizing oral movements, adjusting gaze points, and enhancing the overall video quality. 
    Finally, efforts focus on reducing production costs, enabling real-time streaming platform integration of digital humans, and incorporating practical functionalities. It is worth noting that this review does not cover the reconstruction of 3D assets from 2D images or video generation through rendering techniques such as NeRF \cite{mildenhall2021nerf} or 3DGS \cite{kerbl20233d}. 
    }
    \vspace{-10px}
    \label{fig:pipeline}
\end{figure*}

Existing reviews \cite{survey1, survey2, song2023image, meng2024comprehensive, bigioi2023multilingual} often concentrate on specific subtasks within human motion video generation and do not provide a complete pipeline for generating human-centric videos.
Therefore, we clearly define \textit{five key phases} for the human motion video generation task, as illustrated in Fig. \ref{fig:pipeline}.
These phases collaboratively transform reference inputs into high-fidelity, lifelike human motion videos, enabling real-time and cost-efficient applications.
The process begins with identifying driving sources, which include vision cues, text prompts, or audio signals. 
Notably, generating facial regions and holistic human bodies often requires distinct frameworks. 
The second phase involves motion planning based on these inputs. While previous approaches \cite{shen2023difftalk,zhong2023identity} use input conditions to design implicit feature mappings for human motion, recent studies \cite{geng2023affective, wang2023agentavatar, wang2024instructavatar} explore the use of large language models (LLMs) for motion planning. 
The third phase focuses on generating human motion videos, ensuring body consistency, precise movements, and high-quality output. 
The fourth phase involves enhancing the generated videos, optimizing hand movements, synchronizing mouth and teeth movements, adjusting eye gaze, and improving overall video quality.
Finally, the fifth phase addresses the deployment of digital humans on real-time streaming platforms, integrating practical functionalities.

\begin{figure*}[t]
    \centering
    \vspace{-15px}
    \includegraphics[width=0.89\linewidth]{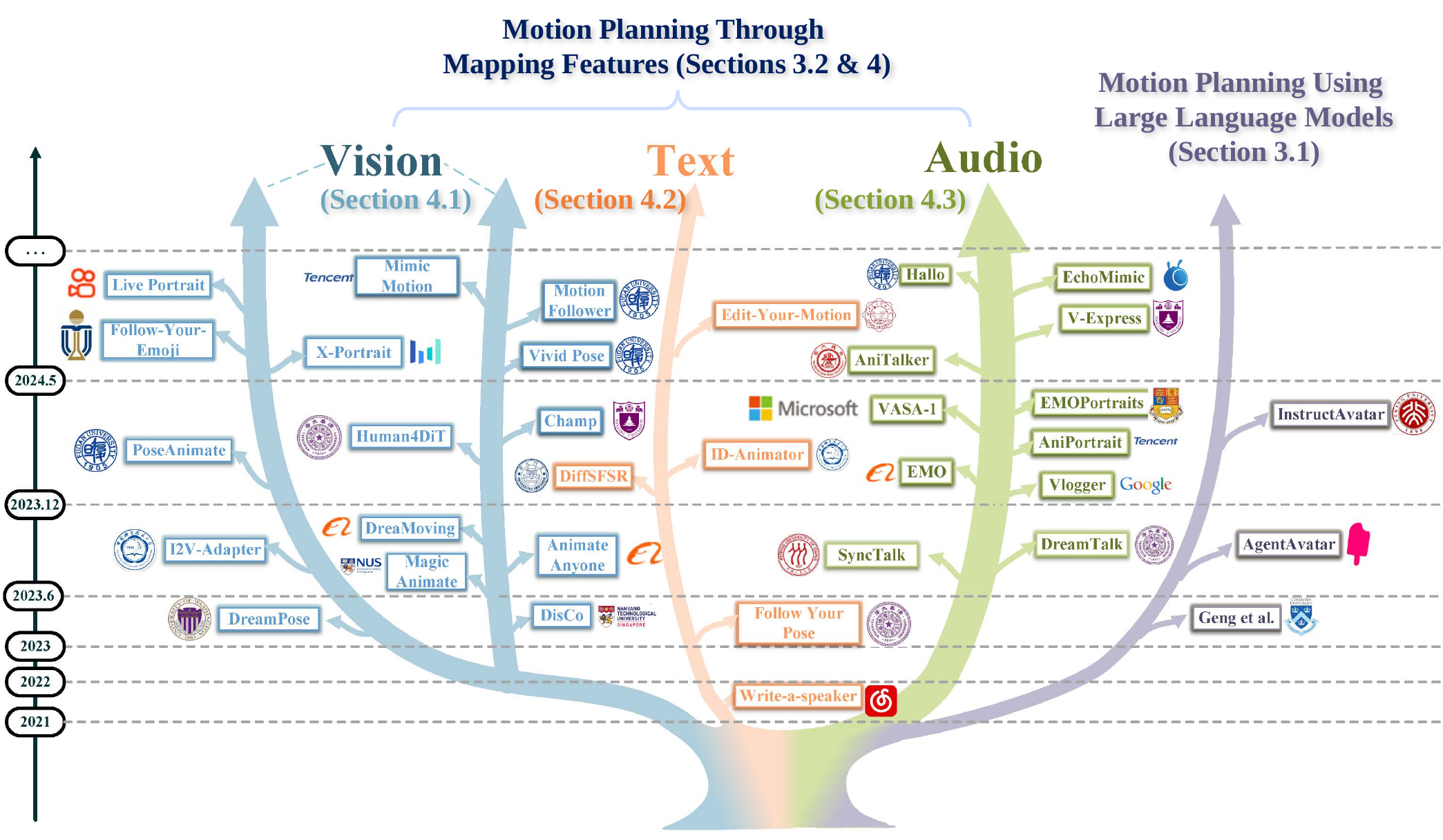}
    \vspace{-10px}
    \caption{Timeline of key advances in vision-, text-, and audio-driven human motion video generation methods.
    }
    \label{fig:timeline}
    \vspace{-10px}
\end{figure*}

We identify three primary modalities that drive human motion video generation: vision, text, and audio. Given that many methods involve multiple modalities, with vision being ubiquitous and audio having a stronger influence than text when both are present, we categorize methods based on the following criteria: 

\begin{itemize}

\item \new{\textbf{Audio-Driven} (Audio + Text + Vision or Audio + Vision). If a method includes audio, it is classified as audio-driven, even if other modalities are also involved.}

\item \new{ \textbf{Text-Driven} (Text + Vision). If a method is not in the audio-driven category but includes text, it is classified as text-driven.}

\item \new{\textbf{Vision-Driven} (only Vision). A method is classified as vision-driven if it only utilizes the vision modality. }

\end{itemize}
This classification system effectively organizes over 200 papers into these three distinct categories, and the timeline of representative works for these three driving modalities is illustrated in Fig.~\ref{fig:timeline}.

For vision-based conditions, portrait animation primarily focuses on generating specific facial expressions.
On a broader scale, pose-driven, video-driven dance video generation, and virtual Try-On techniques represent emerging research hotspots.
It should be noted that we emphasize the distinction between portrait animation and talking head. Portrait animation relies on reference images and pose sequences, whereas talking head utilizes reference images alongside driving audio, often referred to as audio-driven portrait animation.
Text-driven methods include Text2Face, which generates facial animations from first-personal scripts or instructions, and Text2MotionVideo, which extends this concept to create holistic human motions from text prompts. While much of the current research is focused on Text2Motion3D, as discussed by Zhu et al. \cite{zhu2023human}, our survey specifically addresses human motion video generation, placing 3D skeletal motions outside its scope.
In audio-driven scenarios, we elaborate on the related studies in the order of the driving region from small to large, including audio-lip synchronization, head pose driving, and holistic human body motion generation.
\new{
It is noteworthy that the methods discussed in this paper utilize multimodal conditional inputs, where the generation duration is determined by driving motion sequences.
}

Unlike previous surveys \cite{survey1, survey2} listed in Table \ref{tab:survey}, our work offers a comprehensive definition of the five key phases in human motion video generation. 
Notably, we are the first to discuss the application and potential of LLMs in motion planning. 
Our survey covers a broader spectrum, providing detailed categorization of various methods. 
Within this framework, various sub-branches emerge, each focusing on specific aspects of human motion video generation, as summarized in Table \ref{tab:taskdescription}.
Additionally, we collect 64 human-centered datasets to support related tasks, offering detailed information such as data duration and resolution. 
Moreover, we identify current challenges while offering insights into future research and development directions.
The main contributions of this paper are summarized as follows:

\begin{itemize}

\item We decompose human motion video generation into five key phases, covering all subtasks across various driving sources and body regions. To the best of our knowledge, this is the first survey to offer such a comprehensive framework for human motion video generation.

\item We provide an in-depth analysis of human motion video generation from both motion planning and motion generation perspectives, a
dimension that has been underexplored in existing reviews.

\item We clearly delineate established baselines and evaluation metrics, offering detailed insights into the key challenges shaping this field.

\item We present a set of potential future research directions, aiming for inspiring and guiding researchers in the field of human motion video generation.

\end{itemize}

\begin{table}[ht]
\centering
\caption{
Summary of related surveys. 
}
\label{tab:survey}
\resizebox{0.9\columnwidth}{!}{%
\begin{tabular}{|cccc|c|}
\hline
\rowcolor[HTML]{F1F8F8} 
\multicolumn{1}{|c|}{\cellcolor[HTML]{F1F8F8}\textbf{}}                                                              & \multicolumn{1}{c|}{\cellcolor[HTML]{F1F8F8}\textbf{Sha et al.   \cite{survey1}}} & \multicolumn{1}{c|}{\cellcolor[HTML]{F1F8F8}\textbf{Lei et al.   \cite{survey2}}} & \textbf{Ours} & \cellcolor[HTML]{F5F5F5}                      \\ \cline{1-4}
\multicolumn{1}{|c|}{Generation Pipeline}                                                                            & \multicolumn{1}{c|}{\nologo}                                                      & \multicolumn{1}{c|}{\nologo}                                                      & \rightlogo    & \cellcolor[HTML]{F5F5F5}                      \\ \cline{1-4}
\multicolumn{1}{|c|}{Motion Planning}                                                                                & \multicolumn{1}{c|}{\nologo}                                                      & \multicolumn{1}{c|}{\nologo}                                                      & \rightlogo    & \cellcolor[HTML]{F5F5F5}                      \\ \cline{1-4}
\multicolumn{1}{|c|}{LLMs as Motion Planner}                                                                          & \multicolumn{1}{c|}{\nologo}                                                      & \multicolumn{1}{c|}{\nologo}                                                      & \rightlogo    & \cellcolor[HTML]{F5F5F5}                      \\ \cline{1-4}
\multicolumn{1}{|c|}{Motion Video Generation}                                                                        & \multicolumn{1}{c|}{\rightlogo}                                                   & \multicolumn{1}{c|}{\rightlogo}                                                   & \rightlogo    & \cellcolor[HTML]{F5F5F5}                      \\ \cline{1-4}
\multicolumn{1}{|c|}{Refinement}                                                                                     & \multicolumn{1}{c|}{\nologo}                                                      & \multicolumn{1}{c|}{\nologo}                                                      & \rightlogo    & \cellcolor[HTML]{F5F5F5}                      \\ \cline{1-4}
\multicolumn{1}{|c|}{Structure Summary}                                                                              & \multicolumn{1}{c|}{\rightlogo}                                                   & \multicolumn{1}{c|}{\nologo}                                                      & \rightlogo    & \cellcolor[HTML]{F5F5F5}                      \\ \cline{1-4}
\multicolumn{1}{|c|}{Benchmarks}                                                                                     & \multicolumn{1}{c|}{\rightlogo}                                                   & \multicolumn{1}{c|}{\rightlogo}                                                   & \rightlogo    & \cellcolor[HTML]{F5F5F5}                      \\ \cline{1-4}
\multicolumn{1}{|c|}{Evaluation}                                                                                     & \multicolumn{1}{c|}{\nologo}                                                      & \multicolumn{1}{c|}{\nologo}                                                      & \rightlogo    & \cellcolor[HTML]{F5F5F5}                      \\ \cline{1-4}
\multicolumn{1}{|c|}{Challenges}                                                                                     & \multicolumn{1}{c|}{\nologo}                                                      & \multicolumn{1}{c|}{\rightlogo}                                                   & \rightlogo    & \multirow{-9}{*}{\cellcolor[HTML]{F5F5F5}}    \\ \hline
\rowcolor[HTML]{F1F8F8} 
\multicolumn{4}{|c|}{\cellcolor[HTML]{F1F8F8}\textbf{Sub-Tasks}}                                                                                                                                                                                                                                             & \textbf{Other Survey}                         \\ \hline
\multicolumn{1}{|c|}{Portrait Animation}                                                                             & \multicolumn{1}{c|}{\rightlogo}                                                   & \multicolumn{1}{c|}{\nologo}                                                      & \rightlogo    & \nologo                                       \\ \hline
\multicolumn{1}{|c|}{\begin{tabular}[c]{@{}c@{}}Video-Driven \\      Dance      Video Generation\end{tabular}}    & \multicolumn{1}{c|}{\nologo}                                                      & \multicolumn{1}{c|}{\nologo}                                                      & \rightlogo    & \nologo                                       \\ \hline
\multicolumn{1}{|c|}{\begin{tabular}[c]{@{}c@{}}Pose-Driven \\      Dance       Video Generation\end{tabular}}     & \multicolumn{1}{c|}{\rightlogo}                                                   & \multicolumn{1}{c|}{\rightlogo}                                                   & \rightlogo    & \nologo                                       \\ \hline
\multicolumn{1}{|c|}{Try-On}                                                                                         & \multicolumn{1}{c|}{\rightlogo}                                                   & \multicolumn{1}{c|}{\nologo}                                                      & \rightlogo    & Song et al.   \cite{song2023image}            \\ \hline
\multicolumn{1}{|c|}{Text2Face}                                                                                      & \multicolumn{1}{c|}{\rightlogo}                                                   & \multicolumn{1}{c|}{\nologo}                                                      & \rightlogo    & \nologo                                       \\ \hline
\multicolumn{1}{|c|}{Text2MotionVideo}                                                                               & \multicolumn{1}{c|}{\nologo}                                                      & \multicolumn{1}{c|}{\rightlogo}                                                   & \rightlogo    & \nologo                                       \\ \hline
\multicolumn{1}{|c|}{Talking Head}                                                                                   & \multicolumn{1}{c|}{\rightlogo}                                                   & \multicolumn{1}{c|}{\nologo}                                                      & \rightlogo    & Han Li et al.   \cite{li2025survey}    \\ \hline
\multicolumn{1}{|c|}{\begin{tabular}[c]{@{}c@{}}Audio Driven Holistic Human \\      Motion Gerneration\end{tabular}} & \multicolumn{1}{c|}{\nologo}                                                      & \multicolumn{1}{c|}{\rightlogo}                                                   & \rightlogo    & \nologo                                       \\ \hline
\multicolumn{1}{|c|}{\begin{tabular}[c]{@{}c@{}}Music-Driven \\      Dance       Video Generation\end{tabular}}    & \multicolumn{1}{c|}{\nologo}                                                      & \multicolumn{1}{c|}{\rightlogo}                                                   & \rightlogo    & \nologo                                       \\ \hline
\multicolumn{1}{|c|}{Multilingual Video Dubbing}                                                                                  & \multicolumn{1}{c|}{\nologo}                                                      & \multicolumn{1}{c|}{\nologo}                                                      & \rightlogo    & Bigioi et al.   \cite{bigioi2023multilingual} \\ \hline
\end{tabular}%
}
\vspace{-10px}
\end{table}

%
%
Given the significant advancements and extensive applications of human motion video generation, we present a comprehensive survey to help the community track its progress. The remaining sections of this survey are organized as follows:
Section~\ref{Sec:preliminaries} establishes the foundational knowledge necessary for understanding human motion video generation.
Next, Section~\ref{Sec:MP} introduces innovative approaches utilizing LLMs for human motion planning.
Following this, Section~\ref{Sec:MV} delves into the methodologies of motion modeling and video generation.
In Section~\ref{Sec:refinement}, we briefly explore the strategies in the final two phases: refinement and output.
In Section~\ref{Sec:Evaluation}, we consolidate popular metrics and datasets.
Finally, Section~\ref{sec:Challenges} addresses the current challenges and provides potential future directions in human motion video generation.
Note that all statistics referenced in this paper are current as of August 30, 2024. 

\begin{table*}[ht]
\vspace{-15px}
\centering
\caption{ 
Five phases in the human motion video generation pipeline, each emphasizing representative works and techniques across various tasks.
}
\vspace{-5px}
\label{tab:taskdescription}
\resizebox{0.97\textwidth}{!}{%
\begin{tabular}{|cc|c|c|c|c|cc}
\cline{1-6}
\multicolumn{2}{|c|}{Phase 1}                                                                                                                                                                                                                              & Phase 2                                                                                                       & Phase 3                                                                                                                    & Phase 4                                                                                                                                         & Phase 5                                                                                           &                                                                                                                                   &                                                                                                                                                                                                                                                                                          \\ \cline{1-6}
\multicolumn{1}{|c|}{Input}                                                                                                                          & Region                                                                                     & Motion Planning                                                                                                     & Generation Model                                                                                                         & Refinement                                                                                                                                    & Acceleration                                                                                    &                                                                                                                                   &                                                                                                                                                                                                                                                                                          \\ \hline
\multicolumn{1}{|c|}{\begin{tabular}[c]{@{}c@{}}\visuallogo Vision-driven Methods\\ \textlogo Text-driven Methods\\ \audiologo Audio-driven Methods\end{tabular}} & \begin{tabular}[c]{@{}c@{}}\partlogo Part (Face)\\      \\      \holisticlogo Holistic Human\end{tabular} & \begin{tabular}[c]{@{}c@{}}\mappinglogo Feature   Mapping\\      \\      \llmlogo LLMs Planner\end{tabular} & \begin{tabular}[c]{@{}c@{}}GAN\\ Diffusion Model (DM)\\ AutoRegressive (AR)\\ Encoder-Decoder (ED)\end{tabular} & \begin{tabular}[c]{@{}c@{}}\nologo None\\      \handlogo Hand Refinement\\      \facelogo Face Refinement\end{tabular} & \begin{tabular}[c]{@{}c@{}}\nologo None\\      \acceleratelogo Acceleration\end{tabular} & \multicolumn{1}{c|}{Tasks}                                                                                                         & \multicolumn{1}{c|}{Related Works}                                                                                                                                                                                                                                                       \\ \hline
\multicolumn{1}{|c|}{\visuallogo}                                                                                                                           & \partlogo                                                                                  & \mappinglogo                                                                                                & GAN / ED                                                                                                                 & \nologo / \facelogo                                                                                                                           & \nologo                                                                                         & \multicolumn{1}{c|}{\centering \multirow{8}{*}{Portrait Animation}}                                                                          & \multicolumn{1}{c|}{\begin{tabular}[c]{@{}c@{}}      \facelogo GazeGANV2   \cite{zhang2022unsupervised} \\      EDTN \cite{kang2024expression}\\      OTAvatar \cite{ma2023otavatar}\end{tabular}}                                                 \\ \cline{1-6} \cline{8-8} 
\multicolumn{1}{|c|}{\visuallogo}                                                                                                                           & \partlogo                                                                                  & \mappinglogo                                                                                                & DM                                                                                                                       & \nologo / \facelogo                                                                                                                           & \nologo                                                                                         & \multicolumn{1}{c|}{}                                                                                                             & \multicolumn{1}{c|}{\begin{tabular}[c]{@{}c@{}}Follow-Your-Emoji   \cite{ma2024follow}\\      \facelogo LivePortrait \cite{guo2024liveportrait} \\      X-Portrait \cite{xie2024x}\end{tabular}}                                                                                          \\ \cline{1-6} \cline{8-8} 
\multicolumn{1}{|c|}{\visuallogo}                                                                                                                           & \partlogo                                                                                  & \mappinglogo                                                                                                & DM                                                                                                                       & \nologo                                                                                                                                       & \acceleratelogo                                                                                 & \multicolumn{1}{c|}{}                                                                                                             & \multicolumn{1}{c|}{\acceleratelogo MobilePortrait \cite{jiang2024mobileportrait} }                                                                                                                                                                                                                       \\ \hline
\multicolumn{1}{|c|}{\visuallogo}                                                                                                                           & \holisticlogo                                                                              & \mappinglogo                                                                                                & GAN / ED                                                                                                                 & \nologo                                                                                                                                       & \nologo                                                                                         & \multicolumn{1}{c|}{\centering \multirow{2}{*}{\begin{tabular}[c]{@{}c@{}}Video-Driven \\      Dance       Video Generation\end{tabular}}} & \multicolumn{1}{c|}{\begin{tabular}[c]{@{}c@{}}EDN   \cite{chan2019everybody}\\      Human MotionFormer \cite{liu2023human}\end{tabular}}                                                                                                             \\ \cline{1-6} \cline{8-8} 
\multicolumn{1}{|c|}{\visuallogo}                                                                                                                           & \holisticlogo                                                                              & \mappinglogo                                                                                                & DM                                                                                                                       & \nologo                                                                                                                                       & \nologo                                                                                         & \multicolumn{1}{c|}{}                                                                                                             & \multicolumn{1}{c|}{BTDM \cite{adiya2023bidirectional}}                                                                                                                                                                                                                                  \\ \hline
\multicolumn{1}{|c|}{\visuallogo}                                                                                                                           & \holisticlogo                                                                              & \mappinglogo                                                                                                & DM                                                                                                                       & \nologo                                                                                                                                       & \nologo                                                                                         & \multicolumn{1}{c|}{\centering \multirow{6}{*}{\begin{tabular}[c]{@{}c@{}}Pose-Driven \\      Dance      Video Generation\end{tabular}}}  & \multicolumn{1}{c|}{\begin{tabular}[c]{@{}c@{}}DisCo \cite{wang2024disco}\\      Animate Anyone \cite{hu2024animate}\\      Follow-Your-Pose v2 \cite{xue2024follow}\end{tabular}}                                                                                                       \\ \cline{1-6} \cline{8-8} 
\multicolumn{1}{|c|}{\visuallogo}                                                                                                                           & \holisticlogo                                                                              & \mappinglogo                                                                                                & DM (DiT)                                                                                                                       & \nologo                                                                                                                                       & \nologo                                                                                         & \multicolumn{1}{c|}{}                                                                                                             & \multicolumn{1}{c|}{Human4DiT \cite{shao2024human4dit}}                                                                                                                                                                                                                                  \\ \cline{1-6} \cline{8-8} 
\multicolumn{1}{|c|}{\visuallogo}                                                                                                                           & \holisticlogo                                                                              & \mappinglogo                                                                                                & DM                                                                                                                       & \handlogo                                                                                                                                     & \nologo                                                                                         & \multicolumn{1}{c|}{}                                                                                                             & \multicolumn{1}{c|}{\handlogo MimicMotion \cite{zhang2024mimicmotion} }                                                                                                                                                                                                                             \\ \cline{1-6} \cline{8-8} 
\multicolumn{1}{|c|}{\visuallogo}                                                                                                                           & \holisticlogo                                                                              & \mappinglogo                                                                                                & DM                                                                                                                       & \nologo                                                                                                                                       & \acceleratelogo                                                                                 & \multicolumn{1}{c|}{}                                                                                                             & \multicolumn{1}{c|}{\acceleratelogo I2V-Adapter \cite{guo2023i2v} }                                                                                                                                                                                                                                       \\ \hline
\multicolumn{1}{|c|}{\visuallogo}                                                                                                                           & \holisticlogo                                                                              & \mappinglogo                                                                                                & DM                                                                                                                       & \nologo                                                                                                                                       & \nologo                                                                                         & \multicolumn{1}{c|}{Try-On}                                                                                                       & \multicolumn{1}{c|}{\begin{tabular}[c]{@{}c@{}}ViViD \cite{fang2024vivid}\end{tabular}}                                                                                                             \\ \hline
\multicolumn{1}{|c|}{\visuallogo}                                                                                                                           & \holisticlogo                                                                              & \mappinglogo                                                                                                & DM                                                                                                                       & \nologo                                                                                                                                       & \nologo                                                                                         & \multicolumn{1}{c|}{Pose2Video}                                                                                                   & \multicolumn{1}{c|}{\begin{tabular}[c]{@{}c@{}}DreamPose   \cite{karras2023dreampose}\\      Make-Your-Anchor \cite{huang2024make}\end{tabular}}                                                                                                                                         \\ \hline
\multicolumn{1}{|c|}{\textlogo}                                                                                                                             & \partlogo                                                                                  & \mappinglogo                                                                                                & GAN / ED                                                                                                                 & \nologo                                                                                                                                       & \nologo                                                                                         & \multicolumn{1}{c|}{\centering \multirow{5}{*}{Text2Face}}                                                                                   & \multicolumn{1}{c|}{\begin{tabular}[c]{@{}c@{}}Write-a-speaker   \cite{li2021write}\\           Faces that Speak \cite{jang2024faces}\end{tabular}}                                                                                                     \\ \cline{1-6} \cline{8-8} 
\multicolumn{1}{|c|}{\textlogo}                                                                                                                             & \partlogo                                                                                  & \mappinglogo                                                                                                & DM                                                                                                                       & \nologo                                                                                                                                       & \nologo                                                                                         & \multicolumn{1}{c|}{}                                                                                                             & \multicolumn{1}{c|}{ID-Animator \cite{he2024id}}                                                                                                                                                                                                                                         \\ \cline{1-6} \cline{8-8} 
\multicolumn{1}{|c|}{\textlogo}                                                                                                                             & \partlogo                                                                                  & \llmlogo                                                                                                    & DM                                                                                                                       & \nologo                                                                                                                                       & \nologo                                                                                         & \multicolumn{1}{c|}{}                                                                                                             & \multicolumn{1}{c|}{Geng et al. \cite{geng2023affective}}                                                                                                                                                                                                                                \\ \hline
\multicolumn{1}{|c|}{\textlogo}                                                                                                                             & \holisticlogo                                                                              & \mappinglogo                                                                                                & DM                                                                                                                       & \nologo                                                                                                                                       & \nologo                                                                                         & \multicolumn{1}{c|}{\centering \multirow{2}{*}{Text2MotionVideo}}                                                                            & \multicolumn{1}{c|}{\begin{tabular}[c]{@{}c@{}}Edit-Your-Motion   \cite{zuo2024edit}\\        \new{Follow-Your-Pose v1 \cite{ma2024followyp}}\end{tabular}}                                                         \\ \cline{1-6} \cline{8-8} 
\multicolumn{1}{|c|}{\textlogo}                                                                                                                             & \holisticlogo                                                                              & \mappinglogo                                                                                                & ED                                                                                                                       & \nologo                                                                                                                                       & \nologo                                                                                         & \multicolumn{1}{c|}{}                                                                                                             & \multicolumn{1}{c|}{Text2Performer \cite{jiang2023text2performer}}                                                                                                                                                                                                                       \\ \hline
\multicolumn{1}{|c|}{\audiologo}                                                                                                                            & \partlogo                                                                                  & \mappinglogo                                                                                                & GAN / ED                                                                                                                 & \nologo                                                                                                                                       & \nologo                                                                                         & \multicolumn{1}{c|}{\centering \multirow{13}{*}{Talking Head}}                                                                                & \multicolumn{1}{c|}{\begin{tabular}[c]{@{}c@{}}StyleHEAT   \cite{yin2022styleheat}\\      PC-AVS \cite{zhou2021pose}\\      EDTalk \cite{tan2024edtalk}\end{tabular}}                                                                                                                    \\ \cline{1-6} \cline{8-8} 
\multicolumn{1}{|c|}{\audiologo}                                                                                                                            & \partlogo                                                                                  & \llmlogo                                                                                                    & ED                                                                                                                       & \nologo                                                                                                                                       & \nologo                                                                                         & \multicolumn{1}{c|}{}                                                                                                             & \multicolumn{1}{c|}{AgentAvatar \cite{wang2023agentavatar}}                                                                                                                                                                                                                              \\ \cline{1-6} \cline{8-8} 
\multicolumn{1}{|c|}{\audiologo}                                                                                                                            & \partlogo                                                                                  & \mappinglogo                                                                                                & DM                                                                                                                       & \nologo                                                                                                                                       & \nologo                                                                                         & \multicolumn{1}{c|}{}                                                                                                             & \multicolumn{1}{c|}{\begin{tabular}[c]{@{}c@{}}      EchoMimic \cite{chen2024echomimic}\\       EMO \cite{tian2024emo}\\      Hallo \cite{xu2024hallo}\end{tabular}} \\ \cline{1-6} \cline{8-8} 
\multicolumn{1}{|c|}{\audiologo}                                                                                                                            & \partlogo                                                                                  & \llmlogo                                                                                                    & DM                                                                                                                       & \nologo                                                                                                                                       & \nologo                                                                                         & \multicolumn{1}{c|}{}                                                                                                             & \multicolumn{1}{c|}{InstructAvatar \cite{wang2024instructavatar}}                                                                                                                                                                                                                        \\ \cline{1-6} \cline{8-8} 
\multicolumn{1}{|c|}{\audiologo}                                                                                                                            & \partlogo                                                                                  & \mappinglogo                                                                                                & DM                                                                                                                       & \facelogo                                                                                                                                     & \nologo                                                                                         & \multicolumn{1}{c|}{}                                                                                                             & \multicolumn{1}{c|}{\facelogo Liang et al. \cite{liang2024emotional} }                                                                                                                                                                                                                              \\ \cline{1-6} \cline{8-8} 
\multicolumn{1}{|c|}{\audiologo}                                                                                                                            & \partlogo                                                                                  & \mappinglogo                                                                                                & AR                                                                                                                       & \nologo                                                                                                                                       & \nologo                                                                                         & \multicolumn{1}{c|}{}                                                                                                             & \multicolumn{1}{c|}{MakeItTalk \cite{zhou2020makelttalk}}                                                                                                                                                                                                                                \\ \cline{1-6} \cline{8-8} 
\multicolumn{1}{|c|}{\audiologo}                                                                                                                            & \partlogo                                                                                  & \mappinglogo                                                                                                & AR                                                                                                                       & \nologo                                                                                                                                       & \acceleratelogo                                                                                 & \multicolumn{1}{c|}{}                                                                                                             & \multicolumn{1}{c|}{\acceleratelogo Live Speech Portraits \cite{lu2021live} }                                                                                                                                                                                                                             \\ \hline
\multicolumn{1}{|c|}{\audiologo}                                                                                                                            & \holisticlogo                                                                              & \mappinglogo                                                                                                & DM                                                                                                                       & \nologo                                                                                                                                       & \nologo                                                                                         
& \multicolumn{1}{c|}{\begin{tabular}[c]{@{}c@{}}Audio-driven \\ Holistic Human Motion Driving\end{tabular}}
& \multicolumn{1}{c|}{Vlogger \cite{corona2024vlogger}}                                                                                                                                                                                                                                    \\ \hline
\multicolumn{1}{|c|}{\audiologo}                                                                                                                            & \holisticlogo                                                                              & \mappinglogo                                                                                                & DM                                                                                                                       & \nologo                                                                                                                                       & \nologo                                                                                         
& \multicolumn{1}{c|}{\begin{tabular}[c]{@{}c@{}}Music-Driven \\ Dance Video Generation\end{tabular}}
& \multicolumn{1}{c|}{Dance-Any-Beat \cite{wang2024dance}}                                                                                                                                                                                                                                 \\ \hline
\end{tabular}%
}
\vspace{-10px}
\end{table*}

\vspace{-10px}
\section{Preliminaries}
\label{Sec:preliminaries}

\subsection{Generative Frameworks}

\noindent \textbf{Variational Autoencoder~(VAE). } 
VAE, introduced in 2013 by Kingma and Welling~\cite{kingma2013auto}, has become a prominent generative model, recognized for its robust data representation capabilities.
In the human motion video generation task, VAE and its variants are instrumental in encoding vision signals or reference images, facilitating the generation of corresponding videos~\cite{yegeneface,ye2023geneface++}.
These factors lead to the adoption of variants, such as VQ-VAE~\cite{van2017neural}.
However, VAE is susceptible to mode collapse and often produces samples that are less sharp compared to those generated by generative adversarial networks.
Thus, VAE is combined with diffusion models to enhance generated output quality.

\noindent \textbf{Generative Adversarial Networks (GANs). } 
First proposed by Ian Goodfellow et al. in 2014~\cite{goodfellow2014generative}, GANs comprise two adversarial neural networks: a generator \( G \) and a discriminator \( D \). 
The generator creates data that mimics real samples, while the discriminator distinguishes between real and generated data. Unlike the degradation in generation quality caused by the strong mathematical priors in VAE, GANs implicitly map feature relationships, resulting in higher-quality generated outputs. Variants such as StyleGAN~\cite{karras2019style,karras2020analyzing} achieve significant milestones, particularly in human motion video generation, including tasks like motion copy~\cite{ma2024identity, chan2019everybody} and talking head synthesis~\cite{jang2024faces, li2021write}. However, GANs are limited by the lack of diversity in generated samples and are prone to mode collapse, primarily due to the challenges in balancing the training dynamics between the generator and discriminator.

\noindent \textbf{Diffusion Models (DMs). }
Diffusion models~\cite{sohl2015deep} attract widespread attention in the field of generative modelling~\cite{song2020improved, nichol2021improved, song2020denoising, dhariwal2021diffusion, cao2024survey}. 
These models generate samples by progressively denoising an initially noisy input, and their training objective can be expressed as a reweighted variational lower bound~\cite{ho2020denoising}, offering benefits such as distribution coverage, a stationary training objective, and easy scalability~\cite{dhariwal2021diffusion}. 

However, despite achieving high-quality sample generation, diffusion models incur significant computational overhead. 
Latent diffusion models (LDMs) apply the diffusion process in a latent space, significantly reducing computational costs. 
For example, Stable Video Diffusion (SVD)~\cite{blattmann2023stable} and Animatediff~\cite{guo2023animatediff} are classic text-to-video generation methods employing LDMs. 
Several studies~\cite{fang2023dance,chang2023magicdance} on human motion video generation are based on this method, 
extending the text-driven models to multi-condition driven ones.

\vspace{-10px}
\subsection{Human Data Representations}
\begin{figure}[!t]
    \centering
    \includegraphics[width=\linewidth]{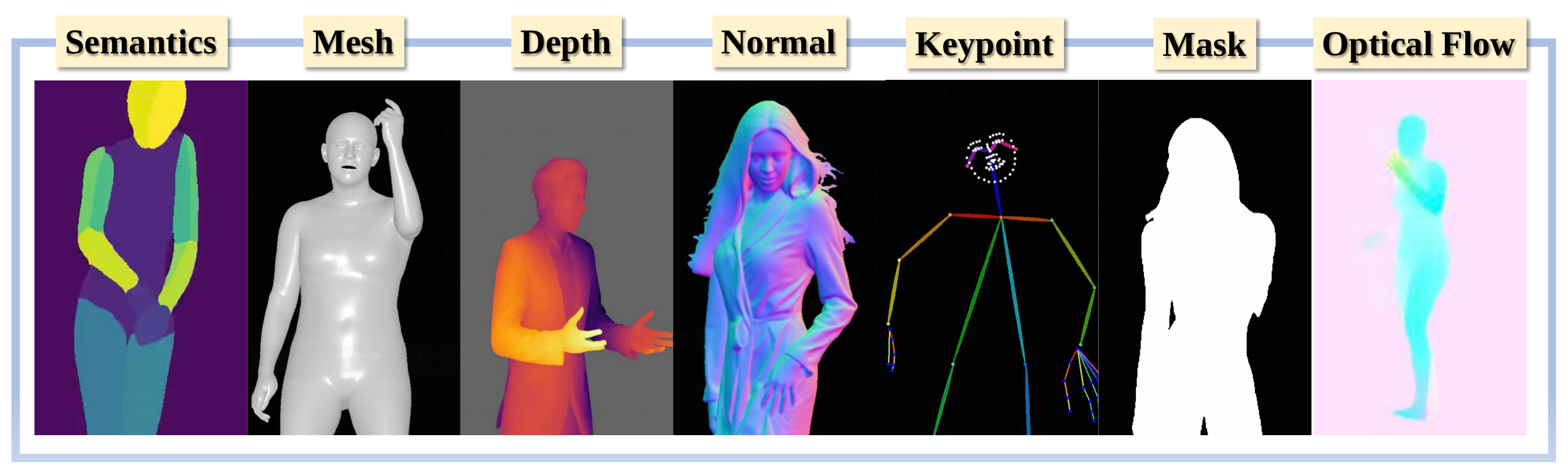}
    \caption{Different human data representations.}
    \label{fig:humanpose}
    \vspace{-10px}
\end{figure}

Following Knap \cite{knap2024human}, we categorize human body posture representations into seven key types
as shown in Fig. \ref{fig:humanpose}.


\noindent \textbf{Mask.}
Mask outlines the basic contours and occupied areas of characters, providing a coarse-grained layout prior \cite{fang2023dance}. However, they lack detailed posture descriptions.


\noindent \textbf{Mesh.}
Mesh offers a more detailed representation of body shape, including limb bending and occlusion.
Currently, mesh-based representation is applied to overcome the loss of 3D spatial information in keypoint estimation \cite{zhu2024champ}.


\noindent \textbf{Depth.}
Insufficient understanding of spatial relations limits existing methods' ability to accurately generate occluded body parts~\cite{xue2024follow}. Incorporating depth cues can effectively leverage spatial information, aiding the model in learning the spatial relationships between characters. 

\noindent \textbf{Normal.}
The normal condition critically emphasizes the orientation of the human body, ensuring precise alignment in the generation of realistic human poses for enhanced visual fidelity~\cite{zhu2024champ}. 
This alignment is paramount for maintaining the integrity of spatial relationships in animations, thereby significantly improving the believability and immersive quality of the rendered scenes.

\noindent \textbf{Keypoint.}
Existing pose keypoint estimation methods typically employ models such as OpenPose~\cite{cao2021openpose} or DWPose~\cite{yang2023dwpose}. 
Recent studies~\cite{hu2024animate, wang2024vividpose} use DWPose~\cite{yang2023dwpose} as the foundational keypoint-based human skeleton representation.

\noindent \textbf{Semantics.}
The semantic condition prioritizes the semantic information of different body parts, facilitating the decoupling of distinct features. 
This approach enables targeted enhancements and independent manipulation of each body part, thereby refining the representation and interaction dynamics within human-centric models \cite{xu2024magicanimate}. 


\noindent \textbf{Optical Flow.}
Directly optimizing the model on a noisy dataset often leads to background instability.
Follow Your Pose2~\cite{xue2024follow} reduces the impact of noisy data by adding additional optical flow conditions to enable the model to adapt to unstable changes in the background.

\section{Human Motion Planning}
\label{Sec:MP}

The human motion planning phase is critical in determining the specific motions that a virtual digital human performs. 
The motion sequence generated from input signals enables the virtual character to exhibit highly natural movements, align with human habits, and interact smoothly with surrounding objects, effectively mimicking human behavior.

Currently, human motion planning is primarily driven by two methodologies: one that leverages the power of LLMs for motion planning, and another that relies on the mapping of distinct features for motion generation. 
This section highlights the growing significance of LLMs in human motion video generation. 
By leveraging the inherent prior knowledge embedded within LLMs, these models can better comprehend semantic nuances and reason about emotions.


%
\vspace{-10px}
\subsection{Motion Planning Using Large Language Models}

In the realm of human motion video generation, studies by Geng et al. \cite{geng2023affective}, AgentAvatar \cite{wang2023agentavatar}, and InstructAvatar \cite{wang2024instructavatar} represent the current advancements in applying LLMs to motion planning. 
Notably, Geng et al. \cite{geng2023affective} pioneer the analysis of dialogue characteristics between two individuals in conversations using LLMs, thereby inferring the appropriate expressions for the listener. 
The other two studies focus on the generation of single-person human motion videos.

\begin{figure}[!t]
    \centering
    \includegraphics[width=0.85\linewidth]{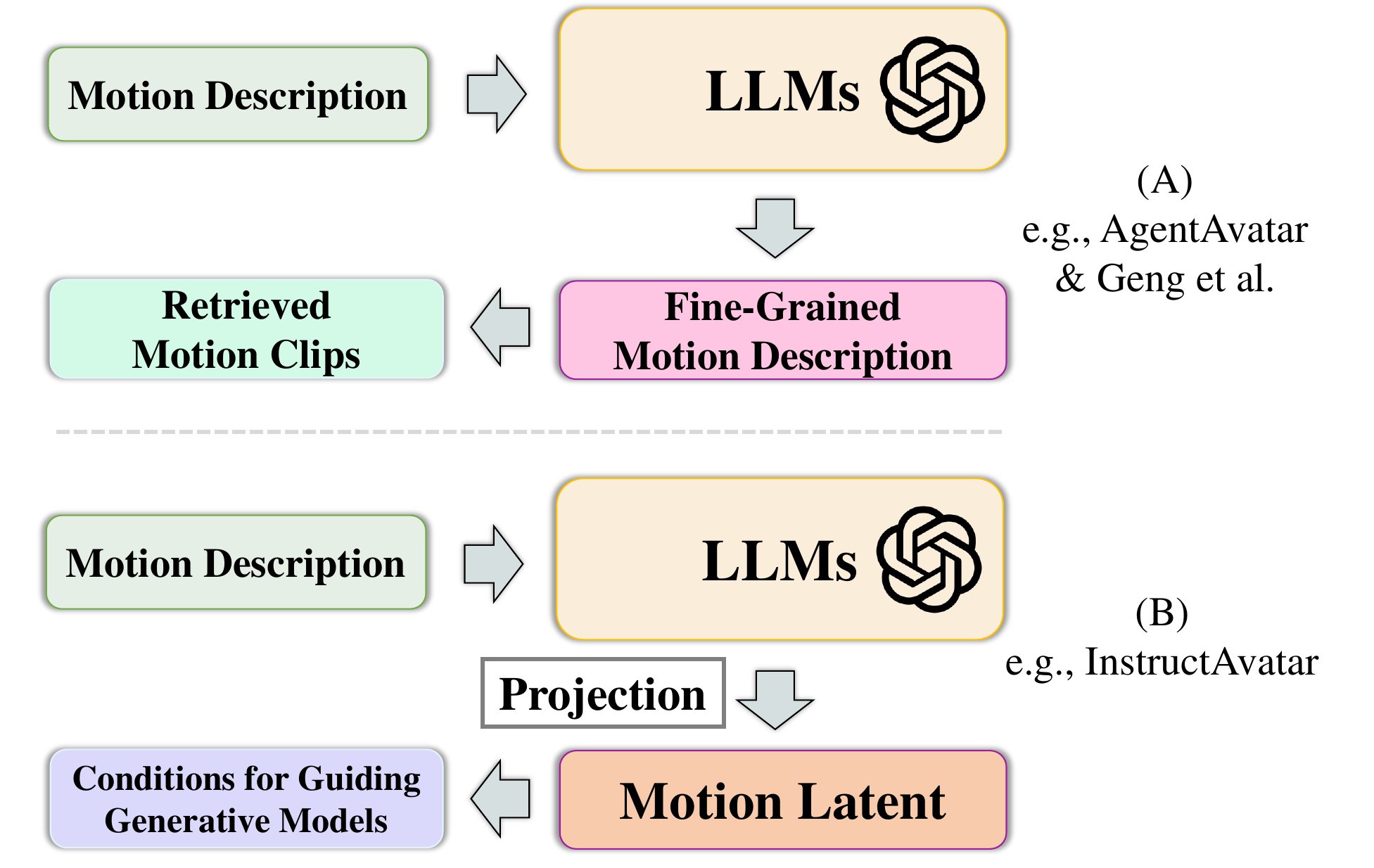}
    \caption{
    Two common forms of human motion planning by LLMs. (A) LLMs generate fine-grained descriptions for retrieval, and (B) LLMs project descriptions into a latent space for guiding generative models.
    }
    \begin{picture}(0,0)
    \put(99, 136){\scalebox{0.7}{\cite{wang2023agentavatar}}} 
    \put(93, 128){\scalebox{0.7}{\cite{geng2023affective}}} 
    \put(99, 68){\scalebox{0.7}{\cite{wang2024instructavatar}}} 
    \end{picture}
    \label{fig:LLMplanner}
    \vspace{-12px}
\end{figure}
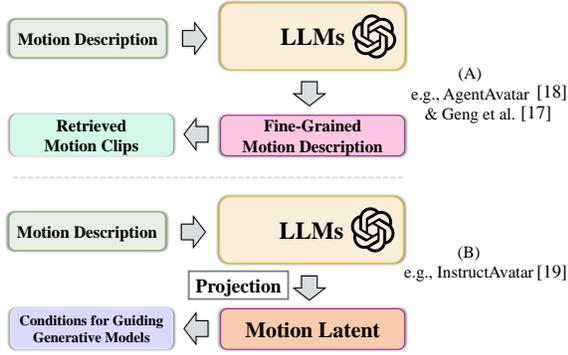

One of the pioneering efforts in this area, Geng et al. \cite{geng2023affective}, demonstrates the innovative use of LLMs in human motion video generation. 
Their approach starts by inputting the speaker’s scripts and conversational intention into LLMs, which then generate plausible reactions for the listener, such as a subtle smile. 
These generated motion descriptions are subsequently used to train a CLIP module \cite{radford2021learning}, effectively integrating the LLM\new{s} with motion generation.
AgentAvatar \cite{wang2023agentavatar} takes a broader approach by incorporating LLMs into the general context of human motion video generation. 
This method extends beyond dialogue-driven scenarios, enabling the planning and generation of human movements across a wide range of contexts. 
Their process begins with an environmental overview and avatar settings for the LLM-based planner. 
The planner generates detailed descriptions of facial movements, which are then passed to the driving engine to produce photorealistic video sequences. 
\begin{figure}[t]
    \centering
    \includegraphics[width=0.74\linewidth]{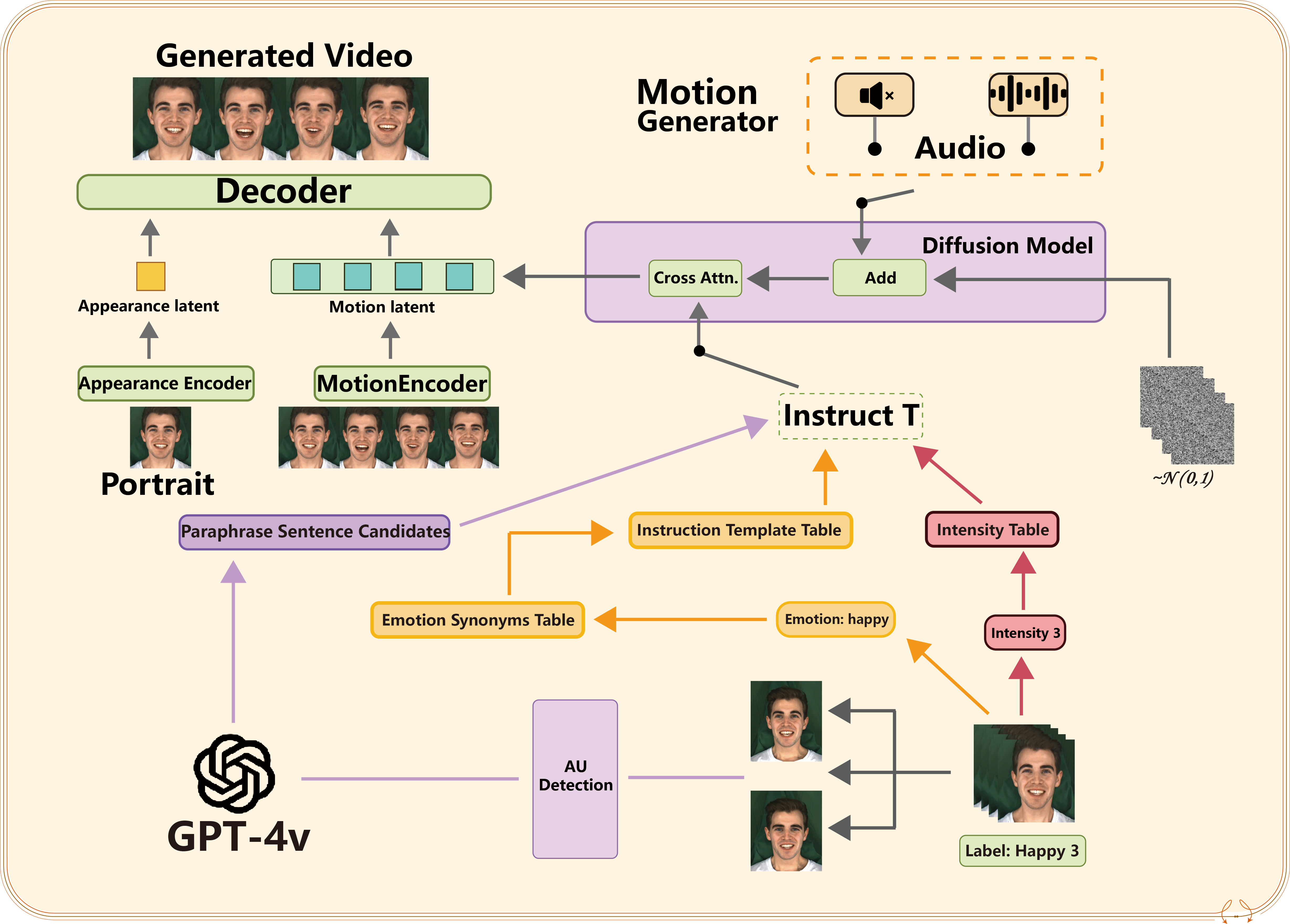}
    \vspace{-10px}
    \caption{Overview of InstructAvatar \cite{wang2024instructavatar}, which employs GPT-4 and diffusion models for video generation, producing expressive and dynamic videos that are synchronized with audio input.
    }
    \label{fig:LLMmethod3}
    \vspace{-12px}
\end{figure}
InstructAvatar \cite{wang2024instructavatar} designs an automatic annotation pipeline to construct a rich dataset of instruction-video pairs. 
This dataset captures fine-grained facial details using Action Units (AUs) to describe facial muscle movements. 
AUs are extracted using an off-the-shelf model and refined through multimodal LLMs, which paraphrase AUs into natural textual descriptions. 
This process not only enriches the dataset with detailed emotion and motion descriptions but also enhances the model’s ability to generalize across various expressions and emotions.
The architecture of InstructAvatar \cite{wang2024instructavatar}, as shown in Fig. \ref{fig:LLMmethod3}, 
is underpinned by VAE \cite{kingma2013auto} that disentangles motion from appearance, allowing for separate and focused manipulation of these elements. 
The motion generator, based on the Conformer architecture~\cite{conformer}, employs a diffusion model to learn the mapping from audio and text instructions to the motion latent space. 
This model is further equipped with a two-branch cross attention mechanism that distinguishes between emotion and motion instructions, ensuring that the avatar maintains a consistent emotional state throughout the video while executing dynamic facial motions as directed by the text.

These studies \cite{geng2023affective,wang2023agentavatar,wang2024instructavatar} reveal two distinct motion planning strategies utilized by LLMs, as depicted in Fig. \ref{fig:LLMplanner}. In Fig. \ref{fig:LLMplanner}~(A), LLMs process motion descriptions to generate fine-grained motion descriptions, which are used to perform retrieval on a video database to obtain relevant clips. This retrieval-based approach is adopted by methods such as Geng et al. \cite{geng2023affective} and AgentAvatar \cite{wang2023agentavatar}. In contrast, Fig. \ref{fig:LLMplanner} (B) illustrates a generative strategy where the derived motion conditions serve as inputs to a generative model, as demonstrated by InstructAvatar \cite{wang2024instructavatar}.


\begin{table*}[ht]
\vspace{-15px}
\renewcommand{\arraystretch}{0.8} 
\centering
\caption{Detailed review of recent developments in portrait animation. This table showcases a spectrum of innovative approaches focusing on facial animations through advanced generative models. 
ED: Encoder-Decoder; 
SIG: SIGGRAPH; SIGA: SIGGRAPH ASIA; 3D-P: 3D Parameterization.
}
\label{tab:portraitanimation}
\resizebox{0.86\textwidth}{!}{%
\begin{tabular}{|ccccccccc|}
\hline

\multicolumn{1}{|c|}{\textbf{\begin{tabular}[c]{@{}c@{}}Driving \\ Region\end{tabular}}}                       & \multicolumn{1}{c|}{\textbf{\begin{tabular}[c]{@{}c@{}}Publication \\ Time\end{tabular}}} & \multicolumn{1}{c|}{\textbf{Paper}}                                                  & \multicolumn{1}{c|}{\textbf{Input Signals}}                         & \multicolumn{1}{c|}{\textbf{\begin{tabular}[c]{@{}c@{}}Motion \\ Representation\end{tabular}}}              & \multicolumn{1}{c|}{\textbf{Backbone}}                                                                  & \multicolumn{1}{c|}{\textbf{\begin{tabular}[c]{@{}c@{}}Open \\ Source\end{tabular}}} & \textbf{\begin{tabular}[c]{@{}c@{}}\new{Real-Time} \\ \new{Inference}\end{tabular}} & \multicolumn{1}{|c|}{\textbf{Venue}}                                                                                                  \\ \hline

\multicolumn{9}{|c|}{\cellcolor[HTML]{D1DADC}\rule{0pt}{2.3ex}\textbf{\begin{tabular}[c]{@{}c@{}} Portrait Animation  \\\end{tabular}}}                                                                                                                                                                                                                                                                                                                                                            \\ \hline
\multicolumn{1}{|c|}{}                                                                                         & \multicolumn{1}{c|}{\cellcolor[HTML]{ECF2F0}Jun. 4, 2024}                                   & \multicolumn{1}{c|}{\cellcolor[HTML]{ECF2F0}Follow-Your-Emoji \cite{ma2024follow}}         & \multicolumn{1}{c|}{\cellcolor[HTML]{ECF2F0}\reallogo + \poselogo} & \multicolumn{1}{c|}{\cellcolor[HTML]{ECF2F0}KeyPoint}                                                       & \multicolumn{1}{c|}{\cellcolor[HTML]{ECF2F0}\begin{tabular}[c]{@{}c@{}}DM\end{tabular}} & \multicolumn{1}{c|}{\cellcolor[HTML]{ECF2F0}×} & \cellcolor[HTML]{ECF2F0}\new{×} & \multicolumn{1}{|c|}{\cellcolor[HTML]{ECF2F0}\begin{tabular}[c]{@{}c@{}}SIGA'24\end{tabular}}                            \\  
\multicolumn{1}{|c|}{}                                                                                         & \multicolumn{1}{c|}{Jul. 5, 2024}                                                           & \multicolumn{1}{c|}{LivePortrait \cite{guo2024liveportrait}}                               & \multicolumn{1}{c|}{\reallogo + \poselogo}                         & \multicolumn{1}{c|}{KeyPoint}                                                                               & \multicolumn{1}{c|}{\begin{tabular}[c]{@{}c@{}}ED\end{tabular}}                          & \multicolumn{1}{c|}{\begin{tabular}[c]{@{}c@{}}\checkmark\end{tabular}} & {\realtimelogo \new{40 FPS}} & \multicolumn{1}{|c|}{arXiv}                                                                                                           \\  
\multicolumn{1}{|c|}{}                                                                                         & \multicolumn{1}{c|}{\cellcolor[HTML]{ECF2F0}Jul. 9, 2024}                                   & \multicolumn{1}{c|}{\cellcolor[HTML]{ECF2F0}MobilePortrait \cite{jiang2024mobileportrait}} & \multicolumn{1}{c|}{\cellcolor[HTML]{ECF2F0}\reallogo + \poselogo} & \multicolumn{1}{c|}{\cellcolor[HTML]{ECF2F0}KeyPoint}                                                       & \multicolumn{1}{c|}{\cellcolor[HTML]{ECF2F0}\begin{tabular}[c]{@{}c@{}}DM\end{tabular}} & \multicolumn{1}{c|}{\cellcolor[HTML]{ECF2F0}×} & {\cellcolor[HTML]{ECF2F0}\realtimelogo \new{99 FPS}} & \multicolumn{1}{|c|}{\cellcolor[HTML]{ECF2F0}arXiv}                                                                                   \\  
\multicolumn{1}{|c|}{}                                                                                         & \multicolumn{1}{c|}{Oct. 16, 2023}                                                           & \multicolumn{1}{c|}{EDTN \cite{kang2024expression}}                                        & \multicolumn{1}{c|}{\reallogo + \poselogo}                         & \multicolumn{1}{c|}{\begin{tabular}[c]{@{}c@{}}3D-P\end{tabular}}                         & \multicolumn{1}{c|}{\begin{tabular}[c]{@{}c@{}}ED\end{tabular}}                          & \multicolumn{1}{c|}{×} & \new{×} & \multicolumn{1}{|c|}{\begin{tabular}[c]{@{}c@{}}ICASSP'24\end{tabular}}                                                          \\  
\multicolumn{1}{|c|}{}                                                                                         & \multicolumn{1}{c|}{\cellcolor[HTML]{ECF2F0}Mar. 26, 2023}                                   & \multicolumn{1}{c|}{\cellcolor[HTML]{ECF2F0}OTAvatar \cite{ma2023otavatar}}                & \multicolumn{1}{c|}{\cellcolor[HTML]{ECF2F0}\reallogo + \poselogo} & \multicolumn{1}{c|}{\cellcolor[HTML]{ECF2F0}\begin{tabular}[c]{@{}c@{}}3D-P\end{tabular}} & \multicolumn{1}{c|}{\cellcolor[HTML]{ECF2F0}\begin{tabular}[c]{@{}c@{}}ED\end{tabular}}  & \multicolumn{1}{c|}{\cellcolor[HTML]{ECF2F0}\checkmark} & {\cellcolor[HTML]{ECF2F0}\realtimelogo \new{35 FPS}} & \multicolumn{1}{|c|}{\cellcolor[HTML]{ECF2F0}\begin{tabular}[c]{@{}c@{}}CVPR'23\end{tabular}}                                    \\  
\multicolumn{1}{|c|}{}                                                                                         & \multicolumn{1}{c|}{Mar. 27, 2023}                                                           & \multicolumn{1}{c|}{OmniAvatar \cite{xu2023omniavatar}}                                    & \multicolumn{1}{c|}{\reallogo + \poselogo}                         & \multicolumn{1}{c|}{Latent}                                                                                 & \multicolumn{1}{c|}{GAN}                                                                                & \multicolumn{1}{c|}{×} & \new{×} & \multicolumn{1}{|c|}{\begin{tabular}[c]{@{}c@{}}CVPR'23\end{tabular}}                                                            \\  
\multicolumn{1}{|c|}{}                                                                                         & \multicolumn{1}{c|}{\cellcolor[HTML]{ECF2F0}Dec. 4, 2023}                                   & \multicolumn{1}{c|}{\cellcolor[HTML]{ECF2F0}GazeGANV2   \cite{zhang2022unsupervised}}      & \multicolumn{1}{c|}{\cellcolor[HTML]{ECF2F0}\reallogo + \poselogo} & \multicolumn{1}{c|}{\cellcolor[HTML]{ECF2F0}Latent}                                                         & \multicolumn{1}{c|}{\cellcolor[HTML]{ECF2F0}GAN}                                                        & \multicolumn{1}{c|}{\cellcolor[HTML]{ECF2F0}\checkmark} & \cellcolor[HTML]{ECF2F0}\new{×} & \multicolumn{1}{|c|}{\cellcolor[HTML]{ECF2F0}\begin{tabular}[c]{@{}c@{}}TIP'22\end{tabular}}  \\  

\multicolumn{1}{|c|}{\multirow{-6}{*}{\begin{tabular}[c]{@{}c@{}}\partlogo \\ Part \\ (Face)\end{tabular}}}                                                                                         & \multicolumn{1}{c|}{Jun. 8, 2024}                                                           & \multicolumn{1}{c|}{MegActor \cite{yang2024megactor}}                                      & \multicolumn{1}{c|}{\reallogo + \poselogo}                         & \multicolumn{1}{c|}{Latent}                                                                                 & \multicolumn{1}{c|}{\begin{tabular}[c]{@{}c@{}}DM\end{tabular}}                         & \multicolumn{1}{c|}{×} & \new{×} & \multicolumn{1}{|c|}{arXiv}                                                                                                           \\  
\multicolumn{1}{|c|}{}                                                                                         & \multicolumn{1}{c|}{\cellcolor[HTML]{ECF2F0}May. 31, 2024}                                   & \multicolumn{1}{c|}{\cellcolor[HTML]{ECF2F0}X-Portrait \cite{xie2024x}}                    & \multicolumn{1}{c|}{\cellcolor[HTML]{ECF2F0}\reallogo + \poselogo} & \multicolumn{1}{c|}{\cellcolor[HTML]{ECF2F0}Latent}                                                         & \multicolumn{1}{c|}{\cellcolor[HTML]{ECF2F0}\begin{tabular}[c]{@{}c@{}}DM\end{tabular}} & \multicolumn{1}{c|}{\cellcolor[HTML]{ECF2F0}×} & \cellcolor[HTML]{ECF2F0}\new{×} & \multicolumn{1}{|c|}{\cellcolor[HTML]{ECF2F0}\begin{tabular}[c]{@{}c@{}}SIG'24\end{tabular}}                                \\  
\multicolumn{1}{|c|}{} & \multicolumn{1}{c|}{Mar. 23, 2024}                                                           & \multicolumn{1}{c|}{FaceOff \cite{agarwal2023faceoff}}                                     & \multicolumn{1}{c|}{\reallogo + \poselogo}                         & \multicolumn{1}{c|}{Latent}                                                                                 & \multicolumn{1}{c|}{\begin{tabular}[c]{@{}c@{}}ED\end{tabular}}                          & \multicolumn{1}{c|}{×} & \new{×} & \multicolumn{1}{|c|}{\begin{tabular}[c]{@{}c@{}}WACV'23\end{tabular}}                                                            \\ \hline
\multicolumn{9}{|l|}{\reallogo: Reference Real Images; \poselogo: Driving Pose Video; \realtimelogo\new{: Real-Time Support 
(Estimated FPS based on NVIDIA RTX 4090 GPU).
}
}                                                                                                                                                                                                                                                                                                                                                                                                                                                                                                                                                                                                                                                                                                                    \\ \hline
\end{tabular}%
}
\vspace{-10px}
\end{table*}

LLMs are beginning to show promising results in 3D skeletal motion generation, acting as central orchestrators in the motion planning phase. 
For instance, FineMoGen \cite{zhang2024finemogen}, PRO-Motion~\cite{liu2023plan}, AvatarGPT \cite{zhou2024avatargpt}, and MotionGPT \cite{zhang2024motiongpt} are capable of generating holistic human motions from text.
MotionScript~\cite{yazdian2023motionscript} facilitates the seamless transformation of motions into descriptive motion scripts and vice versa, establishing a robust bidirectional generation mechanism.
Additionally, Ng et al. \cite{ng2023can} introduce an innovative approach that leverages LLMs to synthesize 3D facial expressions tailored to audience characteristics. 
Furthermore, InterControl \cite{wang2023intercontrol} and MoMat-MoGen \cite{cai2024digital} are beginning to explore the potential of LLMs in facilitating dyadic interaction tasks. 
However, these efforts are limited to 3D skeletal motion and do not extend to the creation of authentic video content so far.  The field of human motion video generation remains ripe for further advancement.

Currently, the generation of human motion videos predominantly relies on text as an intermediary to integrate LLMs with generative models.
All these works are summarized in Appendix A.
However, there remains a lack of exploration into more effective and novel intermediate representations.
From the initial analysis, different challenges emerge:

\begin{itemize}

\item  \new{As indicated in Table~\ref{tab:taskdescription}, most existing works~\cite{guo2024liveportrait, liu2023human, zhang2024mimicmotion, ma2024followyp, tian2024emo} primarily focus on learning implicit relationships between driving conditions and motion sequences. LLMs remain underutilized as motion planners, despite their potential to decompose driving conditions into fine-grained features for enhancing motion planning.}


\item \new{Ng et al.~\cite{ng2023can} pioneer the exploration of intermediate representations through codebooks beyond textual descriptions, although codebook-based controls show limited granularity in motion manipulation. Investigating effective intermediate representations for motion planning remains a crucial research direction to enable LLMs to understand and plan behaviors.}


\item \new{While current evaluation methodologies rely on video quality metrics and user studies~\cite{wang2024instructavatar, zhou2024avatargpt, wang2023agentavatar}, the assessment of LLMs' effectiveness in motion planning remains a fundamental challenge.}

\end{itemize}


\vspace{-12px}
\subsection{Motion Planning through Mapping Features}
In human motion video generation, the complexity of video synthesis has led most studies \cite{zhong2024stylepreservinglipsyncaudioaware, du2023dae, xu2023high} to adopt implicit representations of input features. 
These approaches focus on learning the mapping between input conditions and motion, often generating diverse outcomes by incorporating a degree of stochastic noise. 
A detailed discussion of these methods will be provided in Section \ref{Sec:MV}.

\vspace{-8px}

\section{Motion Modeling and Video Generation}
\label{Sec:MV}

In the human motion video generation phase, our objective is to synthesize realistic videos that capture holistic human motion, grounded in the outcomes from the motion planning phase. To thoroughly examine the various generation approaches, we conduct comprehensive analyses on each specific sub-task. This section concentrates on video generation methods leveraging diffusion models,  excluding other techniques like NeRF~\cite{mildenhall2021nerf} and 3DGS \cite{kerbl20233d}. We delve into three key areas: \textit{vision guidance} (including portrait animation, dance video generation, \new{Try-On}, and pose-to-video), \textit{text guidance} (covering Text2Face and Text2MotionVideo), and \textit{audio-driven scenarios} (such as talking head generation and audio-driven holistic human motion generation).

\vspace{-10px}
\subsection{Vision-Driven Human Motion Video Generation}

\noindent \textbf{Portrait Animation. }
Portrait animation focuses on breathing life into static images, typically portraits, using advanced animation techniques. The process begins with a static image, whether it is a photograph or a digital painting, and transforms it into an animated sequence that conveys the subject’s emotional expressions.
Recent advancements in this field are summarized in Table~\ref{tab:portraitanimation}.

OmniAvatar~\cite{xu2023omniavatar} exemplifies the use of geometric priors to guide the animation process, ensuring 3D consistency and detailed facial expressions. 
Follow-Your-Emoji~\cite{ma2024follow} utilizes a diffusion-based framework for animating portraits with target landmark sequences. 
By using an expression-aware landmark to guide the animation, it ensures motion alignment and identity preservation while enhancing the portrayal of exaggerated expressions. 
Additionally, it employs a fine-grained facial loss function to improve the model's perception of subtle expressions and the reconstruction of the reference portrait's appearance.
LivePortrait~\cite{guo2024liveportrait} introduces an efficient video-driven framework that balances computational efficiency with controllability, enabling rapid generation speeds.

Overall, portrait animation, which requires only a reference image to drive facial expressions based on input conditions, draws significant research focus. 
For instance, EDTN~\cite{kang2024expression} and X-Portrait~\cite{xie2024x} address the challenge of cross-domain head reenactment, allowing human motions to be transferred to non-human domains, such as anime characters. 
MobilePortrait~\cite{jiang2024mobileportrait} focuses on real-time performance, offering one-shot solutions for talking face avatars with controllable rendering. 
FaceOff~\cite{agarwal2023faceoff} presents a novel video-to-video face-swapping system that retains source expressions and identity while adapting to the pose and background of the target video.
From numerous studies, we can identify different key challenges in portrait animation:

\begin{itemize}

\item \new{Current facial driving techniques~\cite{xu2023omniavatar,ma2024follow,xie2024x,agarwal2023faceoff,yang2024megactor} encompass multiple facial attributes, including eye gaze, teeth, lip synchronization, and head posture. However, precise control over eye gaze consistency and teeth details remains insufficiently addressed in existing research.}

\item \new{Existing approaches typically require identity-specific fine-tuning to achieve optimal performance~\cite{ma2024follow,kang2024expression,xu2023omniavatar}, while maintaining identity consistency in zero-shot scenarios presents a significant challenge.}

\item \new{Most current studies~\cite{jiang2024mobileportrait,guo2024liveportrait,xu2023omniavatar,ma2024follow} focus on single-person facial driving, making the exploration of methods for multi-person facial driving a promising research.}


\end{itemize}






\noindent \textbf{Video-Driven Dance Video Generation.}
Video-driven dance video generation enables the transformation of a static individual into a dynamic dancer from a video source. 
This technology allows for the transfer of dance movements from a professional dancer to an amateur or non-dancer, enabling them to perform complex choreographies with lifelike fluidity.


Everybody Dance Now \cite{chan2019everybody} introduces a method for motion transfer using poses as an intermediate representation, enabling the transfer of dance performances onto subjects with different physical attributes.
\begin{table*}[ht]
\vspace{-15px}
\renewcommand{\arraystretch}{0.8} 
\centering
\caption{Comprehensive overview of research on video-driven and pose-driven dance video generation. DiT: Diffusion Transformer.}
\label{tab:dance}
\resizebox{0.86\textwidth}{!}{%
\begin{tabular}{|ccccccccc|}
\hline

\multicolumn{1}{|c|}{\textbf{\begin{tabular}[c]{@{}c@{}}Driving \\ Region\end{tabular}}}                  & \multicolumn{1}{c|}{\textbf{\begin{tabular}[c]{@{}c@{}}Publication \\ Time\end{tabular}}} & \multicolumn{1}{c|}{\textbf{Paper}}                                            & \multicolumn{1}{c|}{\textbf{Input Signals}}                                & \multicolumn{1}{c|}{\textbf{\begin{tabular}[c]{@{}c@{}}Motion \\ Representation\end{tabular}}}              & \multicolumn{1}{c|}{\textbf{Backbone}}                                                                  & \multicolumn{1}{c|}{\textbf{\begin{tabular}[c]{@{}c@{}}Open \\ Source\end{tabular}}} & \multicolumn{1}{c|}{\textbf{\begin{tabular}[c]{@{}c@{}}\new{Real-Time} \\ \new{Inference}\end{tabular}}} & \textbf{Venue}                                                                                     \\ \hline

\multicolumn{9}{|c|}{\cellcolor[HTML]{D1DADC}\textbf{\begin{tabular}[c]{@{}c@{}} Video-Driven Dance Video Generation \\\end{tabular}}}               \\ \hline                                                                                                                                                                                                                                                                  

\multicolumn{1}{|c|}{}                                                                                    & \multicolumn{1}{c|}{\cellcolor[HTML]{ECF2F0}Aug. 22, 2018}                                   & \multicolumn{1}{c|}{\cellcolor[HTML]{ECF2F0}Chen et al. \cite{chan2019everybody}}    & \multicolumn{1}{c|}{\cellcolor[HTML]{ECF2F0}\reallogo + \drivingreallogo} & \multicolumn{1}{c|}{\cellcolor[HTML]{ECF2F0}KeyPoint}                                                       & \multicolumn{1}{c|}{\cellcolor[HTML]{ECF2F0}GAN}                                                        & \multicolumn{1}{c|}{\cellcolor[HTML]{ECF2F0}\checkmark} & \multicolumn{1}{c|}{\cellcolor[HTML]{ECF2F0}\new{×}} & \cellcolor[HTML]{ECF2F0}\begin{tabular}[c]{@{}c@{}}ICCV'19\end{tabular}                       \\  
\multicolumn{1}{|c|}{\multirow{-2}{*}{\begin{tabular}[c]{@{}c@{}}\quad \\ \holisticlogo\\ Holistic \\ Human \end{tabular}}}                                                                                    & \multicolumn{1}{c|}{Jul. 02, 2023}                                                           & \multicolumn{1}{c|}{BTDM \cite{adiya2023bidirectional}}                              & \multicolumn{1}{c|}{\reallogo + \drivingreallogo}                         & \multicolumn{1}{c|}{Region}                                                                                 & \multicolumn{1}{c|}{\begin{tabular}[c]{@{}c@{}}DM\end{tabular}}                         & \multicolumn{1}{c|}{×} & \multicolumn{1}{c|}{\new{×}} & \new{ICLR'24}                                                                                              \\  
\multicolumn{1}{|c|}{}                                                                                    & \multicolumn{1}{c|}{\cellcolor[HTML]{ECF2F0}Feb. 22, 2023}                                   & \multicolumn{1}{c|}{\cellcolor[HTML]{ECF2F0}Human MotionFormer \cite{liu2023human}}  & \multicolumn{1}{c|}{\cellcolor[HTML]{ECF2F0}\reallogo + \drivingreallogo} & \multicolumn{1}{c|}{\cellcolor[HTML]{ECF2F0}KeyPoint}                                                       & \multicolumn{1}{c|}{\cellcolor[HTML]{ECF2F0}\begin{tabular}[c]{@{}c@{}}ED\end{tabular}}  & \multicolumn{1}{c|}{\cellcolor[HTML]{ECF2F0}×} & \multicolumn{1}{c|}{\cellcolor[HTML]{ECF2F0}\new{×}} & \cellcolor[HTML]{ECF2F0}\new{ICLR'23}                                                                      \\  
\multicolumn{1}{|c|}{}  & \multicolumn{1}{c|}{Jun. 24, 2024}                                                           & \multicolumn{1}{c|}{FakeVideo \cite{wu2024pose}}                                     & \multicolumn{1}{c|}{\reallogo + \drivingreallogo}                         & \multicolumn{1}{c|}{KeyPoint}                                                                               & \multicolumn{1}{c|}{GAN}                                                                                & \multicolumn{1}{c|}{×} & \multicolumn{1}{c|}{\new{×}} & TDSC'24\\ \hline
\multicolumn{9}{|c|}{\cellcolor[HTML]{D1DADC}\rule{0pt}{2.3ex}\textbf{\begin{tabular}[c]{@{}c@{}} Pose-Driven Dance Video Generation \\\end{tabular}}}                                                                                                                                                                                                                                                                                                                                                                                                                                                                                                                                                                                                                                            \\ \hline
\multicolumn{1}{|c|}{}                                                                                    & \multicolumn{1}{c|}{\cellcolor[HTML]{ECF2F0}Jun. 30, 2023}                                   & \multicolumn{1}{c|}{\cellcolor[HTML]{ECF2F0}DisCo \cite{wang2024disco}}              & \multicolumn{1}{c|}{\cellcolor[HTML]{ECF2F0}\reallogo + \poselogo}        & \multicolumn{1}{c|}{\cellcolor[HTML]{ECF2F0}KeyPoint}                                                       & \multicolumn{1}{c|}{\cellcolor[HTML]{ECF2F0}\begin{tabular}[c]{@{}c@{}}DM\end{tabular}} & \multicolumn{1}{c|}{\cellcolor[HTML]{ECF2F0}\checkmark} & \multicolumn{1}{c|}{\cellcolor[HTML]{ECF2F0}\new{×}} & \cellcolor[HTML]{ECF2F0}\begin{tabular}[c]{@{}c@{}}CVPR'24\end{tabular}                        \\  
\multicolumn{1}{|c|}{}                                                                                    & \multicolumn{1}{c|}{Oct. 20, 2023}                                                           & \multicolumn{1}{c|}{Dance-Your-Latents \cite{fang2023dance}}                         & \multicolumn{1}{c|}{\reallogo + \poselogo}                                & \multicolumn{1}{c|}{KeyPoint}                                                                               & \multicolumn{1}{c|}{\begin{tabular}[c]{@{}c@{}}DM\end{tabular}}                         & \multicolumn{1}{c|}{×} & \multicolumn{1}{c|}{\new{×}} & arXiv                                                                                              \\  
\multicolumn{1}{|c|}{}                                                                                    & \multicolumn{1}{c|}{\cellcolor[HTML]{ECF2F0}Nov. 18, 2023}                                   & \multicolumn{1}{c|}{\cellcolor[HTML]{ECF2F0}MagicPose \cite{chang2024magicpose}}    & \multicolumn{1}{c|}{\cellcolor[HTML]{ECF2F0}\reallogo + \poselogo}        & \multicolumn{1}{c|}{\cellcolor[HTML]{ECF2F0}KeyPoint}                                                       & \multicolumn{1}{c|}{\cellcolor[HTML]{ECF2F0}\begin{tabular}[c]{@{}c@{}}DM\end{tabular}} & \multicolumn{1}{c|}{\cellcolor[HTML]{ECF2F0}\checkmark} & \multicolumn{1}{c|}{\cellcolor[HTML]{ECF2F0}\new{×}} & \cellcolor[HTML]{ECF2F0}\begin{tabular}[c]{@{}c@{}}ICML'24\end{tabular}                        \\  
\multicolumn{1}{|c|}{}                                                                                    & \multicolumn{1}{c|}{Nov. 27, 2023}                                                           & \multicolumn{1}{c|}{MagicAnimate \cite{xu2024magicanimate}}                          & \multicolumn{1}{c|}{\reallogo + \poselogo}                                & \multicolumn{1}{c|}{Region}                                                                                 & \multicolumn{1}{c|}{\begin{tabular}[c]{@{}c@{}}DM\end{tabular}}                         & \multicolumn{1}{c|}{\begin{tabular}[c]{@{}c@{}}\checkmark\end{tabular}} & \multicolumn{1}{c|}{\new{×}} & \begin{tabular}[c]{@{}c@{}}CVPR'24\end{tabular}                                                \\  
\multicolumn{1}{|c|}{}                                                                                    & \multicolumn{1}{c|}{\cellcolor[HTML]{ECF2F0}Nov. 28, 2023}                                   & \multicolumn{1}{c|}{\cellcolor[HTML]{ECF2F0}Animate Anyone \cite{hu2024animate}}     & \multicolumn{1}{c|}{\cellcolor[HTML]{ECF2F0}\reallogo + \poselogo}        & \multicolumn{1}{c|}{\cellcolor[HTML]{ECF2F0}KeyPoint}                                                       & \multicolumn{1}{c|}{\cellcolor[HTML]{ECF2F0}\begin{tabular}[c]{@{}c@{}}DM\end{tabular}} & \multicolumn{1}{c|}{\cellcolor[HTML]{ECF2F0}×} & \multicolumn{1}{c|}{\cellcolor[HTML]{ECF2F0}\new{×}} & \cellcolor[HTML]{ECF2F0}\begin{tabular}[c]{@{}c@{}}CVPR'24\end{tabular}                        \\  
\multicolumn{1}{|c|}{}                                                                                    & \multicolumn{1}{c|}{Dec. 8, 2023}                                                           & \multicolumn{1}{c|}{DreaMoving \cite{feng2023dreamoving}}                            & \multicolumn{1}{c|}{\reallogo + \poselogo + \textlogo}                    & \multicolumn{1}{c|}{KeyPoint}                                                                               & \multicolumn{1}{c|}{\begin{tabular}[c]{@{}c@{}}DM\end{tabular}}                         & \multicolumn{1}{c|}{\begin{tabular}[c]{@{}c@{}}\checkmark\end{tabular}} & \multicolumn{1}{c|}{\new{×}} & arXiv                                                                                              \\  
\multicolumn{1}{|c|}{}                                                                                    & \multicolumn{1}{c|}{\cellcolor[HTML]{ECF2F0}Dec. 27, 2023}                                   & \multicolumn{1}{c|}{\cellcolor[HTML]{ECF2F0}I2V-Adapter \cite{guo2023i2v}}           & \multicolumn{1}{c|}{\cellcolor[HTML]{ECF2F0}\reallogo + \poselogo}        & \multicolumn{1}{c|}{\cellcolor[HTML]{ECF2F0}KeyPoint}                                                       & \multicolumn{1}{c|}{\cellcolor[HTML]{ECF2F0}\begin{tabular}[c]{@{}c@{}}DM\end{tabular}} & \multicolumn{1}{c|}{\cellcolor[HTML]{ECF2F0}×} & \multicolumn{1}{c|}{\cellcolor[HTML]{ECF2F0}\new{×}} & \cellcolor[HTML]{ECF2F0}\begin{tabular}[c]{@{}c@{}}SIG'24\end{tabular}                    \\  
\multicolumn{1}{|c|}{}                                                                                    & \multicolumn{1}{c|}{May. 26, 2024}                                                           & \multicolumn{1}{c|}{Liu et. al \cite{liu2024disentangling}}                          & \multicolumn{1}{c|}{\reallogo + \poselogo}                                & \multicolumn{1}{c|}{KeyPoint}                                                                               & \multicolumn{1}{c|}{\begin{tabular}[c]{@{}c@{}}DM\end{tabular}}                         & \multicolumn{1}{c|}{×} & \multicolumn{1}{c|}{\new{×}} & arXiv                                                                                              \\  
\multicolumn{1}{|c|}{}                                                                                    & \multicolumn{1}{c|}{\cellcolor[HTML]{ECF2F0}May. 28, 2024}                                   & \multicolumn{1}{c|}{\cellcolor[HTML]{ECF2F0}VividPose \cite{wang2024vividpose}}      & \multicolumn{1}{c|}{\cellcolor[HTML]{ECF2F0}\reallogo + \poselogo}        & \multicolumn{1}{c|}{\cellcolor[HTML]{ECF2F0}\begin{tabular}[c]{@{}c@{}}3D-P\end{tabular}}                                                       & \multicolumn{1}{c|}{\cellcolor[HTML]{ECF2F0}\begin{tabular}[c]{@{}c@{}}DM\end{tabular}} & \multicolumn{1}{c|}{\cellcolor[HTML]{ECF2F0}×} & \multicolumn{1}{c|}{\cellcolor[HTML]{ECF2F0}\new{×}} & \cellcolor[HTML]{ECF2F0}arXiv                                                                      \\  
\multicolumn{1}{|c|}{}                                                                                    & \multicolumn{1}{c|}{May. 30, 2024}                                                           & \multicolumn{1}{c|}{MotionFollower \cite{tu2024motionfollower}}                      & \multicolumn{1}{c|}{\reallogo + \poselogo}                                & \multicolumn{1}{c|}{KeyPoint}                                                                               & \multicolumn{1}{c|}{\begin{tabular}[c]{@{}c@{}}DM\end{tabular}}                         & \multicolumn{1}{c|}{×} & \multicolumn{1}{c|}{\new{×}} & arXiv                                                                                              \\  
\multicolumn{1}{|c|}{}                                                                                    & \multicolumn{1}{c|}{\cellcolor[HTML]{ECF2F0}Jun. 3, 2024}                                   & \multicolumn{1}{c|}{\cellcolor[HTML]{ECF2F0}UniAnimate \cite{wang2024unianimate}}    & \multicolumn{1}{c|}{\cellcolor[HTML]{ECF2F0}\reallogo + \poselogo}        & \multicolumn{1}{c|}{\cellcolor[HTML]{ECF2F0}KeyPoint}                                                       & \multicolumn{1}{c|}{\cellcolor[HTML]{ECF2F0}\begin{tabular}[c]{@{}c@{}}DM\end{tabular}} & \multicolumn{1}{c|}{\cellcolor[HTML]{ECF2F0}×} & \multicolumn{1}{c|}{\cellcolor[HTML]{ECF2F0}\new{×}} & \cellcolor[HTML]{ECF2F0}arXiv                                                                      \\  
\multicolumn{1}{|c|}{}                                                                                    & \multicolumn{1}{c|}{Jun. 5, 2024}                                                           & \multicolumn{1}{c|}{Follow-Your-Pose v2 \cite{xue2024follow}}                        & \multicolumn{1}{c|}{\reallogo + \poselogo}                                & \multicolumn{1}{c|}{KeyPoint}                                                                               & \multicolumn{1}{c|}{\begin{tabular}[c]{@{}c@{}}DM\end{tabular}}                         & \multicolumn{1}{c|}{×} & \multicolumn{1}{c|}{\new{×}} & arXiv                                                                                              \\  
\multicolumn{1}{|c|}{}                                                                                    & \multicolumn{1}{c|}{\cellcolor[HTML]{ECF2F0}May. 27, 2024}                                   & \multicolumn{1}{c|}{\cellcolor[HTML]{ECF2F0}Human4DiT \cite{shao2024human4dit}}      & \multicolumn{1}{c|}{\cellcolor[HTML]{ECF2F0}\reallogo + \poselogo}        & \multicolumn{1}{c|}{\cellcolor[HTML]{ECF2F0}\begin{tabular}[c]{@{}c@{}}3D-P\end{tabular}} &     
\multicolumn{1}{c|}{\cellcolor[HTML]{ECF2F0}\begin{tabular}[c]{@{}c@{}}DiT\end{tabular}} & \multicolumn{1}{c|}{\cellcolor[HTML]{ECF2F0}\checkmark} & \multicolumn{1}{c|}{\cellcolor[HTML]{ECF2F0}\new{×}} & \cellcolor[HTML]{ECF2F0}arXiv                                                                                   \\  
\multicolumn{1}{|c|}{\multirow{-12}{*}{\begin{tabular}[c]{@{}c@{}}\holisticlogo\\ Holistic \\ Human \end{tabular}}}                                                                                    & \multicolumn{1}{c|}{Jan. 19, 2024}                                                           & \multicolumn{1}{c|}{3DHM \cite{li2024synthesizing}}                                  & \multicolumn{1}{c|}{\reallogo + \poselogo}                                & \multicolumn{1}{c|}{\begin{tabular}[c]{@{}c@{}}3D-P\end{tabular}}                         & \multicolumn{1}{c|}{\begin{tabular}[c]{@{}c@{}}DM\end{tabular}}                         & \multicolumn{1}{c|}{×} & \multicolumn{1}{c|}{\new{×}} & arXiv                                                                                              \\  
\multicolumn{1}{|c|}{}                                                                                    & \multicolumn{1}{c|}{\cellcolor[HTML]{ECF2F0}Mar. 21, 2024}                                   & \multicolumn{1}{c|}{\cellcolor[HTML]{ECF2F0}Champ \cite{zhu2024champ}}               & \multicolumn{1}{c|}{\cellcolor[HTML]{ECF2F0}\reallogo + \poselogo}        & \multicolumn{1}{c|}{\cellcolor[HTML]{ECF2F0}\begin{tabular}[c]{@{}c@{}}3D-P\end{tabular}} & \multicolumn{1}{c|}{\cellcolor[HTML]{ECF2F0}\begin{tabular}[c]{@{}c@{}}DM\end{tabular}} & \multicolumn{1}{c|}{\cellcolor[HTML]{ECF2F0}\checkmark} & \multicolumn{1}{c|}{\cellcolor[HTML]{ECF2F0}\new{×}} & \cellcolor[HTML]{ECF2F0}\begin{tabular}[c]{@{}c@{}}ECCV'24\end{tabular}                                                                      \\  
\multicolumn{1}{|c|}{}                                                                                    & \multicolumn{1}{c|}{Jul. 15, 2024}                                                           & \multicolumn{1}{c|}{TCAN \cite{kim2024tcan}}                                         & \multicolumn{1}{c|}{\reallogo + \poselogo}                                & \multicolumn{1}{c|}{KeyPoint}                                                                               & \multicolumn{1}{c|}{\begin{tabular}[c]{@{}c@{}}DM\end{tabular}}                         & \multicolumn{1}{c|}{×} & \multicolumn{1}{c|}{\new{×}} & \new{ECCV'24}                                                                                              \\  
\multicolumn{1}{|c|}{}                                                                                    & \multicolumn{1}{c|}{\cellcolor[HTML]{ECF2F0}Jul. 1, 2024}                                   & \multicolumn{1}{c|}{\cellcolor[HTML]{ECF2F0}MimicMotion \cite{zhang2024mimicmotion}} & \multicolumn{1}{c|}{\cellcolor[HTML]{ECF2F0}\reallogo + \poselogo}        & \multicolumn{1}{c|}{\cellcolor[HTML]{ECF2F0}KeyPoint}                                                       & \multicolumn{1}{c|}{\cellcolor[HTML]{ECF2F0}\begin{tabular}[c]{@{}c@{}}DM\end{tabular}} & \multicolumn{1}{c|}{\cellcolor[HTML]{ECF2F0}×} & \multicolumn{1}{c|}{\cellcolor[HTML]{ECF2F0}\new{×}} & \cellcolor[HTML]{ECF2F0}arXiv                                                                      \\  

\multicolumn{1}{|c|}{} & \multicolumn{1}{c|}{Jul. 16, 2024}                                   & \multicolumn{1}{c|}{IDOL \cite{zhai2024idol}}                & \multicolumn{1}{c|}{\reallogo + \poselogo}        & \multicolumn{1}{c|}{Region}                                                         & \multicolumn{1}{c|}{\begin{tabular}[c]{@{}c@{}}DM\end{tabular}} & \multicolumn{1}{c|}{×} & \multicolumn{1}{c|}{\new{×}} & \new{ECCV'24}                                                                      \\ \hline
\multicolumn{9}{|l|}{\reallogo: Reference Real Images; \poselogo: Driving Pose Video; \drivingreallogo: Driving Real Video; \textlogo: Text Prompts. }                                                                                                                                                                                                                                                                                                                                                                                                                                                                                                                                                                                                                                    \\ \hline
\end{tabular}%
}
\vspace{-10px}
\end{table*}
BTDM \cite{adiya2023bidirectional} enforces temporal coherence by reducing motion ambiguity.
Human MotionFormer \cite{liu2023human}, a hierarchical Vision Transformer \cite{dosovitskiy2020image} framework, is designed for transferring human motions by capturing both global and local perceptions to accurately match large and subtle motions.
FakeVideo \cite{wu2024pose} employs perceptual loss and Gromov-Wasserstein loss \cite{peyre2016gromov} to bridge the gap between pose and appearance.
It also introduces an episodic memory module to support continuous learning and uses facial geometrical cues to enhance facial details.
The domain of human motion copy, especially in video-driven dance video generation, has key challenges that researchers and developers are addressing:

\begin{itemize}

\item \new{Current approaches~\cite{chan2019everybody,liu2023human,wu2024pose} apply DWPose~\cite{yang2023dwpose} to extract pose sequences from videos, but the resulting human representations have inherent accuracy limitations. Motion blur in high-velocity dance movements exacerbates these precision issues, highlighting the critical need for developing accurate human pose representations from videos.}


\item \new{Generating videos with realistic and temporally consistent human body textures remains challenging. Existing methods~\cite{chan2019everybody,wu2024pose,adiya2023bidirectional,liu2023human} require a mass of training data and struggle to maintain visual-temporal coherence, underscoring the need to explore few-shot learning approaches.}



\end{itemize}

\noindent \textbf{Pose-Driven Dance Video Generation. }
Pose-driven dance video generation focuses on synthesizing realistic and temporally coherent video frames that depict human images performing dance movements, where the recent works in this field are shown in Table~\ref{tab:dance}. 

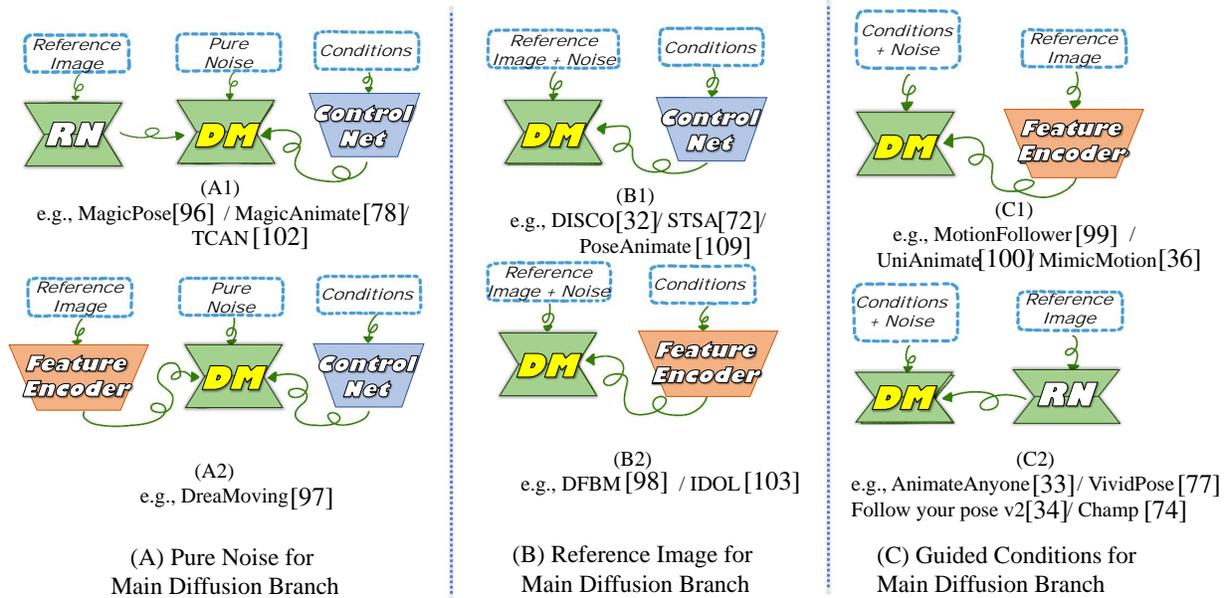
\begin{figure*}[t!]
    \centering
    \vspace{-15px}
    \includegraphics[width=0.9\linewidth]{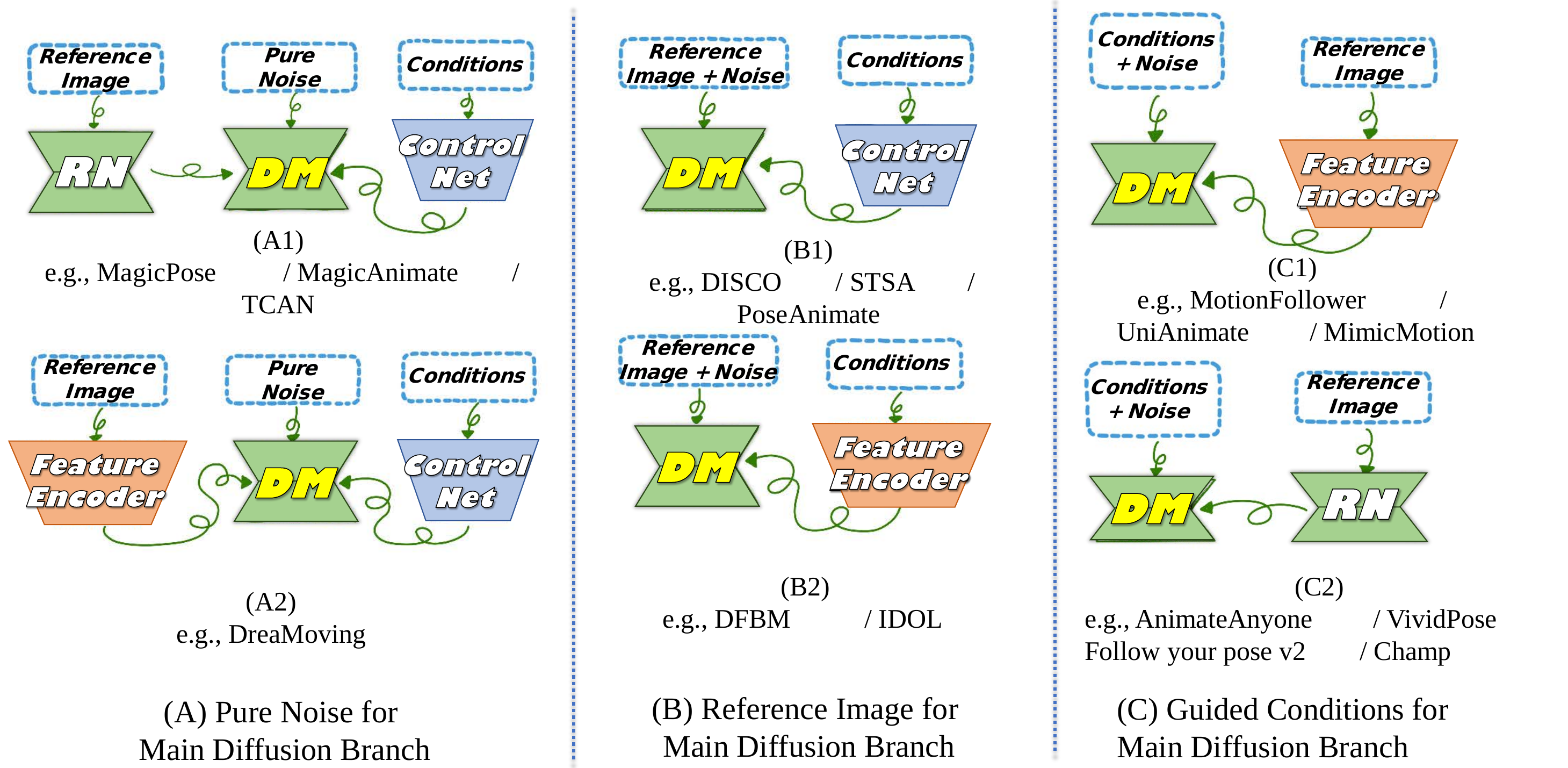}
    \caption{Comparative overview of different generative frameworks based on diffusion models, where pure noise (A),  a reference image (B), and guided conditions (C) are for the main diffusion branch. Each framework integrates unique components, such as ControlNet or feature encoders, to achieve diverse objectives in animation and pose generation objectives.
    The DM refers to the main diffusion model branch, while the ReferenceNet (RN) is a U-Net copy of the main diffusion model.
    }
    \begin{picture}(0,0)
\put(-170, 188.5){\cite{chang2024magicpose}} 
\put(-97.8, 188.5){\cite{xu2024magicanimate}} 
\put(-139.5, 179.5){\cite{kim2024tcan}} 
\put(-2.2, 185.8){\cite{wang2024disco}} 
\put(36.5, 185.8){\cite{fang2023dance}} 
\put(27.9, 175.9){\cite{zhu2024poseanimate}} 
\put(171.7, 180.4){\cite{tu2024motionfollower}} 
\put(134.8, 171.4){\cite{wang2024unianimate}} 
\put(203.2, 171.4){\cite{zhang2024mimicmotion}} 
\put(-125.1, 81){\cite{feng2023dreamoving}} 
\put(1.5, 87){\cite{liu2024disentangling}} 
\put(46.5, 87){\cite{zhai2024idol}} 
\put(155.5, 86){\cite{hu2024animate}} 
\put(210.5, 86){\cite{wang2024vividpose}} 
\put(152.8, 76.1){\cite{xue2024follow}} 
\put(197.8, 76.1){\cite{zhu2024champ}} 
    \end{picture}
    \label{fig:DGF}
\vspace{-10px}
\end{figure*}

This process relies on target pose sequences, which define the desired movements and postures over time, and a reference image, which provides the visual appearance that the synthesized video should retain.
To achieve this, most methods~\cite{wang2024disco,shao2024human4dit,zhang2024mimicmotion} rely on deep learning architectures, predominantly utilizing diffusion models, which are trained to understand the relationship between pose and appearance. 
And these generation frameworks based on diffusion models can be roughly divided into three categories: pure noise input, reference image plus noise input, and guided conditions plus noise input, as shown in Fig. \ref{fig:DGF}.

\textbf{(A) Pure Noise for Main Diffusion Branch. }
As illustrated in Fig. \ref{fig:DGF} (A), in this approach, the input of the main diffusion branch is pure noise, with the target image being embedded in one of two ways: either through ReferenceNet, which is another U-Net copy same as the main diffusion branch, or via a feature encoder. 
Simultaneously, the driving condition is typically encoded using ControlNet \cite{zhang2023adding}.
This generation framework is adopted by methods such as MagicPose \cite{chang2024magicpose}, MagicAnimate \cite{xu2024magicanimate}, TCAN \cite{kim2024tcan}, and DreaMoving \cite{feng2023dreamoving}.

MagicAnimate \cite{xu2024magicanimate} uses pure noise as the input for a diffusion model and applies an image-video joint training strategy to leverage diverse single-frame image data for augmentation. 
MagicPose~\cite{chang2024magicpose} introduces a two-stage training strategy to disentangle human motions and appearance (e.g., facial expressions, skin tone, and dressing), consisting of the pre-training of an appearance-control block and learning appearance-disentangled pose control. 
TCAN \cite{kim2024tcan} leverages a pre-trained ControlNet~\cite{zhang2023adding} and adapts the LoRA \cite{hulora} technique to the U-Net \cite{ronneberger2015u} layers, aligning the latent space between pose and appearance features. 
DreaMoving~\cite{feng2023dreamoving} argues that the cost of using the diffusion model for appearance coding is high. Thus, they replace the appearance encoder with a feature encoder with a multi-layer convolutional network, shown in Fig. \ref{fig:DGF} (A2).
Pure noise input can produce a wider range of diverse outcomes, relying heavily on the generalization ability of pre-trained models. Even with pure noise as the input, the model can leverage prior knowledge to progressively denoise and generate images or videos that adhere to vision logic.



\textbf{(B) Reference Image for Main Diffusion Branch. }
Fig.~\ref{fig:DGF}~(B) illuminates the approach where noise is added to the reference image as input to the main diffusion branch, while the guided conditions are encoded using ControlNet \cite{zhang2023adding} or a feature encoder. 
Disco \cite{wang2024disco}, STSA \cite{fang2023dance}, and PoseAnimate \cite{zhu2024poseanimate} utilize ControlNet to encode the driving signal.
DFBM \cite{liu2024disentangling} and IDOL \cite{zhai2024idol} employ feature encoders with multi-layer convolutional networks to encode the guided conditions.

Disco \cite{wang2024disco} is an early method for pose-driven dance video generation. 
It introduces a disentangled control architecture that separates the manipulation of human foreground, background, and pose. 
The human foreground is processed through a cross attention mechanism that utilizes local CLIP \cite{radford2021learning} image embeddings to capture fine-grained human semantics, while the pose is controlled via a dedicated ControlNet \cite{zhang2023adding} branch. 
STSA \cite{fang2023dance} proposes spatial-temporal subspace-attention blocks that decompose the global space into a combination of regular subspaces, enabling efficient modelling of spatio-temporal consistency. 
PoseAnimate~\cite{zhu2024poseanimate} addresses potential disruptions in character identity and background details by replacing self attention layers in the U-Net \cite{ronneberger2015u} architecture with a dual consistency attention mechanism.
Similar to DreaMoving~\cite{feng2023dreamoving}, DFBM~\cite{liu2024disentangling} replaces ControlNet~\cite{zhang2023adding} with a feature encoder to learn both foreground and background dynamics using distinct motion representations.
By adding noise to the reference image, the model can better preserve key identity features, such as the appearance of the characters, while introducing variations to enhance the diversity of the generated content.

%
%
\textbf{(C) Guided Conditions for Main Diffusion Branch. } 
Another design approach based on the diffusion models, as shown in Fig. \ref{fig:DGF} (C), involves adding noise to the guided conditions as the input to the main diffusion branch. 
In this approach, the reference image is used to encode appearance features either through a feature encoder (C1) or ReferenceNet (C2). 
For example, Animate Anyone \cite{hu2024animate}, VividPose \cite{wang2024vividpose}, Follow-your-pose v2~\cite{xue2024follow}, and Champ \cite{zhu2024champ} adopt the U-Net copy to encode the appearance features of the reference image, while MotionFollower \cite{tu2024motionfollower}, Unianimate \cite{wang2024unianimate}, and MimicMotion \cite{zhang2024mimicmotion} use the feature encoder for the reference image.

Animate Anyone \cite{hu2024animate}, has demonstrated remarkable success in pose-driven dance video generation. 
During the training stage, the model first conditions on individual video frames to prioritize spatial feature extraction and pose guidance, while the temporal layers are not considered. 
Subsequently, the temporal layer is seamlessly integrated, and the model is further refined using different video clips, while the rest of the network's weights remain fixed.
This approach ensures a harmonious blend of spatial and temporal coherence. 
VividPose \cite{wang2024vividpose} utilizes an identity-aware appearance controller that skillfully incorporates facial features, ensuring the preservation of the subject's identity across various poses without compromising other visual details.
Follow-Your-Pose V2~\cite{xue2024follow}  stabilizes the background by leveraging the guidance provided by optical flow. 
Champ \cite{zhu2024champ} integrates a 3D human parametric model within a latent diffusion framework, using the SMPL model to establish a unified representation of body shape.

To reduce network costs, MotionFollower \cite{tu2024motionfollower} employs a feature encoder as a lightweight alternative to the time-consuming DDIM \cite{song2020denoising} inversion, which can lead to significant infidelity, as shown in Fig. \ref{fig:DGF} (C2). 
Unianimate \cite{wang2024unianimate} introduces the use of temporal Mamba~\cite{gu2024mambalineartimesequencemodeling}, a state-space model, as an alternative to the traditional temporal Transformer, significantly enhancing efficiency and enabling the processing of longer video sequences with linear time complexity.
MimicMotion \cite{zhang2024mimicmotion} implements a hand region enhancement strategy that focuses on improving the visual quality of hand regions.



\begin{figure}[!t]
    \centering
    \vspace{-10px}
    \includegraphics[width=\linewidth]{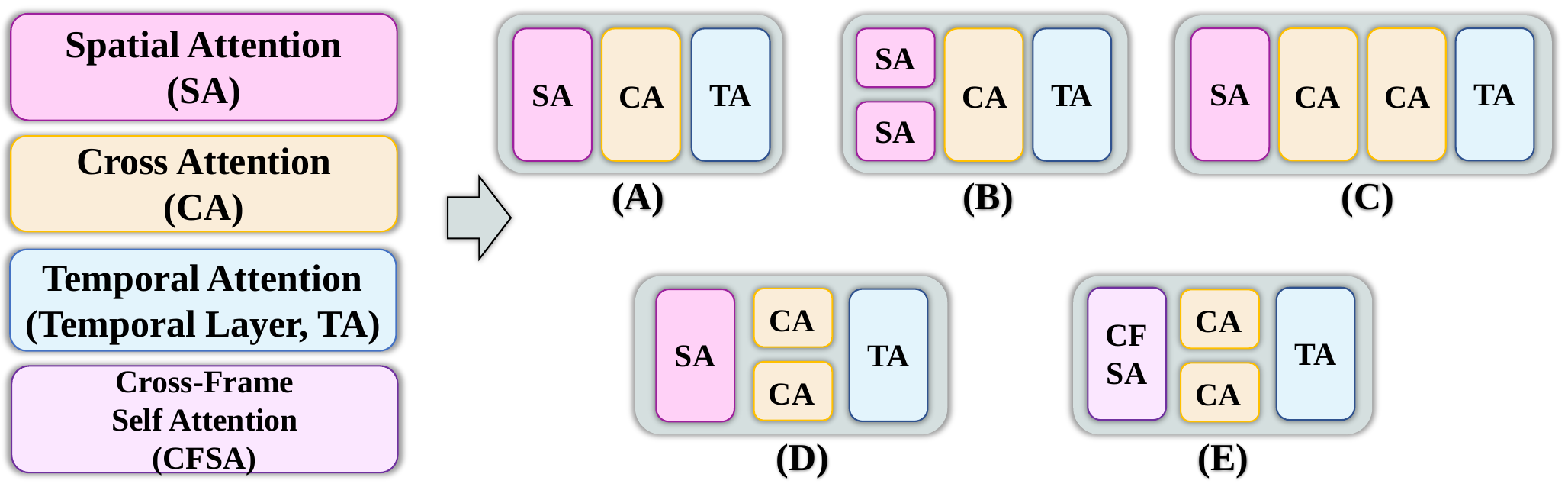}
    \vspace{-20px}
    \caption{Different attention fusion methods of diffusion-based vision-driven human motion video generation.
    }
    \label{fig:sumattention}
    \vspace{-8px}
\end{figure}

Moreover, 
most methods for designing diffusion models primarily incorporate intra-frame and inter-frame attention mechanisms. Since the original diffusion model lacks a temporal dimension, many approaches introduce a temporal layer to implement inter-frame attention, thereby enhancing the consistency of video frames. 
\new{Additionally, by analyzing the placement of spatial attention (SA), cross attention (CA), temporal attention (TA) and cross-frame self attention (CFSA) layers, we further classify different approaches into five variants, as shown in Fig. \ref{fig:sumattention}.}

\textbf{SA-CA-TA.} This variant, as shown in Fig. \ref{fig:sumattention} (A),  employs a single self attention layer followed by a cross attention layer to control single-frame image generation. 
Temporal attention layers are used to ensure multi-frame temporal continuity, with the features and semantics of the reference image injected into the noise predictor of the main diffusion model branch through a CLIP \cite{radford2021learning} encoder or ControlNet~\cite{van2017neural}. 
Disco \cite{wang2024disco}, DreaMoving~\cite{feng2023dreamoving}, MotionFollower~\cite{tu2024motionfollower}, and MimicMotion \cite{zhang2024mimicmotion} adopt this approach.

\textbf{SA\&SA-CA-TA.} Since image encoders alone may not capture fine-grained features in reference images, this method employs an additional ReferenceNet and integrates the self attention layer features of ReferenceNet with the main diffusion model branch. 
We define this category of methods as hierarchical self attention as illustrated in Fig. \ref{fig:sumattention} (B). 
MagicPose \cite{chang2024magicpose}, Animate Anyone \cite{hu2024animate}, MagicAnimate \cite{xu2024magicanimate}, TCAN \cite{kim2024tcan}, Follow-Your-Pose v2~\cite{xue2024follow}, and Champ \cite{zhu2024champ} are based on this calculation method.

\textbf{SA-CA-CA-TA.} To better handle multiple control signals, which can be challenging with a single cross attention layer, as shown in Fig. \ref{fig:sumattention} (C), this approach enhances controllability by adding an extra following cross attention layer. 
Hallo \cite{xu2024hallo} uses this method, combining additional audio signals to generate facial and mouth movements.

\textbf{SA-CA\&CA-TA.} Since cross attention layers typically focus on the global features of the reference image, important semantic details that are critical to human perception may be diminished. 
VividPose \cite{wang2024vividpose}, inspired by IP-Adapter \cite{ye2023ip}, uses a hierarchical cross attention to enhance facial semantics, as illustrated in Fig. \ref{fig:sumattention} (D), thereby improving identity consistency.

\textbf{CFSA-CA\&CA-TA.} A direct way to enhance inter-frame correlation is to perform attention calculations between every pixel of the current frame and every pixel of other parts or all frames, as shown in Fig. \ref{fig:sumattention} (E). 
Follow-Your-Pose~\cite{ma2024followyp} and PoseAnimate \cite{zhu2024poseanimate} extend the basic self attention mechanism to cross-frame self attention, enhancing temporal coherence in video-based pose animation.
However, compared to temporal attention, cross-frame attention results in higher computational costs and increased memory consumption.

In addition to the generative approach of diffusion models, Human4DiT \cite{shao2024human4dit} explores a video generation framework centered around Transformers. 
Human4DiT employs a 4D diffusion Transformer to capture intricate correlations across views, time, and spatial dimensions, enabling the generation of coherent and realistic human videos from a single reference image.
In conclusion, our investigation reveals key insights:

\begin{itemize}

\item \new{Methods that use DWPose skeletal structures~\cite{yang2023dwpose} as driving conditions~\cite{wang2024disco,fang2023dance,chang2024magicpose,xu2024magicanimate,hu2024animate,feng2023dreamoving,guo2023i2v} face structural inconsistencies when processing subjects with varying body proportions, leading to texture distortions in facial, manual, and other anatomical regions. Furthermore, existing approaches~\cite{liu2024disentangling,wang2024vividpose,tu2024motionfollower,wang2024unianimate} employ Unet-based diffusion models for video generation, where frame-by-frame generation leads to temporal instability and flickering artifacts. Investigating alternative video generation architectures, such as those using DiT~\cite{peebles2023scalable} or VAR~\cite{tian2024visual} frameworks, could offer more stable and coherent results.}



\item \new{Recent studies~\cite{liu2024disentangling,xu2024magicanimate,li2024synthesizing} show improved integrity of human representation in videos through techniques like foreground-background separation, multi-modal training with images and videos, and 3D signal incorporation. However, the reliance on multiple disentangled signals increases computational overhead, highlighting the need for efficient end-to-end training paradigms.}

\end{itemize}

\noindent \textbf{Try-On Video Generation.}
Try-On video generation is another intriguing video generation task. 
A recent survey \cite{song2023image} discussed the development of current virtual Try-On technology.
We briefly mention recent Try-On methods, such as ViViD \cite{fang2024vivid}, Tunnel Try-On \cite{xu2024tunnel}, and WildVidFit \cite{he2024wildvidfit}, which are summarized in Appendix B.
Tunnel Try-On \cite{xu2024tunnel} utilizes a Kalman filter to smooth the motion within the focused tunnel, ensuring temporal coherence. 
ViViD \cite{fang2024vivid} collects a new, diverse and the largest dataset for video virtual Try-On tasks.
WildVidFit \cite{he2024wildvidfit} employs a diffusion guidance module that leverages pre-trained models to enhance temporal coherence without the need for explicit temporal training.

\noindent \textbf{Pose2Video.}
Pose2Video extends video generation beyond dance scenarios, encompassing general human motion generation. Notably, Make-Your-Anchor \cite{huang2024make} advances this field by incorporating precise torso and hand movements within a diffusion-based pose sequence generation framework, requiring only a one-minute video clip for effective training. DreamPose \cite{karras2023dreampose} generates human and fabric motions simultaneously by adapting the Stable Diffusion \cite{yang2023diffusion} model into a pose and image-guided video synthesis system. Despite these advancements, recent studies \cite{zhu2024poseanimate} in human motion video generation highlight ongoing challenges in maintaining character consistency and temporal coherence.

\vspace{-10px}
\subsection{Text-Driven Human Motion Video Generation}

Text-driven human motion video generation can be broadly categorized into Text2Face and Text2MotionVideo. 
Recent works are respectively listed in Appendix C and D.

\noindent \textbf{Text2Face.}
Text2Face typically focuses on generating talking faces, where text can be used as first-person scripts to control facial movements~\cite{liu2024towards} or as instructions to influence the style or content of the video~\cite{jiang2023text2performer}.
Text2Face generation can be divided into two approaches: first-person scripts and third-person instructions. 
The first-person statement approach aims to control the facial movements to synchronize with the spoken text~\cite{jang2024faces}, while third-person instruction methods focus on learning the mapping between objective text descriptions and the generated video~\cite{wang2024videocomposer}.

\begin{figure}[!t]
    \centering
    \vspace{-8px}
    \includegraphics[width=\linewidth]{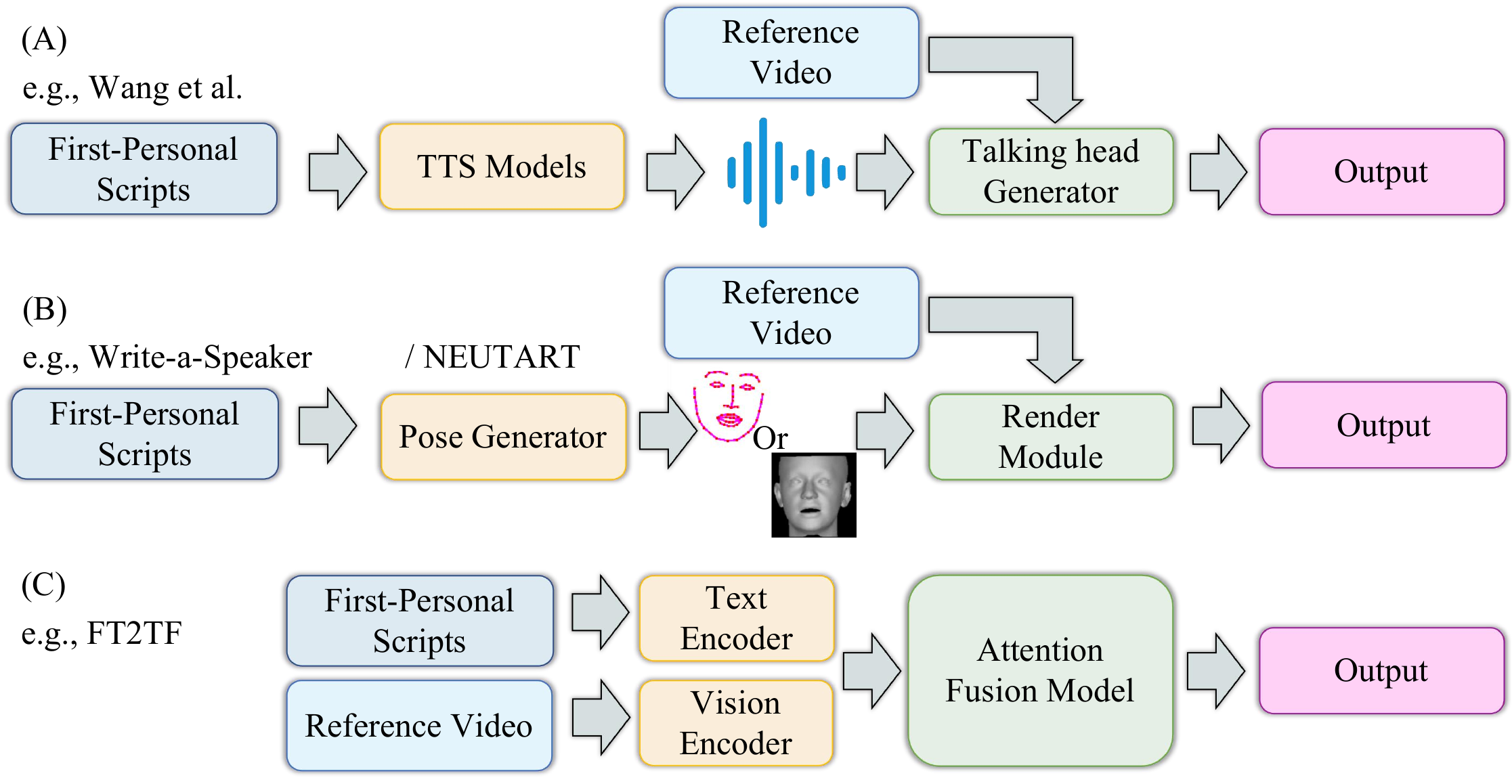}
    \vspace{-15px}
    \caption{
    Different Text2Face pipelines for first-personal scripts.
    }
    \begin{picture}(0,0)
    \put(-85, 141){\scalebox{0.7}{\cite{wang2023text}}} 
    \put(-73, 96){\scalebox{0.7}{\cite{li2021write}}} 
    \put(-29, 96){\scalebox{0.7}{\cite{milis2023neural}}} 
    \put(-96, 50){\scalebox{0.7}{\cite{diao2023ft2tf}}} 
    \end{picture}
    \label{fig:sumtext}
    \vspace{-18px}
\end{figure}
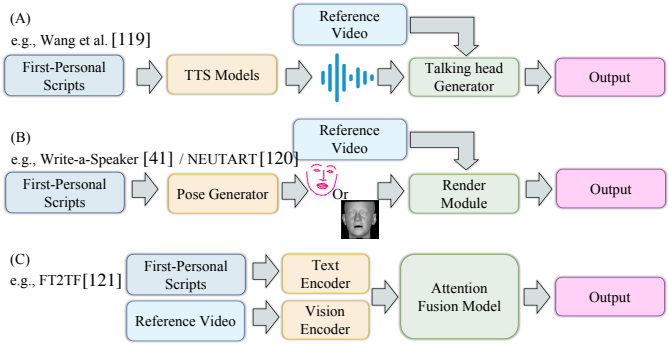

As illustrated in Fig. \ref{fig:sumtext}, when text is utilized for first-person scripts, there are three distinct types of pipelines. Most methods generate intermediate outputs such as audios or landmarks, which subsequently control the production of talking head videos, as demonstrated in Fig. \ref{fig:sumtext} (A) and (B). Notably, only the FT2FT~\cite{diao2023ft2tf} adopts an end-to-end pipeline that directly generates videos without relying on any intermediate outputs.
Wang et al.~\cite{wang2023text} convert text to speech using a Text-To-Speech (TTS) model~\cite{casanova2022yourtts}, effectively transforming the Text2Face task into a traditional talking head generation task. 
Write-a-Speaker.~\cite{li2021write} design three parallel networks to generate 3D parameters from time-aligned text, which are then used to produce the final talking face videos. 
Taking it a step further, NEUTART~\cite{milis2023neural} simultaneously generates audio and employs a lipreading loss~\cite{ma2022visual} to extract lip features. 
When text is used as instructions, it typically influences the content and style of the generated video, with additional signals required to preserve the subject's identity. 
He et al.~\cite{he2024id} address this by training an ID-adapter to control identity, while using text prompts to influence motion and background. 
Ma et al.~\cite{ma2024magic} introduce a simple effective framework for generating subject-controllable videos, focusing on identity reinforcement and frame-wise correlation injection during initialization to ensure stable video outputs.

\noindent \textbf{Text2MotionVideo.}
Text2MotionVideo tasks involve generating human motion videos where textual inputs primarily serve as instructions to influence the content or style of the video~\cite{ma2024follow,khachatryan2023text2video,shi2024bivdiff}. These tasks can be categorized into two types: ID-preservation: generated results retain input identity. ID-transfer: identity may change.

For ID-preserved video generation, the identity is primarily derived from the input images or videos. 
When other explicit conditions (e.g., keypoint) are used to control motion, the text's role is often limited, typically serving as a caption for the generated video. 
Specifically, 
Wang et al.~\cite{wang2024videocomposer} propose a spatio-temporal condition encoder capable of encoding various conditional information, allowing for more flexible control. However, the input text must be closely related to the input image for better results. 
Zhang et al.~\cite{zhang2023magicavatar} adopt a two-stage method that first converts multi-modal inputs into control signals (such as human pose, depth, and DensePose~\cite{guler2018densepose}) and then generate videos by these control signals.

In the absence of other explicit conditions, the text provides a general motion directive, which can guide the animation of a static human portrait or directly generate a human motion video from text, typically accomplished directly through text-to-video models~\cite{guo2023animatediff}.
Recent methods in this area demonstrate a trend toward more stable training, easier control, and finer granularity in identity preservation. VideoCrafter 1~\cite{chen2023videocrafter1} and VideoCrafter 2~\cite{chen2024videocrafter2} explore stable generation video generation model training schemes extended from diffusion model, using text, images, and low-quality videos to achieve high-quality video models~\cite{chen2023videocrafter1,chen2024videocrafter2}.
To enhance control, Guo et al.~\cite{guo2023animatediff} propose a plug-and-play motion module that can be trained once and integrated into any personalized text-to-image (T2I) model, adding motion dynamics to reference images. 
Additionally, AnimateZero~\cite{yu2023animatezero} offers spatial appearance control and temporal consistency by replacing the global time node of the original text-to-video model with a position correction window and utilizing intermediate latent embeddings from T2I generation. 
In terms of finer granularity, Renshuai et al.~\cite{liu2024towards} refine fine-grained emotional expression, expanding from 8 sentiment indicators to 135 detailed emotional descriptions. They introduce a framework that simultaneously controls identity, expression, and background from multi-modal inputs, addressing emotional challenges with 
maintaining identity.


\begin{table*}[!t]
    \vspace{-15px}
    \renewcommand{\arraystretch}{0.8} 
    \centering
    \caption{Comprehensive overview of lip synchronization and head pose driving. 
    AR: Autoregressive Model. 
    \new{Complete table can be found in appendix E.}
    }
    \vspace{-5px}
    \label{tab:talking head}
    \resizebox{0.86\textwidth}{!}{%
    \begin{tabular}{|ccccccccc|}
    \hline
    \multicolumn{1}{|c|}{\textbf{Driving Region}}                                                              & \multicolumn{1}{c|}{\textbf{\begin{tabular}[c]{@{}c@{}}Publication \\ Time\end{tabular}}} & \multicolumn{1}{c|}{\textbf{Paper}}                                                                           & \multicolumn{1}{c|}{\textbf{Input Signals}}                                                                       & \multicolumn{1}{c|}{\textbf{\begin{tabular}[c]{@{}c@{}}Motion \\ Representation\end{tabular}}}              & \multicolumn{1}{c|}{\textbf{Backbone}}                                                                       & \multicolumn{1}{c|}{\textbf{\begin{tabular}[c]{@{}c@{}}Open \\ Source\end{tabular}}}                                                             & \multicolumn{1}{c|}{\textbf{\begin{tabular}[c]{@{}c@{}}\new{Real-Time}\\ \new{Inference}\end{tabular}}} & \textbf{Venue}                                                               \\ \hline
    \multicolumn{9}{|c|}{\cellcolor[HTML]{D1DADC}\rule{0pt}{2.5ex}\textbf{Lip Synchronization}}                                                                                                                                                                                                                                                                                                                                                                                                                                                                                                                                                                                                                                                                                                                                                                                           \\ \hline
    \multicolumn{1}{|c|}{}                                                                                     & \multicolumn{1}{c|}{\cellcolor[HTML]{ECF2F0}Sep. 17, 2020}                                   & \multicolumn{1}{c|}{\cellcolor[HTML]{ECF2F0}AudioDVP \cite{wen2020photorealistic}}                                  & \multicolumn{1}{c|}{\cellcolor[HTML]{ECF2F0}\audiologo + \reallogo(video)}                                       & \multicolumn{1}{c|}{\cellcolor[HTML]{ECF2F0}Region}                                                         & \multicolumn{1}{c|}{\cellcolor[HTML]{ECF2F0}\begin{tabular}[c]{@{}c@{}}ED\end{tabular}}       & \multicolumn{1}{c|}{\cellcolor[HTML]{ECF2F0}\checkmark}                                                         & {\cellcolor[HTML]{ECF2F0}\new{×}} & \multicolumn{1}{|c|}{\cellcolor[HTML]{ECF2F0}\begin{tabular}[c]{@{}c@{}}TVCG'20\end{tabular}}  \\  
    \multicolumn{1}{|c|}{}                                                                                     & \multicolumn{1}{c|}{May. 9, 2019}                                                           & \multicolumn{1}{c|}{ATVGnet \cite{chen2019hierarchical}}                                                            & \multicolumn{1}{c|}{\audiologo + \reallogo}                                                                      & \multicolumn{1}{c|}{KeyPoint}                                                                               & \multicolumn{1}{c|}{\begin{tabular}[c]{@{}c@{}}AR\end{tabular}}                         & \multicolumn{1}{c|}{\checkmark}                                                                                 &  
    {\realtimelogo \new{35 FPS}} & \multicolumn{1}{|c|}{\begin{tabular}[c]{@{}c@{}}CVPR'19\end{tabular}}                          \\  
    \multicolumn{1}{|c|}{\multirow{-3}{*}{\begin{tabular}[c]{@{}c@{}}\quad \\ \partlogo\\ Part\\ (Face)\end{tabular}}}                                                                                     & \multicolumn{1}{c|}{\cellcolor[HTML]{ECF2F0}May. 8, 2019}                                   & \multicolumn{1}{c|}{\cellcolor[HTML]{ECF2F0}VOCA \cite{cudeiro2019capture}}                                         & \multicolumn{1}{c|}{\cellcolor[HTML]{ECF2F0}\begin{tabular}[c]{@{}c@{}}\audiologo + 3D Template\end{tabular}} & \multicolumn{1}{c|}{\cellcolor[HTML]{ECF2F0}KeyPoint}                                                         & \multicolumn{1}{c|}{\cellcolor[HTML]{ECF2F0}\begin{tabular}[c]{@{}c@{}}ED\end{tabular}}       & \multicolumn{1}{c|}{\cellcolor[HTML]{ECF2F0}\checkmark}                                                & {\cellcolor[HTML]{ECF2F0}\realtimelogo \new{60 FPS}} & \multicolumn{1}{|c|}{\cellcolor[HTML]{ECF2F0}\begin{tabular}[c]{@{}c@{}}CVPR'19\end{tabular}}  \\  
    \multicolumn{1}{|c|}{}  & \multicolumn{1}{c|}{Jan. 10, 2023}                                                           & \multicolumn{1}{c|}{Bigioi et al. \cite{bigioi2024speech}}                                                          & \multicolumn{1}{c|}{\audiologo + \reallogo}                                                                      & \multicolumn{1}{c|}{Latent}                                                                                 & \multicolumn{1}{c|}{\begin{tabular}[c]{@{}c@{}}DM\end{tabular}}                              & \multicolumn{1}{c|}{\checkmark}                                                                        & \multicolumn{1}{c}{\new{×}} & \multicolumn{1}{|c|}{\begin{tabular}[c]{@{}c@{}}IVC'24\end{tabular}}                           \\
         \multicolumn{1}{|c|}{} & \multicolumn{1}{c|}{\cellcolor[HTML]{ECF2F0}Nov. 11, 2020}  
     & \multicolumn{1}{c|}{\cellcolor[HTML]{ECF2F0}Wav2lip \cite{wav2lip}}                                         & \multicolumn{1}{c|}{\cellcolor[HTML]{ECF2F0}\begin{tabular}[c]{@{}c@{}}\audiologo + \reallogo \end{tabular}} & \multicolumn{1}{c|}{\cellcolor[HTML]{ECF2F0}Latent}                                                         & \multicolumn{1}{c|}{\cellcolor[HTML]{ECF2F0}\begin{tabular}[c]{@{}c@{}}ED\end{tabular}}       & \multicolumn{1}{c|}{\cellcolor[HTML]{ECF2F0}\checkmark}                                                & {\cellcolor[HTML]{ECF2F0}\new{×}} & \multicolumn{1}{|c|}{\cellcolor[HTML]{ECF2F0}\begin{tabular}[c]{@{}c@{}}ACM MM'20\end{tabular}}
     \\ \hline
    \multicolumn{9}{|c|}{\cellcolor[HTML]{D1DADC}\rule{0pt}{2.5ex}\textbf{Head Pose Driving}}                                                                                                                                                                                                                                                                                                                                                                                                                                                                                                                                                                                                                                                                                                                                                                                             \\ \hline
    \multicolumn{1}{|c|}{}                                                                                     & \multicolumn{1}{c|}{Mar. 26, 2024}                                                           & \multicolumn{1}{c|}{AniPortrait \cite{wei2024aniportrait}}                                                          & \multicolumn{1}{c|}{\audiologo + \reallogo}                                                                      & \multicolumn{1}{c|}{\begin{tabular}[c]{@{}c@{}}KeyPoint + 3D-P\end{tabular}}           & \multicolumn{1}{c|}{\begin{tabular}[c]{@{}c@{}}DM\end{tabular}}                              & \multicolumn{1}{c|}{\checkmark}                                                                        & \multicolumn{1}{c}{\new{×}} & \multicolumn{1}{|c|}{arXiv}                                                                        \\  
    \multicolumn{1}{|c|}{}                                                                                     & \multicolumn{1}{c|}{\cellcolor[HTML]{ECF2F0}Jan. 10, 2023}                                   & \multicolumn{1}{c|}{\cellcolor[HTML]{ECF2F0}DiffTalk \cite{shen2023difftalk}}                                       & \multicolumn{1}{c|}{\cellcolor[HTML]{ECF2F0}\audiologo + \reallogo}                                              & \multicolumn{1}{c|}{\cellcolor[HTML]{ECF2F0}Latent}                                                       & \multicolumn{1}{c|}{\cellcolor[HTML]{ECF2F0}\begin{tabular}[c]{@{}c@{}}DM\end{tabular}}      & \multicolumn{1}{c|}{\cellcolor[HTML]{ECF2F0}\checkmark}                                                & {\cellcolor[HTML]{ECF2F0}\new{×}} & \multicolumn{1}{|c|}{\cellcolor[HTML]{ECF2F0}\begin{tabular}[c]{@{}c@{}}CVPR'23\end{tabular}}  \\  
    \multicolumn{1}{|c|}{}                                                                                     & \multicolumn{1}{c|}{Mar. 16, 2022}                                                           & \multicolumn{1}{c|}{StyleHEAT \cite{yin2022styleheat}}                                                              & \multicolumn{1}{c|}{\audiologo + \reallogo + \drivingreallogo}                                                   & \multicolumn{1}{c|}{Region}                                                                                 & \multicolumn{1}{c|}{GAN}                                                                                     & \multicolumn{1}{c|}{\checkmark}                                                                        & \multicolumn{1}{c}{\new{×}} & \multicolumn{1}{|c|}{\begin{tabular}[c]{@{}c@{}}ECCV'22\end{tabular}}                                                                        \\  
    \multicolumn{1}{|c|}{}                                                                                     & \multicolumn{1}{c|}{\cellcolor[HTML]{ECF2F0}Feb. 20, 2023}                                   & \multicolumn{1}{c|}{\cellcolor[HTML]{ECF2F0}SD-NeRF \cite{shen2023sd}}                                              & \multicolumn{1}{c|}{\cellcolor[HTML]{ECF2F0}\audiologo + \reallogo(video)}                                       & \multicolumn{1}{c|}{\cellcolor[HTML]{ECF2F0}\begin{tabular}[c]{@{}c@{}}3D-P\end{tabular}} & \multicolumn{1}{c|}{\cellcolor[HTML]{ECF2F0}\begin{tabular}[c]{@{}c@{}}ED\end{tabular}}       & \multicolumn{1}{c|}{\cellcolor[HTML]{ECF2F0}×}                                                         & {\cellcolor[HTML]{ECF2F0}\new{×}} & \multicolumn{1}{|c|}{\cellcolor[HTML]{ECF2F0}\begin{tabular}[c]{@{}c@{}}TMM'23\end{tabular}}                                                \\  
    \multicolumn{1}{|c|}{}                                                                                     & \multicolumn{1}{c|}{Jun. 12, 2024}                                                           & \multicolumn{1}{c|}{Liang et al. \cite{liang2024emotional}}                                                         & \multicolumn{1}{c|}{\audiologo + \reallogo + \textlogo}                                                          & \multicolumn{1}{c|}{\begin{tabular}[c]{@{}c@{}}KeyPoint + 3D-P\end{tabular}}           & \multicolumn{1}{c|}{\begin{tabular}[c]{@{}c@{}}DM\end{tabular}}                              & \multicolumn{1}{c|}{×}                                                                                 & \multicolumn{1}{c}{\new{×}} & \multicolumn{1}{|c|}{arXiv}                                                                        \\  
    \multicolumn{1}{|c|}{}                                                                                     & \multicolumn{1}{c|}{\cellcolor[HTML]{ECF2F0}Jan. 6, 2023}                                   & \multicolumn{1}{c|}{\cellcolor[HTML]{ECF2F0}Diffused Heads \cite{stypulkowski2024diffused}}                         & \multicolumn{1}{c|}{\cellcolor[HTML]{ECF2F0}\audiologo + \reallogo}                                              & \multicolumn{1}{c|}{\cellcolor[HTML]{ECF2F0}Latent}                                                         & \multicolumn{1}{c|}{\cellcolor[HTML]{ECF2F0}\begin{tabular}[c]{@{}c@{}}DM\end{tabular}}      & \multicolumn{1}{c|}{\cellcolor[HTML]{ECF2F0}\checkmark}                                                & {\cellcolor[HTML]{ECF2F0}\new{×}} & \multicolumn{1}{|c|}{\cellcolor[HTML]{ECF2F0}\begin{tabular}[c]{@{}c@{}}CVPR'24\end{tabular}}  \\  
    \multicolumn{1}{|c|}{}                                                                                     & \multicolumn{1}{c|}{Mar. 30, 2023}                                                           & \multicolumn{1}{c|}{DAE-Talker \cite{du2023dae}}                                                                    & \multicolumn{1}{c|}{\audiologo + \reallogo}                                                                      & \multicolumn{1}{c|}{Latent}                                                                                 & \multicolumn{1}{c|}{\begin{tabular}[c]{@{}c@{}}DM\end{tabular}}                              & \multicolumn{1}{c|}{×}                                                                                 & \multicolumn{1}{c}{\new{×}} & \multicolumn{1}{|c|}{\begin{tabular}[c]{@{}c@{}}ACM MM'23\end{tabular}}                         \\  
    \multicolumn{1}{|c|}{}                                                                                     & \multicolumn{1}{c|}{\cellcolor[HTML]{ECF2F0}May. 6, 2024}                                   & \multicolumn{1}{c|}{\cellcolor[HTML]{ECF2F0}AniTalker \cite{liu2024anitalker}}                                      & \multicolumn{1}{c|}{\cellcolor[HTML]{ECF2F0}\audiologo + \reallogo}                                              & \multicolumn{1}{c|}{\cellcolor[HTML]{ECF2F0}Latent}                                                         & \multicolumn{1}{c|}{\cellcolor[HTML]{ECF2F0}\begin{tabular}[c]{@{}c@{}}ED\end{tabular}}       & \multicolumn{1}{c|}{\cellcolor[HTML]{ECF2F0}\begin{tabular}[c]{@{}c@{}}\checkmark\end{tabular}}                                                         & {\cellcolor[HTML]{ECF2F0}\new{×}} & \multicolumn{1}{|c|}{\cellcolor[HTML]{ECF2F0}\begin{tabular}[c]{@{}c@{}}\new{ACM MM'24}\end{tabular}}                                                \\  
    \multicolumn{1}{|c|}{}                                                                                     & \multicolumn{1}{c|}{Jul. 12, 2024}                                                           & \multicolumn{1}{c|}{EchoMimic \cite{chen2024echomimic}}                                                             & \multicolumn{1}{c|}{\audiologo + \reallogo}                                                                      & \multicolumn{1}{c|}{KeyPoint}                                                                               & \multicolumn{1}{c|}{\begin{tabular}[c]{@{}c@{}}DM\end{tabular}}                              & \multicolumn{1}{c|}{\begin{tabular}[c]{@{}c@{}}\checkmark\end{tabular}}                         & \multicolumn{1}{c}{\new{×}} & \multicolumn{1}{|c|}{arXiv}                                                                        \\  
    \multicolumn{1}{|c|}{\multirow{-8}{*}{\begin{tabular}[c]{@{}c@{}}\partlogo\\ Part\\ (Face)\end{tabular}}}                                                                                     & \multicolumn{1}{c|}{\cellcolor[HTML]{ECF2F0}May. 4, 2023}                                   & \multicolumn{1}{c|}{\cellcolor[HTML]{ECF2F0}Xu et al. \cite{xu2023high}}                                            & \multicolumn{1}{c|}{\cellcolor[HTML]{ECF2F0}\audiologo + \reallogo}                                              & \multicolumn{1}{c|}{\cellcolor[HTML]{ECF2F0}Latent}                                                         & \multicolumn{1}{c|}{\cellcolor[HTML]{ECF2F0}Transformer}                                                     & \multicolumn{1}{c|}{\cellcolor[HTML]{ECF2F0}×}                                                         & {\cellcolor[HTML]{ECF2F0}\new{×}} & \multicolumn{1}{|c|}{\cellcolor[HTML]{ECF2F0}\begin{tabular}[c]{@{}c@{}}CVPR'23\end{tabular}}  \\  
    \multicolumn{1}{|c|}{}                                                                                     & \multicolumn{1}{c|}{Apr. 2, 2024}                                                           & \multicolumn{1}{c|}{EDTalk \cite{tan2024edtalk}}                                                                    & \multicolumn{1}{c|}{\audiologo + \reallogo}                                                                      & \multicolumn{1}{c|}{Latent}                                                                                 & \multicolumn{1}{c|}{GAN}                                                                                     & \multicolumn{1}{c|}{\checkmark}                                                                                 & \multicolumn{1}{c}{\new{×}} & \multicolumn{1}{|c|}{\begin{tabular}[c]{@{}c@{}}ECCV'24\end{tabular}}                                                                        \\  
    \multicolumn{1}{|c|}{}                                                                                     & \multicolumn{1}{c|}{\cellcolor[HTML]{ECF2F0}Nov. 29, 2023}                                   & \multicolumn{1}{c|}{\cellcolor[HTML]{ECF2F0}SyncTalk \cite{peng2024synctalk}}                                       & \multicolumn{1}{c|}{\cellcolor[HTML]{ECF2F0}\audiologo + \reallogo(video)}                                       & \multicolumn{1}{c|}{\cellcolor[HTML]{ECF2F0}\begin{tabular}[c]{@{}c@{}}3D-P\end{tabular}} & \multicolumn{1}{c|}{\cellcolor[HTML]{ECF2F0}\begin{tabular}[c]{@{}c@{}}ED\end{tabular}}       & \multicolumn{1}{c|}{\cellcolor[HTML]{ECF2F0}\checkmark}                                                & {\cellcolor[HTML]{ECF2F0}\realtimelogo \new{50 FPS}} & \multicolumn{1}{|c|}{\cellcolor[HTML]{ECF2F0}\begin{tabular}[c]{@{}c@{}}CVPR'24\end{tabular}}  \\  
    \multicolumn{1}{|c|}{}                                                                                     & \multicolumn{1}{c|}{Apr. 23, 2024}                                                           & \multicolumn{1}{c|}{TalkingGaussian \cite{li2024talkinggaussian}}                                                   & \multicolumn{1}{c|}{\audiologo + \reallogo(video)}                                                               & \multicolumn{1}{c|}{\begin{tabular}[c]{@{}c@{}}3D-P\end{tabular}}                         & \multicolumn{1}{c|}{\begin{tabular}[c]{@{}c@{}}ED\end{tabular}}                               & \multicolumn{1}{c|}{\checkmark}                                                                                 & {\realtimelogo \new{108 FPS}} & \multicolumn{1}{|c|}{\begin{tabular}[c]{@{}c@{}}ECCV'24\end{tabular}}                                                                        \\  
    \multicolumn{1}{|c|}{}                                                                                     & \multicolumn{1}{c|}{\cellcolor[HTML]{ECF2F0}Apr. 28, 2024}                                   & \multicolumn{1}{c|}{\cellcolor[HTML]{ECF2F0}GaussianTalker \cite{yu2024gaussiantalker}}                             & \multicolumn{1}{c|}{\cellcolor[HTML]{ECF2F0}\audiologo + \reallogo}                                              & \multicolumn{1}{c|}{\cellcolor[HTML]{ECF2F0}\begin{tabular}[c]{@{}c@{}}3D-P\end{tabular}} & \multicolumn{1}{c|}{\cellcolor[HTML]{ECF2F0}\begin{tabular}[c]{@{}c@{}}ED\end{tabular}}       & \multicolumn{1}{c|}{\cellcolor[HTML]{ECF2F0}\checkmark}                                                         & {\cellcolor[HTML]{ECF2F0}\realtimelogo \new{120 FPS}} & \multicolumn{1}{|c|}{\cellcolor[HTML]{ECF2F0}\begin{tabular}[c]{@{}c@{}}ACM MM'24\end{tabular}}                                                \\  
    \multicolumn{1}{|c|}{}                                                                                     & \multicolumn{1}{c|}{Dec. 10, 2021}                                                           & \multicolumn{1}{c|}{FaceFormer \cite{fan2022faceformer}}                                                            & \multicolumn{1}{c|}{\audiologo}                                                                                  & \multicolumn{1}{c|}{KeyPoint}                                                                                 & \multicolumn{1}{c|}{Transformer}                                                                             & \multicolumn{1}{c|}{\checkmark}                                                                        & \multicolumn{1}{c}{\new{×}} & \multicolumn{1}{|c|}{\begin{tabular}[c]{@{}c@{}}CVPR'22\end{tabular}}                          \\  
    \multicolumn{1}{|c|}{}                                                                                     & \multicolumn{1}{c|}{\cellcolor[HTML]{ECF2F0}Sep. 15, 2023}                                   & \multicolumn{1}{c|}{\cellcolor[HTML]{ECF2F0}Delbosc et al. \cite{delbosc2023towards}}                               & \multicolumn{1}{c|}{\cellcolor[HTML]{ECF2F0}\audiologo}                                                          & \multicolumn{1}{c|}{\cellcolor[HTML]{ECF2F0}Latent}                                                         & \multicolumn{1}{c|}{\cellcolor[HTML]{ECF2F0}GAN}                                                             & \multicolumn{1}{c|}{\cellcolor[HTML]{ECF2F0}×}                                                         & {\cellcolor[HTML]{ECF2F0}\new{×}} & \multicolumn{1}{|c|}{\cellcolor[HTML]{ECF2F0}\begin{tabular}[c]{@{}c@{}}ICMI '23\end{tabular}} \\  
    \multicolumn{1}{|c|}{} & \multicolumn{1}{c|}{Oct. 17, 2023}                                                           & \multicolumn{1}{c|}{CorrTalk \cite{chu2024corrtalk}}                                                                & \multicolumn{1}{c|}{\audiologo}                                                                                  & \multicolumn{1}{c|}{Latent}                                                                                 & \multicolumn{1}{c|}{\begin{tabular}[c]{@{}c@{}}ED\end{tabular}}                               & \multicolumn{1}{c|}{×}                                                                                 & \multicolumn{1}{c}{\new{×}} & \multicolumn{1}{|c|}{\begin{tabular}[c]{@{}c@{}}TCSVT'24\end{tabular}}                          \\ \hline
    \multicolumn{9}{|l|}{\reallogo: Reference Real Images; \drivingreallogo: Driving Real Video; \textlogo: Text Prompts; \audiologo: Audio; \new{\realtimelogo: Real-Time Support 
(Estimated FPS based on NVIDIA RTX 4090 GPU).} }                                                                                                                                                                                                                                                                                                                                                                                                                                                                                                                                                                                                                                                                                               \\ \hline
    \end{tabular}%
    }
    \vspace{-10px}
    \end{table*}

In the context of identity transfer, 
current research primarily focuses on maintaining global features such as spatial and temporal consistency during video generation and editing, as well as utilizing lightweight networks to reduce training costs and enhance efficiency. For spatial consistency, Liu et al.~\cite{liu2023dual} propose a dual-stream diffusion network that improves spatial variation consistency in video generation. Ren et al.~\cite{ren2024customize} and Geyer et al.~\cite{geyer2023tokenflow, ma2024follow} enhance temporal consistency by incorporating additional modules and enforcing semantic correspondence across frames. 
Recent efforts have focused on integrating spatial and temporal capabilities. Khachatryan et al.~\cite{khachatryan2023text2video} enhance latent embeddings with motion dynamics to preserve global spatial features, introducing cross-frame attention mechanisms to maintain consistent background timing. Eldesokey and Wonka~\cite{eldesokey2024latentman} develop spatial latent alignment and pixel-wise guidance modules to improve temporal consistency. For cost reduction, Shi et al. introduced BIVDiff~\cite{shi2024bivdiff}, a training-free framework utilizing image diffusion models for video synthesis, while Yang et al. proposed ZeroSmooth~\cite{yang2024zerosmooth}, a training-free video interpolation method for generative video diffusion models. 
Based on our investigation, we discover the following challenges:

\begin{itemize}

\item \new{In Text2MotionVideo, recent research efforts~\cite{zuo2024edit, wang2024videocomposer, zhang2023magicavatar, qin2023dancing} demonstrate that textual inputs exhibit constrained effectiveness when combined with other explicit conditional inputs. The generation performance notably deteriorates for complex motions, particularly when the intricate movement patterns exceed the descriptive capabilities of textual specifications.
}

\item \new{While diffusion-based video generation models demand extensive computational resources, requiring 80GB of GPU memory for fine-tuning, training-free methods~\cite{shi2024bivdiff,yu2023animatezero} are emerging as computationally efficient alternatives.}


\end{itemize}

\vspace{-2px}

\vspace{-10px}
\subsection{Audio-Driven Human Motion Video Generation}

\begin{figure}[!t]
    \centering
    \vspace{-5px}
    \includegraphics[width=\linewidth]{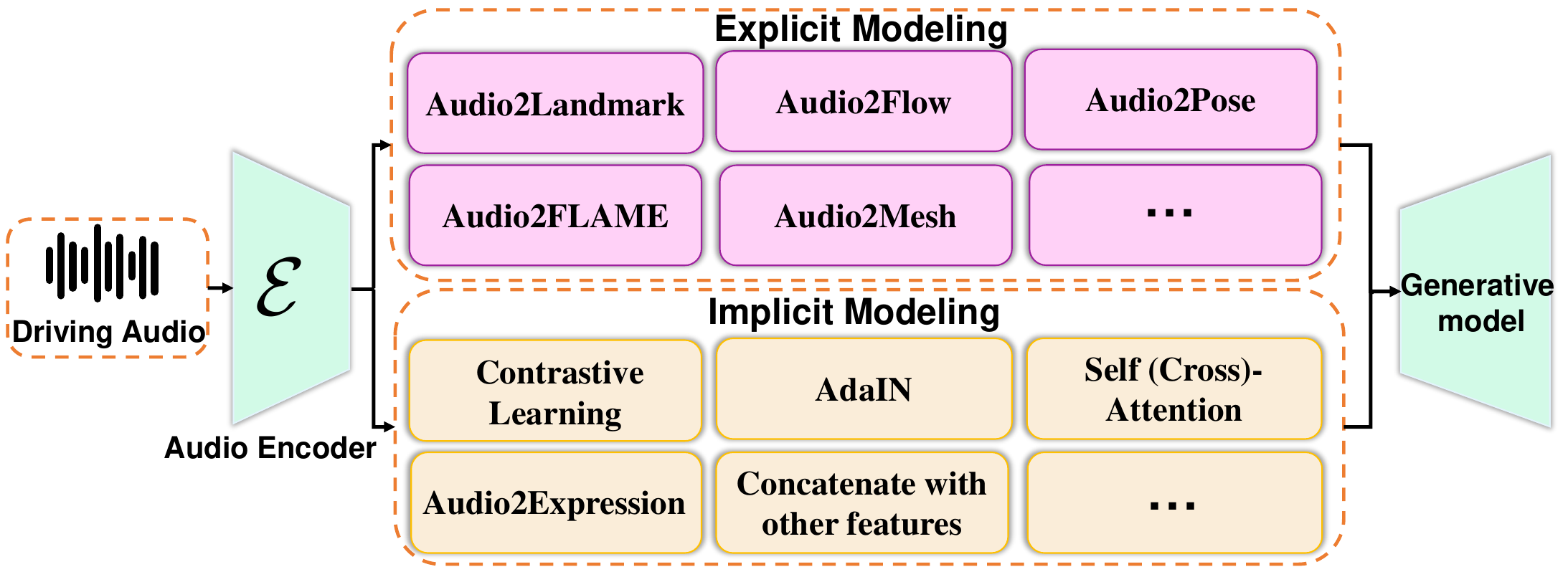}
    \vspace{-20px}
    \caption{Paradigm summary of audio-driven human motion video generation. 
    }
    \vspace{-15px}
    \label{fig:sumaudio}
\end{figure}

\new{This section focuses on the generation of human motion under audio-driven circumstances. By inputting audio and reference signals (e.g., reference images or videos), we deduce human motion videos corresponding to the audio signals.}
Modeling human motion from audio involves tackling several significant research challenges, such as accurately capturing lip movements, head poses, audio-driven holistic human motion, and producing fine-grained animations. 
Furthermore, we investigate and summarize the paradigms of audio used in human motion video generation, and provide clear instructions for the use of speech signals, as shown in Fig. \ref{fig:sumaudio}.

\noindent \textbf{Lip Synchronization.}
Early research primarily focused on lip motion. 
Wen et al.~\cite{wen2020photorealistic} map the audio to facial expression and re-render the synthetic video through a regional mask.
Wav2lip~\cite{wav2lip} synchronizes the given audio with the lip movements of characters in a video, achieving realistic lip-synchronization effects.
Chen et al.~\cite{chen2019hierarchical} employ a cascaded GANs approach to model the synchronization between speech and lip motion, effectively decoupling content-related representations from non-content-related signals in speech, thereby enhancing robustness to diverse facial shapes and viewpoints. 
Additionally, Cudeiro et al.~\cite{cudeiro2019capture} propose explicit modeling by 3D model structure for lip synchronization. 

\noindent \textbf{Head Pose Driving.}
Focusing solely on lip motions fails to produce a realistic visual effect of the speaker. 
Greenwood et al.~\cite{greenwood2017predicting} are pioneers in using a Bi-LSTM model~\cite{graves2005bidirectional} to predict head pose motions from audio. 
Zhou et al.~\cite{zhou2020makelttalk} explicitly model the speaker's head movements through facial landmarks, training a Transformer~\cite{vaswani2017attention} to capture long-term dependencies through adversarial learning, thereby generating natural head poses. 

Lu et al.~\cite{lu2021live} propose an autoregressive probabilistic model conditioned on historical head poses and speech representations to predict the motion distribution at the current time. 
Key head poses are sampled from this predicted probabilistic model. Their approach focuses on real-time audio-driven head pose video generation, capable of producing head pose videos at 30 frames per second.
Since audio alone does not provide cues for head pose and global head movements, directly inferring head movements from audio can lead to significant mismatches between the head pose and the motion of the edited head in the target video. 
Therefore, Ji et al.~\cite{ji2021audio} propose a target-adaptive facial synthesis technique that bridges the pose gap between the inferred landmarks and the target video portrait in 3D space. By employing a precisely crafted 3D perception keypoint alignment algorithm, 2D landmarks can be accurately projected onto the target video.

\begin{table*}[t]
\vspace{-15px}
\renewcommand{\arraystretch}{0.8} 
\centering
\caption{Comprehensive overview of audio-driven holistic human driving and fine-grained animation methods. 
}
\vspace{-5px}
\label{tab:fine}
\resizebox{0.935\textwidth}{!}{%
\begin{tabular}{|ccccccccc|}
\hline
\multicolumn{1}{|c|}{\textbf{Driving Region}}                                                                    & \multicolumn{1}{c|}{\textbf{\begin{tabular}[c]{@{}c@{}}Publication \\ Time\end{tabular}}} & \multicolumn{1}{c|}{\textbf{Paper}}                                               & \multicolumn{1}{c|}{\textbf{Input Signals}}                                 & \multicolumn{1}{c|}{\textbf{\begin{tabular}[c]{@{}c@{}}Motion \\ Representation\end{tabular}}}              & \multicolumn{1}{c|}{\textbf{Backbone}}                                                                                 & \multicolumn{1}{c|}{\textbf{\begin{tabular}[c]{@{}c@{}}Open \\ Source\end{tabular}}}                   & \multicolumn{1}{c|}{\textbf{\begin{tabular}[c]{@{}c@{}}\new{Real-Time}\\ \new{Inference}\end{tabular}}} & \textbf{Venue}
\\ \hline
\multicolumn{9}{|c|}{\cellcolor[HTML]{D1DADC}\rule{0pt}{2.5ex}\textbf{\begin{tabular}[c]{@{}c@{}} Fine-Grained Style and Emotion-Driven Animation \\\end{tabular}}}\\
\hline
\multicolumn{1}{|c|}{}                                                                                           & \multicolumn{1}{c|}{{\cellcolor[HTML]{ECF2F0}May. 19, 2021}}                                   & \multicolumn{1}{c|}{{\cellcolor[HTML]{ECF2F0}EVP \cite{ji2021audio}}}                     & \multicolumn{1}{c|}{{\cellcolor[HTML]{ECF2F0}\audiologo + \reallogo(video)}} & \multicolumn{1}{c|}{{\cellcolor[HTML]{ECF2F0}KeyPoint}}                                                       & \multicolumn{1}{c|}{{\cellcolor[HTML]{ECF2F0}\begin{tabular}[c]{@{}c@{}}ED\end{tabular}}}                 & \multicolumn{1}{c|}{{\cellcolor[HTML]{ECF2F0}\checkmark}}     & {\cellcolor[HTML]{ECF2F0}\new{×}} & \multicolumn{1}{|c|}{{\cellcolor[HTML]{ECF2F0}\begin{tabular}[c]{@{}c@{}}CVPR'21\end{tabular}}}   \\  
\multicolumn{1}{|c|}{}                                                                                           & \multicolumn{1}{c|}{Jun. 10, 2023}                                                           & \multicolumn{1}{c|}{StyleTalk \cite{ma2023styletalk}}                                   & \multicolumn{1}{c|}{\audiologo + \reallogo(video)}                         & \multicolumn{1}{c|}{\begin{tabular}[c]{@{}c@{}}3D-P\end{tabular}}                         & \multicolumn{1}{c|}{\begin{tabular}[c]{@{}c@{}}ED\end{tabular}}                                         & \multicolumn{1}{c|}{Only Inference}                         & \new{×} & \multicolumn{1}{|c|}{\begin{tabular}[c]{@{}c@{}}AAAI'23\end{tabular}}                           \\  
\multicolumn{1}{|c|}{}                                                                                           & \multicolumn{1}{c|}{{\cellcolor[HTML]{ECF2F0}Jan. 16, 2024}}                                   & \multicolumn{1}{c|}{{\cellcolor[HTML]{ECF2F0}Real3D-Portrait \cite{ye2024real3d}}}        & \multicolumn{1}{c|}{{\cellcolor[HTML]{ECF2F0}\audiologo + \reallogo}}        & \multicolumn{1}{c|}{{\cellcolor[HTML]{ECF2F0}\begin{tabular}[c]{@{}c@{}}3D-P\end{tabular}}} & \multicolumn{1}{c|}{{\cellcolor[HTML]{ECF2F0}\begin{tabular}[c]{@{}c@{}}ED\end{tabular}}}                 & \multicolumn{1}{c|}{{\cellcolor[HTML]{ECF2F0}\checkmark}} & {\cellcolor[HTML]{ECF2F0}\new{×}} & \multicolumn{1}{|c|}{{\cellcolor[HTML]{ECF2F0}\begin{tabular}[c]{@{}c@{}}ICLR'24\end{tabular}}}                                              \\  
\multicolumn{1}{|c|}{}                                                                                           & \multicolumn{1}{c|}{Dec. 15, 2023}                                                           & \multicolumn{1}{c|}{DreamTalk \cite{ma2023dreamtalk}}                                   & \multicolumn{1}{c|}{\audiologo + \reallogo(video)}                         & \multicolumn{1}{c|}{\begin{tabular}[c]{@{}c@{}}3D-P\end{tabular}}                         & \multicolumn{1}{c|}{\begin{tabular}[c]{@{}c@{}}DM\end{tabular}}                                        & \multicolumn{1}{c|}{Only Inference}                         & \new{×} & \multicolumn{1}{|c|}{arXiv}                                                                          \\  
\multicolumn{1}{|c|}{}                                                                                           & \multicolumn{1}{c|}{{\cellcolor[HTML]{ECF2F0}Jun. 4, 2024}}                                   & \multicolumn{1}{c|}{{\cellcolor[HTML]{ECF2F0}V-Express \cite{wang2024v}}}                 & \multicolumn{1}{c|}{{\cellcolor[HTML]{ECF2F0}\audiologo + \reallogo}}        & \multicolumn{1}{c|}{{\cellcolor[HTML]{ECF2F0}KeyPoint}}                                                       & \multicolumn{1}{c|}{{\cellcolor[HTML]{ECF2F0}\begin{tabular}[c]{@{}c@{}}DM\end{tabular}}}                & \multicolumn{1}{c|}{{\cellcolor[HTML]{ECF2F0}\checkmark}}     & {\cellcolor[HTML]{ECF2F0}\new{×}} & \multicolumn{1}{|c|}{{\cellcolor[HTML]{ECF2F0}arXiv}}                                                  \\  
\multicolumn{1}{|c|}{}                                                                                           & \multicolumn{1}{c|}{Jul, 21, 2021}                                                           & \multicolumn{1}{c|}{Eskimez et al. \cite{eskimez2021speech}}                            & \multicolumn{1}{c|}{\audiologo + \reallogo}                                & \multicolumn{1}{c|}{Latent}                                                                                 & \multicolumn{1}{c|}{GAN}                                                                                               & \multicolumn{1}{c|}{\checkmark}                             & \new{×} & \multicolumn{1}{|c|}{\begin{tabular}[c]{@{}c@{}}TMM'21\end{tabular}}                     \\  
\multicolumn{1}{|c|}{}                                                                                           & \multicolumn{1}{c|}{{\cellcolor[HTML]{ECF2F0}Nov. 22, 2022}}                                   & \multicolumn{1}{c|}{{\cellcolor[HTML]{ECF2F0}SadTalker \cite{zhang2023sadtalker}}}        & \multicolumn{1}{c|}{{\cellcolor[HTML]{ECF2F0}\audiologo + \reallogo}}        & \multicolumn{1}{c|}{{\cellcolor[HTML]{ECF2F0}Latent + 3D-P}}                                                         & \multicolumn{1}{c|}{{\cellcolor[HTML]{ECF2F0}\begin{tabular}[c]{@{}c@{}}GAN\end{tabular}}}                & \multicolumn{1}{c|}{{\cellcolor[HTML]{ECF2F0}Only Inference}} & {\cellcolor[HTML]{ECF2F0}\new{×}} & \multicolumn{1}{|c|}{{\cellcolor[HTML]{ECF2F0}\begin{tabular}[c]{@{}c@{}}CVPR'23\end{tabular}}}    \\  
\multicolumn{1}{|c|}{}                                                                                           & \multicolumn{1}{c|}{Nov. 28, 2022}                                                           & \multicolumn{1}{c|}{Bai et al. \cite{bai2023high}}                                      & \multicolumn{1}{c|}{\audiologo + \reallogo}                                & \multicolumn{1}{c|}{\begin{tabular}[c]{@{}c@{}}3D-P\end{tabular}}                         & \multicolumn{1}{c|}{GAN}                                                                                               & \multicolumn{1}{c|}{\checkmark}                             & \new{×} & \multicolumn{1}{|c|}{\begin{tabular}[c]{@{}c@{}}CVPR'23\end{tabular}}                            \\  
\multicolumn{1}{|c|}{}                                                                                           & \multicolumn{1}{c|}{{\cellcolor[HTML]{ECF2F0}May. 9, 2023}}                                   & \multicolumn{1}{c|}{{\cellcolor[HTML]{ECF2F0}StyleSync \cite{guan2023stylesync}}}         & \multicolumn{1}{c|}{{\cellcolor[HTML]{ECF2F0}\audiologo + \reallogo(video)}} & \multicolumn{1}{c|}{{\cellcolor[HTML]{ECF2F0}Latent}}                                                         & \multicolumn{1}{c|}{{\cellcolor[HTML]{ECF2F0}GAN}}                                                                       & \multicolumn{1}{c|}{{\cellcolor[HTML]{ECF2F0}×}}              & {\cellcolor[HTML]{ECF2F0}\new{×}} & \multicolumn{1}{|c|}{{\cellcolor[HTML]{ECF2F0}\begin{tabular}[c]{@{}c@{}}CVPR'23\end{tabular}}}    \\  
\multicolumn{1}{|c|}{}                                                                                           & \multicolumn{1}{c|}{Feb. 27, 2024}                                                           & \multicolumn{1}{c|}{EMO \cite{tian2024emo}}                                             & \multicolumn{1}{c|}{\audiologo + \reallogo}                                & \multicolumn{1}{c|}{Latent}                                                                                 & \multicolumn{1}{c|}{\begin{tabular}[c]{@{}c@{}}DM\end{tabular}}                                        & \multicolumn{1}{c|}{×}                                      & \new{×} & \multicolumn{1}{|c|}{\new{ECCV'24}}                                                                          \\  
\multicolumn{1}{|c|}{}                                                                                           & \multicolumn{1}{c|}{{\cellcolor[HTML]{ECF2F0}Mar. 4, 2024}}                                   & \multicolumn{1}{c|}{{\cellcolor[HTML]{ECF2F0}FaceChain-ImagineID \cite{xu2024facechain}}} & \multicolumn{1}{c|}{{\cellcolor[HTML]{ECF2F0}\audiologo + \reallogo}}        & \multicolumn{1}{c|}{{\cellcolor[HTML]{ECF2F0}Latent}}                                                         & \multicolumn{1}{c|}{{\cellcolor[HTML]{ECF2F0}\begin{tabular}[c]{@{}c@{}}DM\end{tabular}}}                & \multicolumn{1}{c|}{{\cellcolor[HTML]{ECF2F0}\checkmark}} & {\cellcolor[HTML]{ECF2F0}\new{×}} & \multicolumn{1}{|c|}{{\cellcolor[HTML]{ECF2F0}\begin{tabular}[c]{@{}c@{}}CVPR'24\end{tabular}}}    \\  
\multicolumn{1}{|c|}{}                                                                                           & \multicolumn{1}{c|}{Apr. 29, 2024}                                                           & \multicolumn{1}{c|}{EMOPortraits \cite{drobyshev2024emoportraits}}                      & \multicolumn{1}{c|}{\audiologo + \reallogo}                                & \multicolumn{1}{c|}{Latent}                                                                                 & \multicolumn{1}{c|}{GAN}                                                                                               & \multicolumn{1}{c|}{×}                                      & \new{×} & \multicolumn{1}{|c|}{\begin{tabular}[c]{@{}c@{}}CVPR'24\end{tabular}}                            \\  
\multicolumn{1}{|c|}{\multirow{-17}{*}{\begin{tabular}[c]{@{}c@{}}\partlogo\\ Part\\ (Face)\end{tabular}}}                                                                                           & \multicolumn{1}{c|}{{\cellcolor[HTML]{ECF2F0}May. 15, 2024}}                                   & \multicolumn{1}{c|}{{\cellcolor[HTML]{ECF2F0}SPEAK \cite{cai2024listen}}}                 & \multicolumn{1}{c|}{{\cellcolor[HTML]{ECF2F0}\audiologo + \reallogo}}        & \multicolumn{1}{c|}{{\cellcolor[HTML]{ECF2F0}Latent}}                                                         & \multicolumn{1}{c|}{{\cellcolor[HTML]{ECF2F0}GAN}}                                                                       & \multicolumn{1}{c|}{{\cellcolor[HTML]{ECF2F0}×}}              & {\cellcolor[HTML]{ECF2F0}\new{×}} & \multicolumn{1}{|c|}{{\cellcolor[HTML]{ECF2F0}arXiv}}                                                  \\  
\multicolumn{1}{|c|}{}                                                                                           & \multicolumn{1}{c|}{Jun. 16, 2024}                                                           & \multicolumn{1}{c|}{Hallo \cite{xu2024hallo}}                                           & \multicolumn{1}{c|}{\audiologo + \reallogo}                                & \multicolumn{1}{c|}{Latent}                                                                                 & \multicolumn{1}{c|}{\begin{tabular}[c]{@{}c@{}}DM\end{tabular}}                                        & \multicolumn{1}{c|}{\checkmark}                         & \new{×} & \multicolumn{1}{|c|}{arXiv}                                                                          \\  
\multicolumn{1}{|c|}{}                                                                                           & \multicolumn{1}{c|}{{\cellcolor[HTML]{ECF2F0}Apr. 16, 2024}}                                   & \multicolumn{1}{c|}{{\cellcolor[HTML]{ECF2F0}VASA-1 \cite{xu2024vasa}}}                   & \multicolumn{1}{c|}{{\cellcolor[HTML]{ECF2F0}\audiologo + \reallogo(video)}} & \multicolumn{1}{c|}{{\cellcolor[HTML]{ECF2F0}Latent}}                                                         & \multicolumn{1}{c|}{{\cellcolor[HTML]{ECF2F0}\begin{tabular}[c]{@{}c@{}}DiT\end{tabular}}} & \multicolumn{1}{c|}{{\cellcolor[HTML]{ECF2F0}×}}              & {\cellcolor[HTML]{ECF2F0}\realtimelogo \new{40 FPS}} & \multicolumn{1}{|c|}{{\cellcolor[HTML]{ECF2F0}arXiv}}                                                  \\  
\multicolumn{1}{|c|}{}                                                                                           & \multicolumn{1}{c|}{Jan. 5, 2023}                                                           & \multicolumn{1}{c|}{chen et al. \cite{chen2023expressive}}                              & \multicolumn{1}{c|}{\audiologo}                                            & \multicolumn{1}{c|}{Latent}                                                                                 & \multicolumn{1}{c|}{\begin{tabular}[c]{@{}c@{}}ED\end{tabular}}                                         & \multicolumn{1}{c|}{\checkmark}                             & \new{×} & \multicolumn{1}{|c|}{\begin{tabular}[c]{@{}c@{}}ICMEW'23\end{tabular}}                           \\  
\multicolumn{1}{|c|}{}       & \multicolumn{1}{c|}{{\cellcolor[HTML]{ECF2F0}Jan. 28, 2024}}                                   & \multicolumn{1}{c|}{{\cellcolor[HTML]{ECF2F0}Media2Face \cite{zhao2024media2face}}}       & \multicolumn{1}{c|}{{\cellcolor[HTML]{ECF2F0}\audiologo}}                    & \multicolumn{1}{c|}{{\cellcolor[HTML]{ECF2F0}Latent}}                                                         & \multicolumn{1}{c|}{{\cellcolor[HTML]{ECF2F0}\begin{tabular}[c]{@{}c@{}}DM\end{tabular}}}                & \multicolumn{1}{c|}{{\cellcolor[HTML]{ECF2F0}×}}              & {\cellcolor[HTML]{ECF2F0}\new{×}} & \multicolumn{1}{|c|}{{\cellcolor[HTML]{ECF2F0}\begin{tabular}[c]{@{}c@{}}SIG'24\end{tabular}}}                                                  \\ \hline
\multicolumn{9}{|c|}{\cellcolor[HTML]{D1DADC}\rule{0pt}{2.5ex}\textbf{\begin{tabular}[c]{@{}c@{}} Audio-driven Holistic Human Driving \\\end{tabular}}}                                                                                                                                                                                                                                                                                                                                                                                                                                                                                                                                                                                                                                            \\ \hline
\multicolumn{1}{|c|}{}                                                                                           & \multicolumn{1}{c|}{Mar. 16, 2024}                                                           & \multicolumn{1}{c|}{VLOGGER \cite{corona2024vlogger}}                                   & \multicolumn{1}{c|}{\audiologo + \reallogo}                                & \multicolumn{1}{c|}{\begin{tabular}[c]{@{}c@{}}3D-P\end{tabular}}                         & \multicolumn{1}{c|}{\begin{tabular}[c]{@{}c@{}}DM\end{tabular}}                                        & \multicolumn{1}{c|}{×}                                      & \new{×} & \multicolumn{1}{|c|}{arXiv}                                                                          \\  
\multicolumn{1}{|c|}{\multirow{-3}{*}{\begin{tabular}[c]{@{}c@{}}\quad \\ \holisticlogo\\ Holistic\\ Human\end{tabular}}}      

& \multicolumn{1}{c|}{{\cellcolor[HTML]{ECF2F0}Dec. 15, 2022}} &
\multicolumn{1}{c|}{{\cellcolor[HTML]{ECF2F0}ANGIE \cite{liu2022audio}}} &
\multicolumn{1}{c|}{{\cellcolor[HTML]{ECF2F0}\audiologo + \reallogo}} &
\multicolumn{1}{c|}{{\cellcolor[HTML]{ECF2F0}Latent}} &
\multicolumn{1}{c|}{{\cellcolor[HTML]{ECF2F0}\begin{tabular}[c]{@{}c@{}}ED\end{tabular}}} &
\multicolumn{1}{c|}{{\cellcolor[HTML]{ECF2F0}×}} & {\cellcolor[HTML]{ECF2F0}\new{×}} & \multicolumn{1}{|c|}{{\cellcolor[HTML]{ECF2F0}\begin{tabular}{@{}c@{}}NeurIPS'22\end{tabular}}} \\ 

\multicolumn{1}{|c|}{} & \multicolumn{1}{c|}{May. 15, 2024}                                                           & \multicolumn{1}{c|}{Dance-Any-Beat \cite{wang2024dance}}                                & \multicolumn{1}{c|}{\audiologo + \reallogo}                                & \multicolumn{1}{c|}{Region}                                                                                 & \multicolumn{1}{c|}{\begin{tabular}[c]{@{}c@{}}DM\end{tabular}}                                        & \multicolumn{1}{c|}{×}                                      & \new{×} & \multicolumn{1}{|c|}{arXiv}                                                                          \\ \hline
\multicolumn{9}{|l|}{\reallogo: Reference Real Images;  \audiologo: Audio; \realtimelogo: \new{Real-Time Support 
(Estimated FPS based on NVIDIA RTX 4090 GPU)}.}                                                                                                                                                                                                                                                                                                                                                                                                                                                                                                                                                                                                                                                                                                \\ \hline
\end{tabular}%
}
\vspace{-10px}
\end{table*}

For more controllability, recent studies~\cite{wei2024aniportrait, yu2024make, liang2024emotional,chen2023vast} have proposed a two-stage framework: audio-to-landmark and landmark-to-video. 
In the first stage, audio features are extracted using wav2vec~\cite{baevski2020wav2vec} and converted into 2D facial landmark sequences. 
In the second stage, a diffusion model and motion modules transform these sequences into temporally consistent and realistic portrait animations. 
Similar two-stage approaches~\cite{ji2024realtalk, ye2023r2} are employed to refine facial expression prediction and achieve high-fidelity facial rendering.
Furthermore, Meng et al.~\cite{meng2024comprehensive} elaborately summarize the latest advancements regarding the talking head task. 
To provide a broader perspective, Table~\ref{tab:talking head} offers a comprehensive overview of the diverse works in the field of head pose driving.

\noindent \textbf{Audio-Driven Holistic Human Driving.}
Previous audio-driven human motion video generation methods have typically focused on facial expressions and lip synchronization, neglecting the generation of head, upper body, and hand movements synchronized with audio. 

This limitation reduces the effectiveness of these videos in conveying richer human communication, as the absence of body language and gestures makes the synthesized videos appear less natural and realistic.
Vlogger~\cite{corona2024vlogger} addresses this gap by generating realistic and temporally coherent videos based on a single input image and audio sample. It employs a Transformer-based network that inputs Mel spectrograms to predict a series of 3D facial expressions and body pose parameters, capturing the complex mapping relationship between audio signals and human movements. This method not only includes head movements, gaze, and lip motions but also generates upper body and hand gestures, thus advancing audio-driven synthesis technology to a new level.

Similarly, some work has focused on holistic motion video generation using only speech to provide motion information. 
ANGIE~\cite{liu2022audio} employs a unified framework to generate speaker image sequences driven by speech audio. The key insight is that co-speech gestures can be decomposed into common motion patterns and subtle rhythmic dynamics. 
Specifically, Dance Any Beat~\cite{wang2024dance} pioneers the task of music-driven dance video generation. Their method is based on the LFDM~\cite{ni2023conditional} framework, which generates corresponding dance movements from music using just a single human photo.

\noindent \textbf{Fine-Grained Style and Emotion-Driven Animation.}
Controlling facial actions based on emotions derived from audio presents the primary challenge of extracting emotions, as emotional information is intricately entangled with other factors like speech content. 
Ji et al.~\cite{ji2021audio} address this by proposing a cross-reconstruction emotion disentanglement technique for audio, which extracts two independent latent spaces: a duration-independent space that encodes emotion without regard to content, and a duration-dependent space that encodes the speech content of the audio. 

Additionally, 
DreamTalk~\cite{ma2023dreamtalk} not only manages the generation of stylized facial animations but also enables fine-grained control over facial expressions through style-aware lip experts and style predictors. 
Recently, 
emotion and speech are frequently used simultaneously as conditions~\cite{eskimez2021speech, eskimez2020end, xu2023high, guan2023stylesync, xu2024facechain}, providing implicit motion representations for generating human motion videos.
To address the complexity of integrating audio with diffusion models due to the ambiguity in mapping audio to facial expressions, 
EMO~\cite{tian2024emo} introduces stable control mechanisms, including a speed controller and a face region controller, to enhance generation stability. 
Furthermore, VASA-1~\cite{xu2024vasa} integrates several additional signals to make the generative modeling more manageable and improve the controllability of the generation process.

A comprehensive overview of these innovative approaches is provided in Table~\ref{tab:fine}.  
Indeed, addressing the complexity of audio-driven human motion video generation faces these challenges: 

\newpage
\begin{itemize}



\item \new{
Recent research has advanced lip synchronization~\cite{zhong2024stylepreservinglipsyncaudioaware,bigioi2024speech} and head motion synthesis~\cite{wei2024aniportrait,liu2024anitalker,chen2024echomimic,zhong2024stylepreservinglipsyncaudioaware}, incorporating emotion control~\cite{tian2024emo,drobyshev2024emoportraits,xu2024hallo}. 
However, a major challenge remains: developing a unified framework that integrates lip synchronization, audio-driven head motion synthesis, and rhythmic gesture generation~\cite{corona2024vlogger}, while ensuring stable backgrounds and preserving holistic body details.
}

\item \new{Diffusion-based approaches~\cite{xu2024hallo,drobyshev2024emoportraits,ma2023dreamtalk,wang2024v} face significant challenges in achieving real-time performance, particularly for audio-driven holistic human body motion video generation. While lightweight neural architectures~\cite{cho2024gaussiantalker,yu2024gaussiantalker} offer potential solutions, achieving low-latency inference in diffusion models remains a critical research focus.}





\end{itemize}

\noindent \textbf{Multilingual Video Dubbing.}
Multilingual video dubbing is another intriguing task in human motion video generation, where videos are translated from one language to another. 
The speech content in the source language is transcribed into text, translated, and then automatically synthesized into speech in the target language, retaining the original speaker's voice. 
The vision content is aligned with the translated audio by synthesizing the speaker's lip movements, creating a seamless audio-visual experience in the target language.
Yang et al.~\cite{yang2020largescalemultilingualaudiovisual} are among the first to tackle this task. 
Bigioi et al.~\cite{bigioi2023multilingual} further elaborate on the challenges in this field, highlighting the complexities of achieving realism, cross-lingual adaptability, and overcoming limitations in data diversity and generalization, which continue to drive research and development efforts.

\vspace{-7px}
\section{Refinement and Output}
\label{Sec:refinement}

The domain of human motion video generation has recently attracted considerable attention due to its extensive potential applications. However, the current generation frameworks are still in their nascent stages and exhibit limited control capabilities. To address these challenges, advanced refinement strategies have been introduced. 
This section details both the refinement phase (Step 4) and the output phase (Step 5).

\vspace{-10px}
\noindent \subsection{Refinement}
Refinement processes are crucial for enhancing the output of generation models can be broadly classified into two categories: part specific refinement and general refinement. Part-specific refinement targets particular, whereas general refinement aims to enhance overall video quality.

\noindent \textbf{Part Specific Refinement}
involves targeted enhancements of specific body parts, such as the mouth, eyes, teeth, and hands, which are commonly affected by generation errors. This refinement is essential in addressing the limitations of generation models in these delicate regions.
Targeted restoration methods, falling under this broader strategy, are utilized to correct inaccuracies in these specific regions.
These methods~\cite{guo2024liveportrait, zhang2022unsupervised} include the development of specialized loss functions which are tailored to targeted regions during the training phase, and the application of pre-trained networks for post-processing improvements. For instance, the MimicMotion~\cite{zhang2024mimicmotion} exemplifies the use of advanced pose guidance mechanisms and design a hand region enhance method to enhance the precision of subtle movements, minimize distortions, and improve overall motion fidelity. In addition, post-processing pipelines often incorporate tools like Codeformer~\cite{zhou2022towards} or the approach by Feng et al.~\cite{24kalman} to eliminate facial artifacts in the outputs.

\noindent \textbf{General Refinement}
techniques include super-resolution, frame rate enhancement, and denoising networks. These collectively work to enhance the resolution, frame rate, and overall clarity of videos, thereby significantly improving the  quality of viewing.

\vspace{-10px}
\subsection{Output}
Real-time generation of human motion videos remains a challenging and relatively unexplored domain.
While GAN-based methods such as those developed by Guo et al.~\cite{guo2024liveportrait} and Jiang et al.~\cite{jiang2024mobileportrait} demonstrate some capacity for real-time performance, primarily in applications like talking head and portrait animations, they face issues with training instability and lower video quality.
The focus on diffusion models has increased due to their exceptional capability for generating high-quality videos. Nevertheless, the high computational costs for training and inference present significant challenges for real-time applications. Although these models hold great promise, research on cost optimization and inference acceleration is still in its nascent stages, posing a considerable barrier to their practical, real-time use.
Emerging works such as Kodaira et al.~\cite{kodaira2023streamdiffusion} and Liang et al.~\cite{liang2024looking} explore innovative real-time video editing techniques using stream-based diffusion, presenting a promising direction for integrating these methods with human motion video generation. Further advancements in model distillation, as seen in the work of Sauer et al.~\cite{sauer2023adversarial} and Zhai et al.~\cite{zhai2024motion}, are anticipated to enhance model sampling speeds significantly, paving the way for real-time video generation capabilities in the near future.

\vspace{-8px}
\section{Evaluation}
\label{Sec:Evaluation}

\vspace{-5px}
\subsection{Evaluation Metrics}

\begin{figure}[t]
    \centering
    \includegraphics[width=0.9\linewidth]{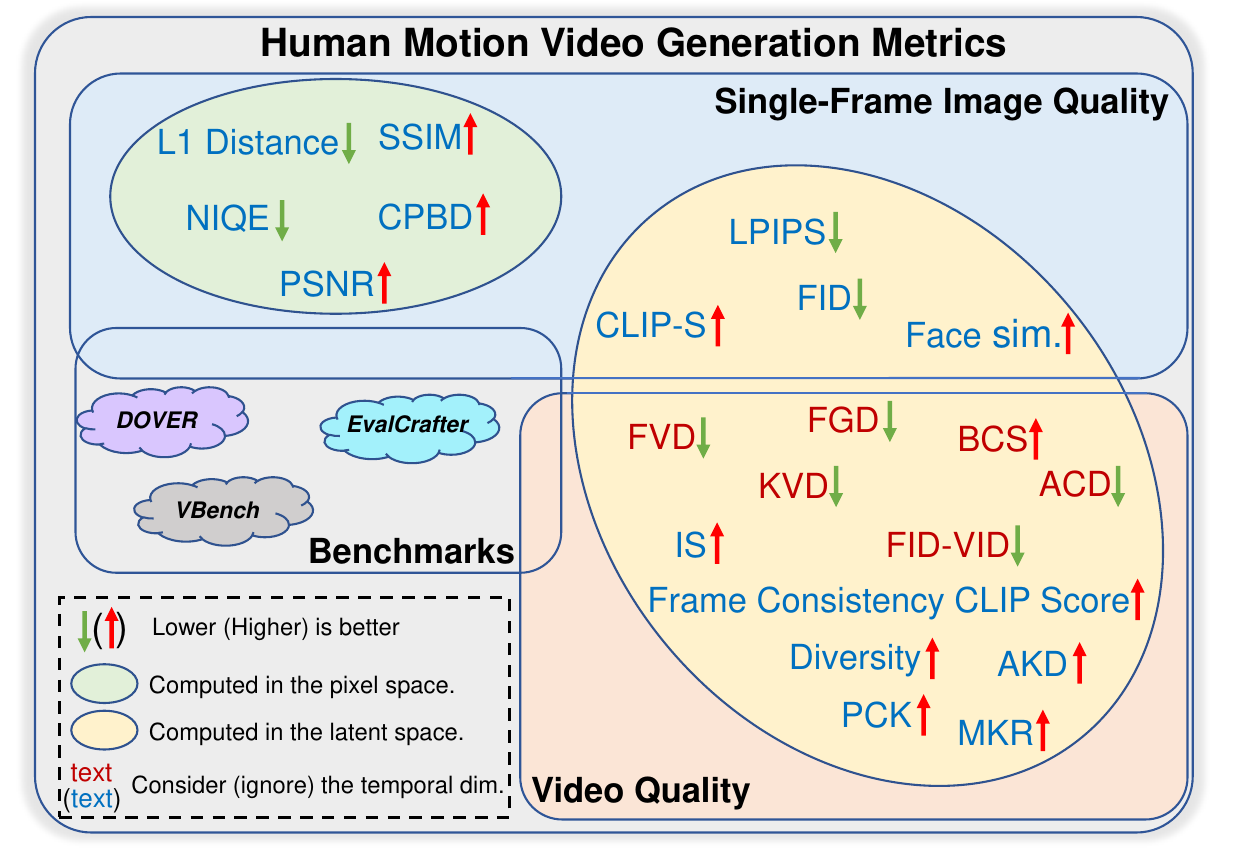}
    \vspace{-10px}
    \caption{Overview of common metrics in human motion video generation. 
    }
    \label{fig:metrics}
    \vspace{-10px}
\end{figure}

Appropriate evaluation metrics are essential for comparing different approaches and advancing the field. However, a uniform evaluation system for generated human motion videos is still lacking. In this section, we summarize the different aspects of commonly used evaluation metrics (Fig. \ref{fig:metrics}).

\subsubsection{Single-Frame Image Quality}\label{frame}
In the video generation task, the goal of measuring the quality of each generated frame is to ensure that the video not only appears coherent and smooth as a whole but also that each frame maintains high quality.
Common metrics include L1 Distance, Fr\'{e}chet Inception Distance (FID)~\cite{heusel2017gans, Seitzer2020FID}, Structural Similarity Index Measure (SSIM)~\cite{wang2004image}, Peak Signal-to-Noise Ratio (PSNR)~\cite{hore2010image}, and Learned Perceptual Image Patch Similarity (LPIPS) ~\cite{zhang2018unreasonable}.  
Additional metrics include:
CLIP-S~\cite{radford2021learning}, Face Similarity~\cite{chang2024magicpose}, CPBD~\cite{narvekar2009sharpness, narvekar2011blur}, and NIQE~\cite{Mittal2013CompletelyBlind}.

\subsubsection{Video Quality Assessment}\label{video}
\new{Evaluation of} examining temporal consistency, motion coherence, and overall visual appeal. 
Key metrics include: KVD~\cite{FVDKVD}, FVD~\cite{FVDKVD}, FGD~\cite{Yoon2020Speech}, and ACD~\cite{Tulyakov2018MoCoGAN}.
\subsubsection{Video Characteristics Assessment}
Frame Consistency CLIP Score~\cite{Esser2023VideoSynthesis}, IS~\cite{hinz2021improved}, Diversity~\cite{liu2022learning}, BCS~\cite{
li2021audio2gestures}, and Pose Accuracy~\cite{6380498,siarohin2019first,zhao2022thinplate}.
\subsubsection{Video Evaluators and Benchmarks}
Comprehensive tools for assessing the quality of generated videos, including DOVER~\cite{Wu2023VQAssessment}, VBench~\cite{huang2024vbench}, and EvalCrafter~\cite{liu2024evalcrafter}.
\new{
Although comprehensive video evaluators and benchmarks offer systematic frameworks, researchers mainly rely on conventional metrics (such as SSIM~\cite{wang2004image} and FID-VID~\cite{heusel2017gans, Seitzer2020FID}) because of the implementation complexity and computational cost of sophisticated evaluation frameworks.
}

\newR{
\vspace{-10px}
\subsubsection{LLM Planner Evaluation}
The LLM planner evaluation focuses on two key aspects: }

\begin{itemize}
    \item \newR{The evaluation of the ability to make appropriate responses based on the input signals from LLMs (e.g., Recall@k and M.Rank~\cite{geng2023affective,wang2023agentavatar,wang2024instructavatar}).}
    \item \newR{The evaluation of the quality of instructions from LLMs (e.g., CLIP Score~\cite{wang2024instructavatar}).}
\end{itemize}
\newR{For example, Geng et al.~\cite{geng2023affective} employ Recall@k, which is defined as the percentage of speakers in
the evaluation set for which the correct ground-truth listener
appears among the top-k predictions, together with the median retrieval rank~(M. Rank) of the ground-truth listener as key evaluation metrics.
Additionally, InstructAvatar \cite{wang2024instructavatar} employs CLIP Score to evaluate how well the LLM's instructions are followed by the generation models. We provide a simple evaluation experiment for the LLM planner in Appendix H.
}

\vspace{-10px}
\subsection{Comparative Analysis}

\begin{figure}[!t]
    \centering
    \resizebox{1\linewidth}{!}{  
    \begin{minipage}{\linewidth}  
        \vspace{-10px}
        \subfloat[\small SSIM (↑)]{\includegraphics[width=0.5\linewidth]{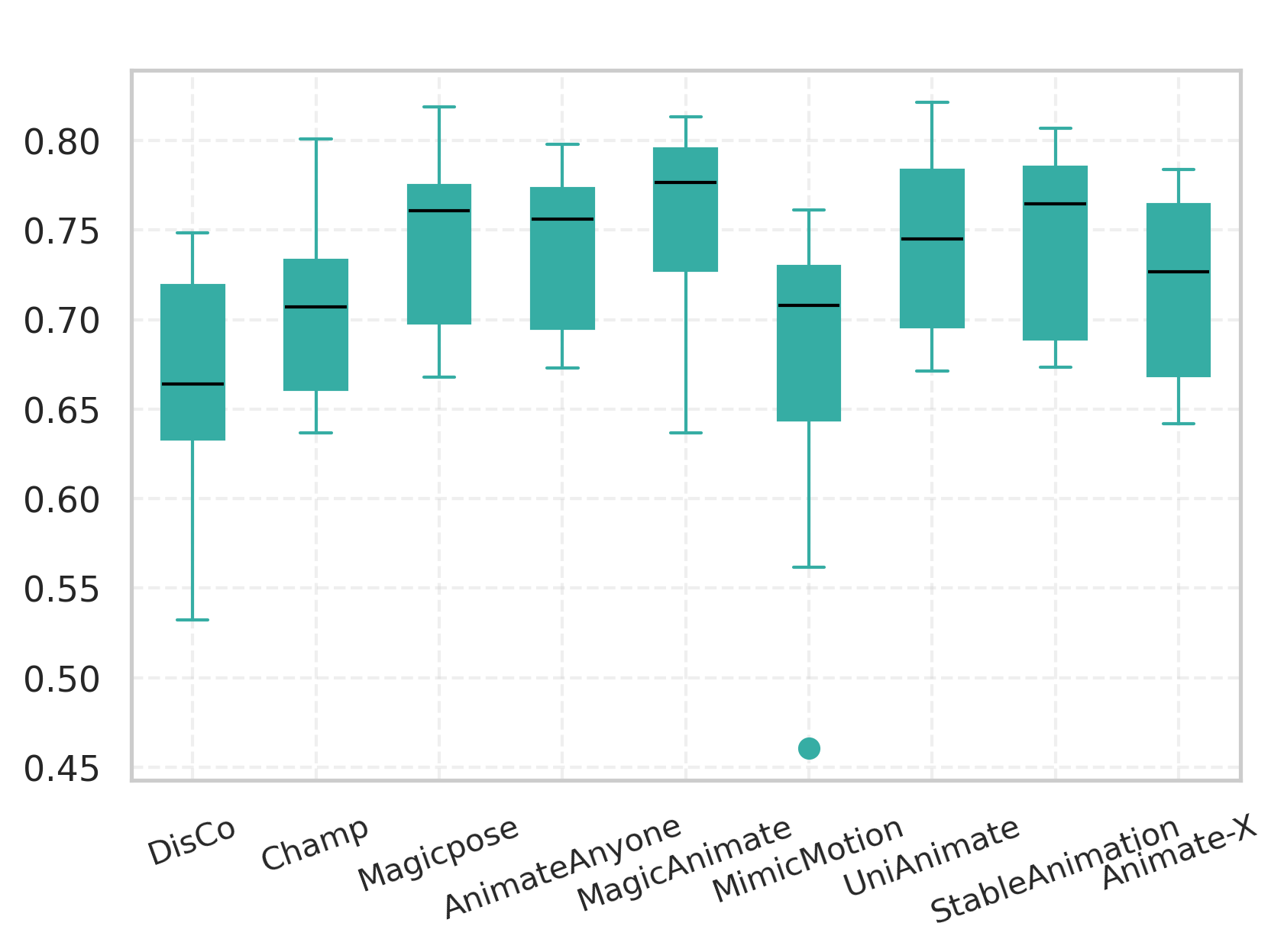}}
        \hfill
        \subfloat[\small PSNR (↑)]{\includegraphics[width=0.5\linewidth]{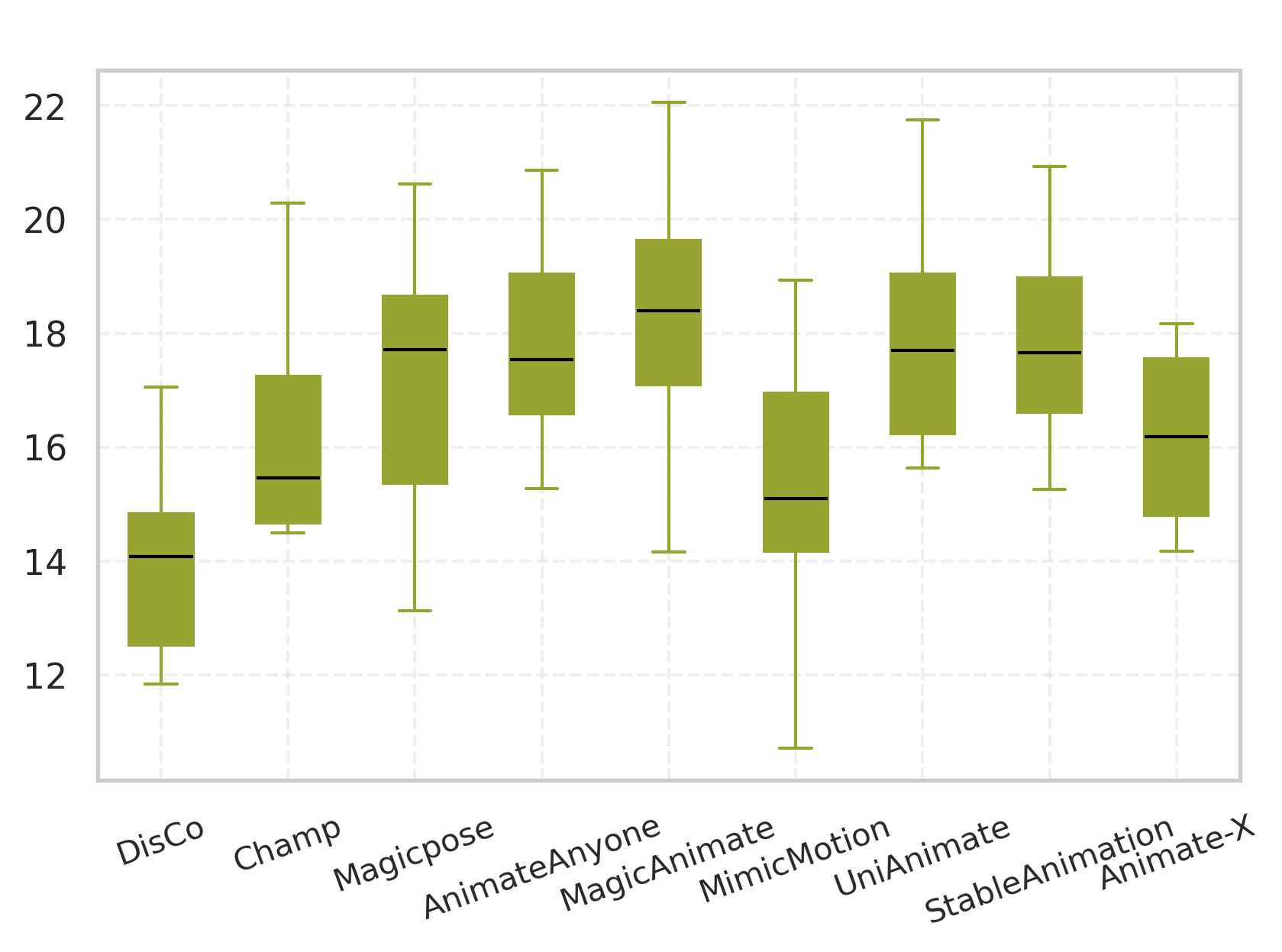}}
        \hfill
        \vspace{-10px}
        \subfloat[\small LPIPS (↓)]{\includegraphics[width=0.5\linewidth]{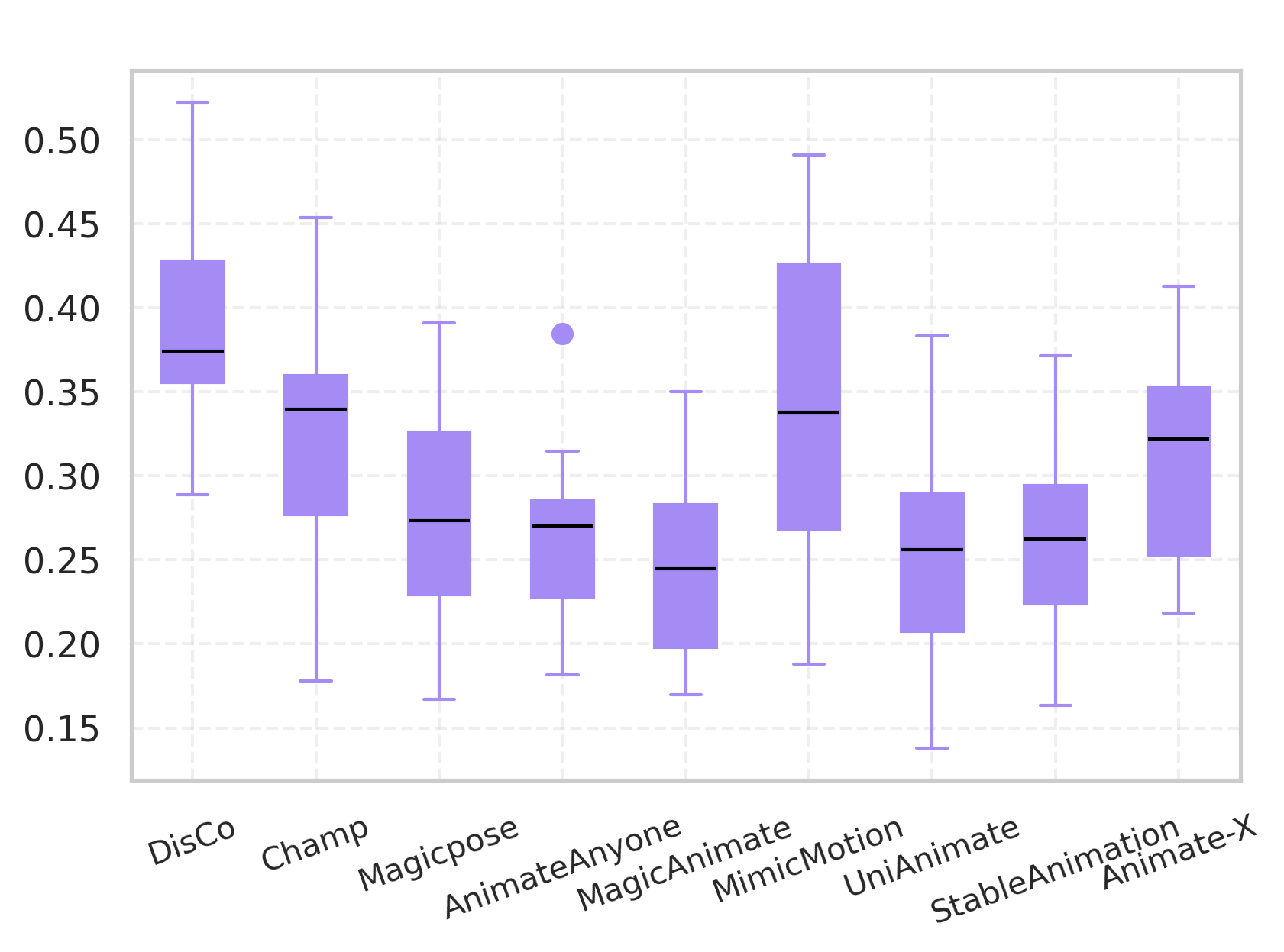}}
        \subfloat[\small L1 (↓)]{\includegraphics[width=0.5\linewidth]{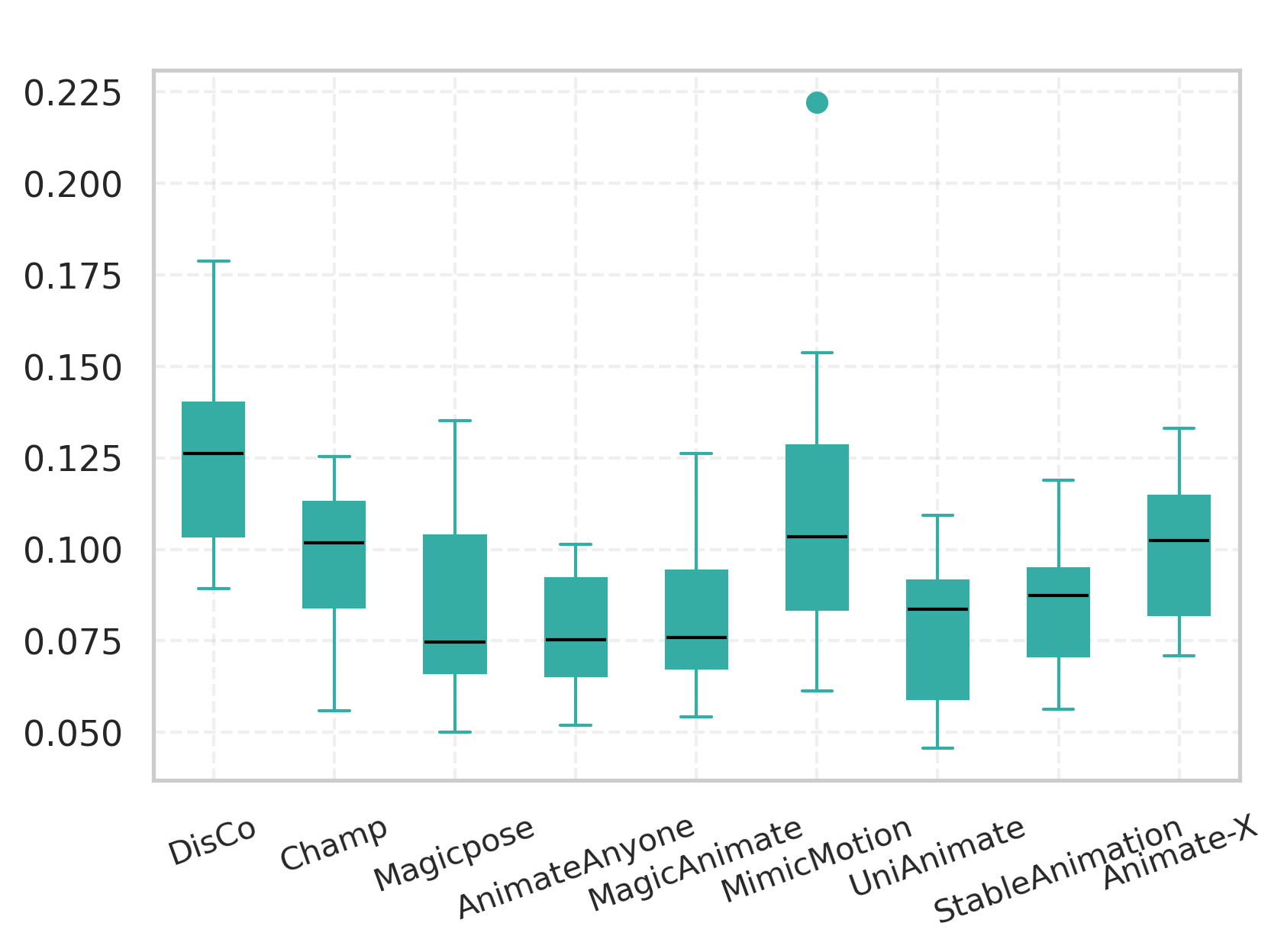}}
        \hfill
    \end{minipage}
    }
    \caption{
    Quantitative performances of five open-source methods across different matrices for pose-guided dance video 
generation.
    }
    \vspace{-10px}
    \label{fig:evaluation1}
\end{figure}
\begin{figure*}[t]
    \centering
    \includegraphics[width=0.975\linewidth]{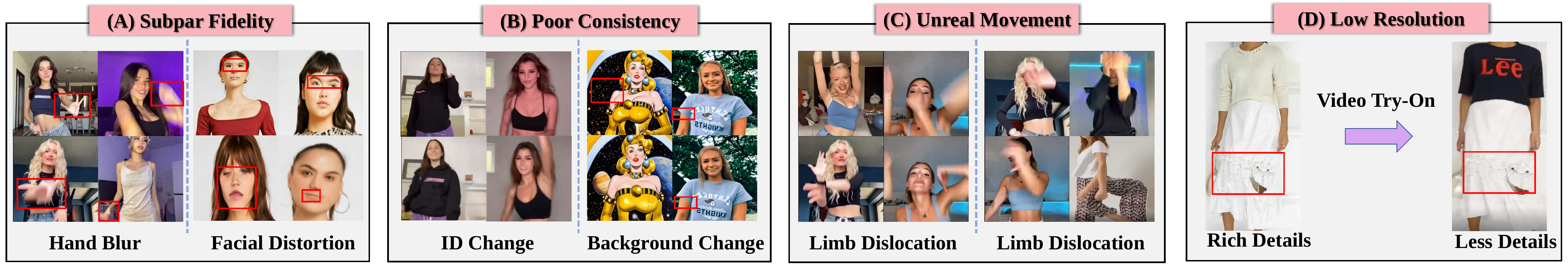}
    \caption{
    Main challenges in human motion video generation: subpar fidelity with examples like hand blur and facial distortion, poor consistency with identity and background changes, unrealistic movements, and low resolution.
    }
    \label{fig:challenges}
    \vspace{-8px}
\end{figure*}

This section centers on the task of pose-guided dance video generation. 
Through quantitative comparative experiments on the TikTok test dataset with 10 videos~\cite{Jafarian_2021_CVPR_TikTok}, we benchmark \new{nine} related open-source methods: Disco~\cite{wang2024disco}, Champ~\cite{zhu2024champ}, MagicPose~\cite{chang2024magicpose}, Animate Anyone~\cite{hu2024animate}, \new{MagicAnimate~\cite{xu2024magicanimate}, MimicMotion~\cite{zhang2024mimicmotion}, UniAnimate~\cite{wang2024unianimate}, StableAnimator~\cite{tu2024stableanimator}, and Animate-X~\cite{tan2024animate}.} 
Since Animate Anyone~\cite{hu2024animate} has not been officially open-sourced, the version used in our experiments is an unofficial implementation\footnote{https://github.com/MooreThreads/Moore-AnimateAnyone}.
\new{We assess the performances of the nine methods with six most commonly used evaluation metrics}: L1, PSNR~\cite{hore2010image}, SSIM~\cite{wang2004image}, LPIPS~\cite{zhang2018unreasonable}, FID~\cite{Seitzer2020FID}, and FID-VID~\cite{heusel2017gans, Seitzer2020FID}, as illustrated in Fig. \ref{fig:evaluation1} and Table \ref{tab:exp}. 
All the test videos are standardized to a resolution of 512$\times$512. 
\new{
The experimental results indicate that different methods exhibit varying strengths across different metrics. 
MagicAnimate achieves the highest scores in SSIM (\new{0.7558}) and PSNR (\new{18.1666}), demonstrating its superiority in maintaining structural integrity and image quality. 
UniAnimate shows remarkable performance by achieving the best scores in LPIPS (\new{0.2496}), FID (\new{73.8868}), and FID-VID (\new{12.6624}), indicating its excellence in perceptual quality and temporal consistency. 
Animate Anyone leads in the L1 metric (\new{0.0779}), suggesting its effectiveness in producing outputs that closely align with the ground truth. 
Overall, while MagicAnimate excels in image quality metrics, UniAnimate demonstrates superior performance in perceptual and video-specific metrics, making it particularly effective for video generation tasks.
We visualize most common visual artifacts and quality issues of these methods in Fig.~\ref{fig:challenges}.
The visualization and experimental results reveal common issues in most current holistic human motion video generation approaches.
}

\begin{table}[!t]
    \vspace{-10px}
    \centering
    \caption{Quantitative performances of nine open-source methods on the TikTok test dataset for pose-guided dance video 
    generation.}
    \label{tab:exp}
    \resizebox{\linewidth}{!}{%
    \begin{tabular}{@{}p{2.6cm} p{0.5cm} p{0.63cm} p{0.56cm} p{0.5cm} p{0.85cm} p{0.95cm}@{}}
    \toprule
    & \multicolumn{1}{c}{SSIM} & \multicolumn{1}{c}{PSNR} & \multicolumn{1}{c}{LPIPS} & \multicolumn{1}{c}{L1} & \multicolumn{1}{c}{FID} & \multicolumn{1}{c}{FID-VID} \\
    & \multicolumn{1}{c}{(↑)} & \multicolumn{1}{c}{(↑)} & \multicolumn{1}{c}{(↓)} & \multicolumn{1}{c}{(↓)} & \multicolumn{1}{c}{(↓)} & \multicolumn{1}{c}{(↓)}         \\ \midrule
    DisCo \cite{wang2024disco}              & \new{0.6645}               & \new{14.0959}              & \new{0.3911}               & \new{0.1284}                   & \new{117.6868}             & \new{79.9380}             \\
    Champ \cite{zhu2024champ}               & \new{0.7042}               & \new{16.1558}              & \new{0.3236}               & \new{0.0984}                   & \new{90.4657}             & \new{20.1173}             \\
    MagicPose \cite{chang2024magicpose}    & \new{0.7427}               & \new{17.2344}              & \new{0.2765}               & \new{0.0842}                   & \new{76.1036}   & \new{49.8028}             \\
    Animate Anyone \cite{hu2024animate}     & \new{0.7397}               & \new{17.8857}              & \new{0.2671}               & \textbf{\new{0.0779}}    
                                                                                                                                            & \new{87.0663}             & \new{21.0533}            \\
    MagicAnimate \cite{xu2024magicanimate}  & \textbf{\new{0.7558}}    
                                                                   & \textbf{\new{18.1666}} 
                                                                                          & \new{0.2506}  
                                                                                                                 & \new{0.0827}                   & \new{82.1320}             & \new{24.1402}             \\
    \new{MimicMotion} \cite{zhang2024mimicmotion}              & \new{0.6689}               & \new{15.3815}              & \new{0.3417}               & \new{0.1138}                   & \new{85.4918}             & \new{17.4705}             \\
    \new{UniAnimate} \cite{wang2024unianimate}              & \new{0.7436}               & \new{17.9129}              & \textbf{\new{0.2496}}               & \new{0.0785}                   & \textbf{\new{73.8868}}             & \textbf{\new{12.6624}}             \\
    \new{StableAnimation} \cite{tu2024stableanimator}              & \new{0.7433}               & \new{17.7722}              & \new{0.2601}               & \new{0.0845}                   & \new{90.9057}             & \new{21.5695}             \\
    \new{Animate-X}~\cite{tan2024animate}              & \new{0.7167}               & \new{16.1982}              & \new{0.3087}               & \new{0.0994}                   & \new{83.8176}             & \new{17.2247}             \\
                                                                                                                 \bottomrule
    \end{tabular}%
    }
    \vspace{-10px}
    \end{table}

\vspace{-10px}
\subsection{Datasets}

Video generation models require a large quantity of data to learn the prior distribution of human knowledge, making data a crucial component. Since crawling data to train human motion models is sensitive considering current privacy issues, to facilitate the training of human action generation models in future research, we carefully collect 64 publicly available human-related video datasets to support future research in this area. Table \ref{tab:datasets} provides an overview of these datasets along with their respective download links. Unlike other dataset overviews~\cite{survey1,survey2}, we include a concise one-sentence description for each dataset, enabling researchers to quickly identify the most relevant data for their work.


\begin{table*}[!ht]
    \vspace{-15px}
    \renewcommand{\arraystretch}{1.0}
    \centering
    \caption{Overview of key attributes across various human video datasets.
    \new{More datasets can be found in appendix F.}
    }
    \label{tab:datasets}
    \resizebox{0.965\textwidth}{!}{%
    \begin{tabular}{|ccccccccccc|}
    \hline
    \rowcolor[HTML]{D1DADC} 
    \multicolumn{1}{|c|}{\cellcolor[HTML]{D1DADC} \textbf{Item} } & \multicolumn{1}{c|}{\cellcolor[HTML]{D1DADC}\textbf{Dataset}}                                                                                                                                                                              & \multicolumn{1}{c|}{\cellcolor[HTML]{D1DADC}\textbf{Year}} & \multicolumn{1}{c|}{\cellcolor[HTML]{D1DADC}\textbf{Hours}} & \multicolumn{1}{c|}{\cellcolor[HTML]{D1DADC}\textbf{Resolution}} & \multicolumn{1}{c|}{\cellcolor[HTML]{D1DADC}\textbf{ID}} & \multicolumn{1}{c|}{\cellcolor[HTML]{D1DADC}\textcolor{black}{\textbf{Camera}}} & \multicolumn{1}{c|}{\cellcolor[HTML]{D1DADC}\textbf{Person}} & \multicolumn{1}{c|}{\cellcolor[HTML]{D1DADC} \textbf{\begin{tabular}[c]{@{}c@{}}Clean \\ Background\end{tabular}}} & \multicolumn{1}{c|}{\cellcolor[HTML]{D1DADC}\textbf{Description}} & \textbf{\begin{tabular}[c]{@{}c@{}}\new{Support} \\ \new{Tasks}\end{tabular}}                                                                      \\ [0.8ex]  
    \hline
    \multicolumn{11}{|c|}{\cellcolor[HTML]{EFEFEF}\raisebox{-0.8ex}[0pt][0pt]{\new{\textbf{Only Head}}}} \\ [0.8ex] \hline
    \multicolumn{1}{|c|}{1}                                     & \multicolumn{1}{c|}{\href{https://liangbinxie.github.io/projects/vfhq/}{\textcolor{blue}{VFHQ   }} \cite{xie2022vfhq}}                                                                                                                     & \multicolumn{1}{c|}{2022}                                  & \multicolumn{1}{c|}{1800}                                   & \multicolumn{1}{c|}{Mid}                                         & \multicolumn{1}{c|}{16000}                               & \multicolumn{1}{c|}{Static}                                  & \multicolumn{1}{c|}{Multiple}                                & \multicolumn{1}{c|}{\nologo}                                           & \multicolumn{1}{c|}{A high-quality video face dataset} & \palogo \textfacePartlogo \talkingheadlogo                                                         \\ \hline
    \multicolumn{1}{|c|}{2}                                    & \multicolumn{1}{c|}{\href{https://celebv-hq.github.io/}{\textcolor{blue}{CelebV-HQ   }} \cite{zhu2022celebvhq}}                                                                                                                            & \multicolumn{1}{c|}{2022}                                  & \multicolumn{1}{c|}{68}                                     & \multicolumn{1}{c|}{Mid}                                         & \multicolumn{1}{c|}{>1000}                               & \multicolumn{1}{c|}{Static}                                  & \multicolumn{1}{c|}{Single}                                  & \multicolumn{1}{c|}{\nologo}                                           & \multicolumn{1}{c|}{A   face video dataset} & \palogo \textfacePartlogo \talkingheadlogo                                                                   \\ \hline
    \multicolumn{1}{|c|}{3}                                    & \multicolumn{1}{c|}{\href{https://github.com/midusi/lsa-t}{\textcolor{blue}{LSA-T   }} \cite{dal2022lsa}}                                                                                                                                  & \multicolumn{1}{c|}{2022}                                  & \multicolumn{1}{c|}{22}                                     & \multicolumn{1}{c|}{High}                                        & \multicolumn{1}{c|}{103}                                 & \multicolumn{1}{c|}{Static}                                  & \multicolumn{1}{c|}{Multiple}                                & \multicolumn{1}{c|}{\nologo}                                           & \multicolumn{1}{c|}{The first continuous Argentinian sign language dataset} & \palogo \posevideologo \textmotionlogo                   \\ \hline
    \multicolumn{1}{|c|}{4}                                    & \multicolumn{1}{c|}{\href{https://github.com/celebv-text/CelebV-Text/tree/main}{\textcolor{blue}{CelebV-Text   }} \cite{yu2022celebvtext}}                                                                                                 & \multicolumn{1}{c|}{2022}                                  & \multicolumn{1}{c|}{279}                                    & \multicolumn{1}{c|}{Mid}                                         & \multicolumn{1}{c|}{>1000}                               & \multicolumn{1}{c|}{Static}                                  & \multicolumn{1}{c|}{Single}                                  & \multicolumn{1}{c|}{\nologo}                                           & \multicolumn{1}{c|}{A   large dataset of facial text-video pairs} & \textfacePartlogo \textmotionlogo \talkingheadlogo                                              \\ \hline
    \multicolumn{11}{|c|}{\cellcolor[HTML]{EFEFEF}\raisebox{-0.8ex}[0pt][0pt]{\new{\textbf{Half-body Data}}}} \\ [0.8ex] \hline
    \multicolumn{1}{|c|}{5}                                    & \multicolumn{1}{c|}{\href{https://project.mhzhou.com/vico/}{\textcolor{blue}{Vico   }} \cite{zhou2022responsive}}                                                                                                                          & \multicolumn{1}{c|}{2022}                                  & \multicolumn{1}{c|}{1.5}                                    & \multicolumn{1}{c|}{Low}                                         & \multicolumn{1}{c|}{92}                                  & \multicolumn{1}{c|}{Static}                                  & \multicolumn{1}{c|}{Single}                                  & \multicolumn{1}{c|}{\nologo}                                           & \multicolumn{1}{c|}{A   dataset of face-to-face conversation} & \palogo \posevideologo \textmotionlogo \talkingheadlogo                                                 \\ \hline
    \multicolumn{1}{|c|}{6}                                    & \multicolumn{1}{c|}{\href{https://github.com/snap-research/articulated-animation}{\textcolor{blue}{TED-Dataset   }} \cite{siarohin2021motion}}                                                                                             & \multicolumn{1}{c|}{2021}                                  & \multicolumn{1}{c|}{3}                                      & \multicolumn{1}{c|}{Low}                                         & \multicolumn{1}{c|}{411}                                 & \multicolumn{1}{c|}{Static}                                  & \multicolumn{1}{c|}{Single}                                  & \multicolumn{1}{c|}{\nologo}                                           & \multicolumn{1}{c|}{Videos   of presentations that include movement and audio} & \posevideologo \textmotionlogo \holisticbodylogo                                \\ \hline
    \multicolumn{1}{|c|}{7}                                    & \multicolumn{1}{c|}{\href{https://www.yasamin.page/hdnet_tiktok}{\textcolor{blue}{TikTok   Dataset }} \cite{Jafarian_2021_CVPR_TikTok}}                                                                                                    & \multicolumn{1}{c|}{2021}                                  & \multicolumn{1}{c|}{1}                                      & \multicolumn{1}{c|}{Mid}                                         & \multicolumn{1}{c|}{>300}                                & \multicolumn{1}{c|}{Static}                                  & \multicolumn{1}{c|}{Single}                                  & \multicolumn{1}{c|}{\rightlogo}                                        & \multicolumn{1}{c|}{Social   media dance videos} & \posevideologo \dancelogo \textmotionlogo                                                               \\ \hline
    \multicolumn{1}{|c|}{8}                                     & \multicolumn{1}{c|}{\href{https://github.com/chevalierNoir/OpenASL/tree/main}{\textcolor{blue}{OpenASL   }} \cite{shi2022open}}                                                                                                            & \multicolumn{1}{c|}{2022}                                  & \multicolumn{1}{c|}{288}                                    & \multicolumn{1}{c|}{Low}                                         & \multicolumn{1}{c|}{>200}                                & \multicolumn{1}{c|}{Static}                                  & \multicolumn{1}{c|}{Single}                                  & \multicolumn{1}{c|}{\rightlogo}                                        & \multicolumn{1}{c|}{An open-domain sign language translation dataset} & \palogo \textfacePartlogo \textmotionlogo                                           \\ \hline
    \multicolumn{1}{|c|}{9}                                    & \multicolumn{1}{c|}{\href{https://how2sign.github.io/}{\textcolor{blue}{How2sign   }} \cite{Duarte_CVPR2021}}                                                                                                                              & \multicolumn{1}{c|}{2021}                                  & \multicolumn{1}{c|}{80}                                     & \multicolumn{1}{c|}{High}                                        & \multicolumn{1}{c|}{11}                                  & \multicolumn{1}{c|}{Static}                                  & \multicolumn{1}{c|}{Single}                                  & \multicolumn{1}{c|}{\nologo}                                           & \multicolumn{1}{c|}{A multiview continuous American sign language dataset} & \palogo \posevideologo \textmotionlogo                                    \\ \hline
    \multicolumn{1}{|c|}{10}                                    & \multicolumn{1}{c|}{\href{https://gewu-lab.github.io/MUSIC-AVQA/}{\textcolor{blue}{MUSIC-AVQA   }} \cite{Li2022Learning}}                                                                                                                  & \multicolumn{1}{c|}{2022}                                  & \multicolumn{1}{c|}{150}                                    & \multicolumn{1}{c|}{Mid}                                         & \multicolumn{1}{c|}{>1000}                               & \multicolumn{1}{c|}{Moving}                                  & \multicolumn{1}{c|}{Multiple}                                & \multicolumn{1}{c|}{\nologo}                                           & \multicolumn{1}{c|}{A   dataset for audio-visual question answering task} &  \posevideologo \textmotionlogo                                      \\ \hline
    \multicolumn{11}{|c|}{\cellcolor[HTML]{EFEFEF}\raisebox{-0.8ex}[0pt][0pt]{\new{\textbf{Full-body Data}}}} \\ [0.8ex] \hline
    \multicolumn{1}{|c|}{11}                                     & \multicolumn{1}{c|}{\href{https://etsin.fairdata.fi/dataset/5b8c55b5-7e4d-492d-9f8a-df707bd56de2}{\textcolor{blue}{HEADSET   }} \cite{10.23729/c086f60d-300d-4d44-928c-dd057846a3c9}}                                                      & \multicolumn{1}{c|}{2022}                                  & \multicolumn{1}{c|}{20}                                     & \multicolumn{1}{c|}{Mid}                                         & \multicolumn{1}{c|}{27}                                  & \multicolumn{1}{c|}{Static}                                  & \multicolumn{1}{c|}{Multiple}                                & \multicolumn{1}{c|}{\nologo}                                           & \multicolumn{1}{c|}{A multimodal XR dataset with facial expressions} & \dancelogo \posevideologo \textmotionlogo                            \\ \hline
    \multicolumn{1}{|c|}{12}                                    & \multicolumn{1}{c|}{\href{https://github.com/Shunli-Wang/TSA-Net?tab=readme-ov-file}{\textcolor{blue}{FR-FS   Dataset }} \cite{TSA-Net}}                                                                                                   & \multicolumn{1}{c|}{2021}                                  & \multicolumn{1}{c|}{1}                                      & \multicolumn{1}{c|}{Low}                                         & \multicolumn{1}{c|}{>10}                                 & \multicolumn{1}{c|}{Moving}                                  & \multicolumn{1}{c|}{Multiple}                                & \multicolumn{1}{c|}{\nologo}                                           & \multicolumn{1}{c|}{Fall recognition in figure skating} & \dancelogo \posevideologo                                      \\ \hline
    \multicolumn{1}{|c|}{13}                                    & \multicolumn{1}{c|}{\href{https://grail.cs.washington.edu/projects/background-matting-v2/\#/datasets}{\textcolor{blue}{VideoMatte240K   }} \cite{lin2021real}}                                                                             & \multicolumn{1}{c|}{2021}                                  & \multicolumn{1}{c|}{3}                                      & \multicolumn{1}{c|}{High}                                        & \multicolumn{1}{c|}{484}                                 & \multicolumn{1}{c|}{Static}                                  & \multicolumn{1}{c|}{Multiple}                                & \multicolumn{1}{c|}{\rightlogo}                                        & \multicolumn{1}{c|}{High-resolution   alpha matte and foreground video clips} & \dancelogo \posevideologo \textmotionlogo                                  \\ \hline
    \multicolumn{1}{|c|}{14}                                     & \multicolumn{1}{c|}{\href{https://github.com/seonokkim/3DYoga90}{\textcolor{blue}{3DYoga90   }} \cite{kim20233dyoga90}}                                                                                                                    & \multicolumn{1}{c|}{2023}                                  & \multicolumn{1}{c|}{80}                                     & \multicolumn{1}{c|}{High}                                        & \multicolumn{1}{c|}{>200}                                & \multicolumn{1}{c|}{Moving}                                  & \multicolumn{1}{c|}{Single}                                  & \multicolumn{1}{c|}{\rightlogo}                                        & \multicolumn{1}{c|}{A Yoga   dataset} & \posevideologo \textmotionlogo \dancelogo                                                                           \\ \hline
    \multicolumn{1}{|c|}{15}                                     & \multicolumn{1}{c|}{\href{https://visym.github.io/cap/}{\textcolor{blue}{CAP   }} \cite{Byrne2023Fine}}                                                                                                                                    & \multicolumn{1}{c|}{2023}                                  & \multicolumn{1}{c|}{1000}                                   & \multicolumn{1}{c|}{Low}                                         & \multicolumn{1}{c|}{780}                                 & \multicolumn{1}{c|}{Static}                                  & \multicolumn{1}{c|}{Single}                                  & \multicolumn{1}{c|}{\nologo}                                           & \multicolumn{1}{c|}{A   large dataset for fine-grained activity classfication} & \posevideologo \textmotionlogo \holisticbodylogo                                 \\ \hline
    \multicolumn{1}{|c|}{16}                                     & \multicolumn{1}{c|}{\href{https://github.com/Kitware/MEVID}{\textcolor{blue}{MEVID   }} \cite{Davila2023mevid}}                                                                                                                            & \multicolumn{1}{c|}{2023}                                  & \multicolumn{1}{c|}{144}                                    & \multicolumn{1}{c|}{Mid}                                         & \multicolumn{1}{c|}{176}                                 & \multicolumn{1}{c|}{Moving}                                  & \multicolumn{1}{c|}{Single}                                  & \multicolumn{1}{c|}{\nologo}                                           & \multicolumn{1}{c|}{A   dataset for large-scale, video person re-identification} & \posevideologo \textmotionlogo \holisticbodylogo                          \\ \hline
    \multicolumn{11}{|c|}{\cellcolor[HTML]{EFEFEF}\raisebox{-0.8ex}[0pt][0pt]{\new{\textbf{General Data (Including Human Data)}}}} \\ [0.8ex] \hline
    \multicolumn{1}{|c|}{17}             & \multicolumn{1}{c|}{\href{https://github.com/showlab/loveu-tgve-2023}{\textcolor{blue}{LOVEU-TGVE-2023   }} \cite{wu2023tune}}                                                                                     & \multicolumn{1}{c|}{2023}          & \multicolumn{1}{c|}{0.6}            & \multicolumn{1}{c|}{Low}                                         & \multicolumn{1}{c|}{76}                                  & \multicolumn{1}{c|}{Static}                                  & \multicolumn{1}{c|}{Single}                                  & \multicolumn{1}{c|}{\rightlogo}                                        & \multicolumn{1}{c|}{Text-driven   video editing competition} & \textmotionlogo \posevideologo \holisticbodylogo                                            \\ \hline
    \multicolumn{1}{|c|}{18}                                     & \multicolumn{1}{c|}{\href{https://mbzuai-oryx.github.io/Video-ChatGPT/}{\textcolor{blue}{VideoChatGPT   }} \cite{Maaz2023VideoChatGPT}}                                                                                                    & \multicolumn{1}{c|}{2023}                                  & \multicolumn{1}{c|}{30}                                     & \multicolumn{1}{c|}{Low}                                         & \multicolumn{1}{c|}{>1000}                               & \multicolumn{1}{c|}{Moving}                                  & \multicolumn{1}{c|}{Multiple}                                & \multicolumn{1}{c|}{\nologo}                                           & \multicolumn{1}{c|}{A   dataset for video-based conversation} & \textmotionlogo \holisticbodylogo \talkingheadlogo                                                  \\ \hline
    \multicolumn{1}{|c|}{19}                                     & \multicolumn{1}{c|}{\href{https://github.com/doc-doc/NExT-GQA}{\textcolor{blue}{NExTVideo   }} \cite{xiao2024can}}                                                                                                                         & \multicolumn{1}{c|}{2024}                                  & \multicolumn{1}{c|}{65}                                     & \multicolumn{1}{c|}{Mid}                                         & \multicolumn{1}{c|}{>1000}                               & \multicolumn{1}{c|}{Moving}                                  & \multicolumn{1}{c|}{Multiple}                                & \multicolumn{1}{c|}{\nologo}                                           & \multicolumn{1}{c|}{A   dataset for video grounding extended from NExt-QA} & \posevideologo \textmotionlogo \holisticbodylogo                                    \\ \hline
    \multicolumn{1}{|c|}{20}                                    & \multicolumn{1}{c|}{\href{https://github.com/dmoltisanti/brace/}{\textcolor{blue}{BRACE   BRACE }} \cite{moltisanti22brace}}                                                                                                               & \multicolumn{1}{c|}{2022}                                  & \multicolumn{1}{c|}{3}                                      & \multicolumn{1}{c|}{Mid}                                         & \multicolumn{1}{c|}{81}                                  & \multicolumn{1}{c|}{Moving}                                  & \multicolumn{1}{c|}{Multiple}                                & \multicolumn{1}{c|}{\rightlogo}                                        & \multicolumn{1}{c|}{A breakdancing competition dataset for dance motion synthesis} & \dancelogo \posevideologo \textmotionlogo                         \\ \hline
\multicolumn{11}{|l|}{\new{\palogo: Portrait Animation;  \dancelogo: Dance Video Generation; \posevideologo: Pose2Video; \textfacePartlogo: Text2Face; \textmotionlogo: Text2MotionVideo; \liplogo: Lip Sync; \talkingheadlogo: Talking Head; \holisticbodylogo: Audio-Driven Holistic Body Driving }} \\
\hline
    \end{tabular}%
    }
    \vspace{-10px}
    \end{table*}

\vspace{-10px}
\section{Challenges and Future Work}
\label{sec:Challenges}
\vspace{-4px}

Human motion video generation presents several challenges, including data availability, signal understanding, motion planning, and the quality of the generated video, as shown in Fig.~\ref{fig:challenges}.

\textbf{Lack of Data. }
The field of human motion video generation is hindered by limited data availability due to privacy concerns, poor quality, and high collection costs. This scarcity weakens model robustness and compromises real-world reliability. Expanding datasets is essential for training models to better recognize, understand, and replicate human behavior, ultimately improving the quality and diversity of generated videos.


\textbf{Motion Planner. }
Current motion planning relies heavily on pre-existing data distributions, which limits its ability to grasp the deeper semantic layers of human actions. This approach constrains adaptability and sophistication. To advance motion planning, we must shift from purely statistical methods to those that incorporate meaning, context, and intent. Leveraging LLMs can enhance this process by analyzing complex human movement patterns, enabling the generation of more realistic and contextually appropriate human motion in video, which is treated as a promising direction \cite{geng2023affective,wang2023agentavatar,wang2024instructavatar}.

\textbf{Lack of Photorealism. }
In human motion video generation, numerous challenges, as illustrated in Fig. \ref{fig:challenges}, require sophisticated solutions. The fidelity of generated human forms, particularly in the face and hands, is crucial for realism and expressiveness.

Maintaining visual consistency throughout a video is another challenge, requiring not only stable visual characteristics but also seamless integration of the subject within the environment.
The plausibility of movements is equally critical. Regardless of how realistic the forms and backgrounds appear, if the depicted actions are not physically or logically sound, the overall effect is compromised. 

\textbf{Expanding Duration and Refining Control.}
Most current methods for human motion video generation produce only short clips, usually lasting a few seconds. Extending this to longer videos, spanning minutes or even hours, remains a major challenge. Future research should aim to develop techniques that maintain coherence and quality over extended durations. Additionally, current multimodal approaches, even with signals like meshes and depth maps, often struggle with detailed control of specific body parts. Enhancing the realism and expressiveness of videos requires improved control over intricate areas such as hands and faces.


\textbf{Real-Time Platform and Cost. }
Real-time streaming of virtual humans demands low latency to ensure smooth and natural interactions, while high latency disrupts communication and reduces lifelike responsiveness. High-quality graphics also need substantial bandwidth, with limited bandwidth leading to lower video quality and buffering issues. Additionally, the user interface must be highly responsive, providing immediate feedback to keep users engaged. While diffusion models \cite{yang2023diffusion} are pivotal in human motion video generation, their high computational demands, as shown in Appendix G, necessitate the development of more efficient models to lower costs and improve accessibility.


\textbf{Ethics. } 
Creating digital humanoids introduces significant ethical concerns, particularly regarding the use of personal data and privacy. A robust ethical framework is needed to guide their development and integration, ensuring informed consent for biometric data and establishing accountability for their actions. Safeguarding privacy and addressing negative impacts are essential. 




{
\tiny
		\bibliographystyle{ieeetr}
		\bibliography{main}
}

\vspace{-15px}
\section{Biography Section}

\footnotesize
{
\noindent \textbf{Haiwei Xue}
received his B.S. degree from the Guangdong University of Foreign Studies in 2022. He is currently pursuing a master’s degree at the Tsinghua University. His research interests focus on AIGC, including 2D/3D digital human generation, multimodal large language models, and multimodal sentiment analysis. 

\noindent \textbf{Xiangyang Luo}
received his B.S. degree from the University of Electronic Science and Technology of China in 2024. He is currently pursuing a master’s degree at the Tsinghua University. His research interests focus on AIGC.

\noindent \textbf{Zhanghao Hu}
is an intern at 01.AI and a graduate Master of Science student in Artificial Intelligence at the University of Edinburgh, UK. His research centers on multi-modal natural language processing models and human-machine interactive AI.

\noindent \textbf{Xin Zhang}
received the B.S. degree from Minzu University of China in 2022. He is currently pursuing the master's degree with the School of Mathematics and Statistics, Xi'an Jiaotong University.

\noindent \textbf{Xunzhi Xiang}
received his B.S. degree from Shanghai Jiao Tong University in 2021 and master's degree from University of Chinese Academic of Sciences in 2024. His research interests focus on AIGC, including 2D/3D digital human and 3D scene and asset generation.

\noindent \textbf{Yuqin Dai}
received the B.E. and MA.Eng. degree from the Nanjing University of Science and Technology, Nanjing, China, in 2022 and 2025. Her research interests include computer vision, especially in 3D content generation, cross-modal generation, and 3D reconstruction.

\noindent \textbf{Jianzhuang Liu}
(Senior Member, IEEE) received the PhD degree in computer vision from The Chinese University of Hong Kong, in 1997. From 1998 to 2000, he was a research fellow with Nanyang Technological University, Singapore. From 2000 to 2012, he was a post-doctoral fellow, an assistant professor, and an adjunct associate professor with The Chinese University of Hong Kong, Hong Kong. In 2011, he joined the Shenzhen Institute of Advanced Technology, University of Chinese Academy of Sciences, Shenzhen, China, as a professor. He was a principal researcher in Huawei Company from 2012 to 2023. He has authored more than 200 papers in the areas of computer vision, image processing, deep learning, and AIGC.

\noindent \textbf{Zhensong Zhang}
received the B.E. degree from Xidian University in 2011, the M.S. degree from the University of Chinese Academy of Sciences in 2014, and the Ph.D. degree from The Chinese University of Hong Kong in 2018. He is currently a senior research engineer with the Huawei Noah’s Ark Lab. His research interests include MLLM, AIGC and 3DGS. He published more than 20 articles at international venues, including CVPR, ECCV, and IJCV.

\noindent \textbf{Minglei Li}
holds a Ph.D. from The Hong Kong Polytechnic University. His primary research focuses on digital humans and large language models.

\noindent \textbf{Jian Yang}
Received a PhD in pattern recognition and intelligence systems from NUST in 2002.
Now a Chang-Jiang professor at NUST, he has published over 200 papers with 50,000+ Google Scholar citations.
His research interests include pattern recognition, computer vision, and machine learning. Currently, he is/was an associate editor of Pattern Recognition, Pattern Recognition Letters, IEEE Trans. Neural Networks and Learning Systems, and Neurocomputing. He is a Fellow of IAPR.

\noindent \textbf{Fei Ma}
is currently a researcher at Guangdong Laboratory of Artificial Intelligence and Digital Economy (SZ). Before that, he received the B.S. degree in Communication Engineering from University of Electronic Science and Technology of China in 2017 and the Ph.D. degree in Information and Communication Engineering from Tsinghua University in 2022. 
His research interests include generative AI, and multimodal learning.

\noindent \textbf{Zhiyong Wu}
(Member, IEEE) received the B.S. and Ph.D. degrees in computer science and technology from Tsinghua University, Beijing, China, in 1999 and 2005, respectively. From 2005 to 2007, he was a Postdoctoral Fellow with the Department of Systems Engineering and Engineering Management, The Chinese University of Hong Kong (CUHK), Hong Kong. He then joined the Graduate School at Shenzhen (now Shenzhen International Graduate School), Tsinghua University, Shenzhen, China, and is currently an Associate Professor. He is also a Coordinator with the Tsinghua-CUHK Joint Research Center for Media Sciences, Technologies and Systems. His research interests include intelligent speech interaction, more specially, speech processing, text-to-audio-visual-speech synthesis, and natural language understanding and generation. He is a Member of International Speech Communication Association (ISCA) and China Computer Federation (CCF).

\noindent \textbf{Changpeng Yang}
received Ph.D. from the ATMRI of Nanyang Technological University, Singapore. He was also affiliated with NEXTOR II at the University of California, Berkeley, USA. 

\noindent \textbf{Zonghong Dai}
is primarily responsible for AI infrastructure, digital humans, and B2B commercialization, as a co-founder of 01.AI. He is dedicated to creating globally leading foundational models and product applications.

\noindent \textbf{Fei Richard Yu}
(Fellow, IEEE) is a Professor at College of Computer Science and Software Engineering, Shenzhen University. Before that, He received the Ph.D. degree in electrical engineering from The University of British Columbia (UBC), Vancouver, BC, Canada, in 2003. He is currently an Elected Member of the Board of Governors of IEEE VTS. He is a fellow of the Canadian Academy of Engineering (CAE), the Engineering Institute of Canada (EIC), and IET. He received several best paper awards from some first-tier conferences. Since 2019, he has been named in the Clarivate Analytics list of Highly Cited Researchers. He is the Editor-in-Chief of the IEEE VTS Mobile World Newsletter. 
He is a Distinguished Lecturer of IEEE in both VTS and ComSoc. 
His research interests include autonomous machine intelligence, embodied intelligence, and information networking.

}





\vfill

\end{document}